\documentclass[a4paper,fleqn]{cas-sc}

\usepackage{caption}
\usepackage{subfigure}
\usepackage{algorithm}
\usepackage{algorithmic}

\usepackage{float}
\usepackage{graphicx}
\usepackage{bm}
\usepackage{bookmark}
\usepackage{cite}

\def\tsc#1{\csdef{#1}{\textsc{\lowercase{#1}}\xspace}}
\tsc{WGM}
\tsc{QE}

\begin{document}
\let\WriteBookmarks\relax
\def\floatpagepagefraction{1}
\def\textpagefraction{.001}

\shorttitle{UAV 3-D path planning based on MOEA/D with adaptive areal weight adjustment}    

\shortauthors{Yougang Xiao et al.}  

\title [mode = title]{UAV 3-D path planning based on MOEA/D with adaptive areal weight adjustment}  

\author[1]{Yougang Xiao}
\credit{Conceptualization, Resources, Supervision}

\author[1]{Hao Yang}
\credit{Conceptualization, Methodology, Software, Validation, Formal analysis, Investigation, Data Curation, Writing - Original Draft, Writing - Review \& Editing, Visualization}

\author[1]{Huan Liu}
\credit{Validation, Writing - Review \& Editing, Project administration}

\author[2]{Keyu Wu}
\credit{Supervision, Project administration}

\author[1]{Guohua Wu\corref{cor1}}
\ead{guohuawu@csu.edu.cn}
\credit{Writing - Review \& Editing, Supervision, Funding acquisition}
\cortext[cor1]{ Corresponding author.}

\affiliation[1]{organization={School of Traffic and Transportation Engineering, Central South University},
            city={Changsha 410075},
            country={China}}
\affiliation[2]{organization={College of Systems Engineering and the College of Electronic Sciences, National University of Defense Technology},
            addressline={}, 
            city={Changsha 410073},
            country={China}}
            
\let\printorcid\relax       

\begin{abstract}
Unmanned aerial vehicles (UAVs) are desirable platforms for time-efficient and cost-effective task execution. 3-D path planning is a key challenge for task decision-making. This paper proposes an improved multi-objective evolutionary algorithm based on decomposition (MOEA/D) with an adaptive areal weight adjustment (AAWA) strategy to make a tradeoff between the total flight path length and the terrain threat. AAWA is designed to improve the diversity of the solutions. More specifically, AAWA first removes a crowded individual and its weight vector from the current population and then adds a sparse individual from the external elite population to the current population. To enable the newly-added individual to evolve towards the sparser area of the population in the objective space, its weight vector is constructed by the objective function value of its neighbors. The effectiveness of MOEA/D-AAWA is validated in twenty synthetic scenarios with different number of obstacles and four realistic scenarios in comparison with other three classical methods.
\end{abstract}

\begin{highlights}
\item Introduction of multi-objective path planning with sharp peak and low tail PFs.
\item AAWA enables individuals to evolve towards sparser areas on PFs.
\item The AAWA improves the convergence and diversity of the population.
\item The MOEA/D-AAWA achieves shorter paths and safer paths.
\end{highlights}


\maketitle

\section{Introduction}\label{cha1}

Due to the advantages of low cost, small size, high flexibility and strong concealment \cite{huang2020reliable, wang2019joint}, UAVs have been widely exploited for various fields, such as meteorological detection \cite{sziroczak2022review}, emergency rescue \cite{xiong2023multi}, traffic monitoring \cite{fan2022deep}, and target reconnaissance \cite{pokhrel2019mobility}. As an important part of task planning, path planning aims to provide feasible flight paths for UAVs. Reasonable path planning schemes can effectively avoid potential threats and reduce flight cost \cite{ramirez2017solving}. However, it is still a significant challenge to provide decision-makers with a diverse and feasible set of paths, while considering the UAV maneuverability and obstacle avoidance.

Currently, path planning methods have achieved effective results in various fields. However, unlike traditional unmanned ground vehicles \cite{liu2022modified,tokekar2016sensor}, the motion space of UAVs has an additional dimension (i.e., height), i.e., UAVs can perform ascent and descent movements. Therefore, UAV path planning in 3-D space is more in line with the flight reality \cite{zhao2018survey}. In addition, limiting UAVs in the 2-D space with a fixed altitude will lead to the loss of promising solutions. The consideration of the third dimension will inevitably bring challenges in additional maneuverability of UAVs, such as the minimum flight altitude, maximum horizontal angle, and maximum vertical angle. In order to simulate real UAV flight, the abovementioned three factors are considered in this article.

Although the difficulty of path planning increases with an additional space dimension \cite{jones2023path}, there are still many relevant studies \cite{wu2018hybrid, phung2021safety, zhang2018social, radmanesh2018grey}. However, these studies only focus on a single objective or construct the objective function as a weighted combination. Determining a weight vector to cater to decision-maker preferences is a hard task, especially if there are many requirements and constraints \cite{peng2022decomposition}. Furthermore, a flight path with a specific weight vector cannot meet the diverse task needs. From our point of view, it is essential to consider a multi-objective optimization problem (MOP) to provide a diverse set of solutions for path planning tasks. Specifically, our path planning problem simultaneously optimizes two conflicting objectives (i.e., the path length and the degree of threats). The shorter path can diminish task execution time, and the safer path can reduce damage rates of UAVs.

Due to the conflicting objectives of MOPs, the optimization of one objective typically accompanies by the deterioration of at least one of the other objectives \cite{chen2019hyperplane}. Thus, a sole solution with optimal values for all objectives almost does not exist in a MOP. The main goal of solving MOPs is to obtain a set of compromise solutions. Owing to the population-based nature, multi-objective evolutionary algorithms (MOEAs) can generate a set of solutions in one run \cite{chen2019adaptive, qiu2020ensemble}. Therefore, MOEAs, such as NSGA-II \cite{ahmed2011multi}, MOEA/D \cite{peng2022decomposition}, MOPSO \cite{guo2020global}, have been developed to deal with MOPs, and have been applied in path planning in recent years. 

During the iteration of MOEAs, the population diversity plays a significant role in the convergence and uniformity of the population \cite{chen2019adaptive}. First, lacking diversity weakens the population ability to leave the current local optimal subspace, thereby preventing the population from further approaching to the Pareto front (PF). In addition, the population diversity determines whether individuals can be uniformly distributed on PF simultaneously. In particular, when the target MOP has a complex PF \cite{qi2014moea,trivedi2016survey}, such as the PF with a sharp peak and a low tail in our paper, it is hard to search for the sharp peak and the low tail. The complexity of PFs may lead to the loss of the population diversity and uniformity, and even the inability of population to leave the current local optimal subspace.

The above observations motivate us to propose an improved MOEA/D with a new adaptive weight adjustment strategy. MOEA/D decomposes a MOP into a number of scalar optimization subproblems represented by a set of uniformly distributed wight vectors. Due to the decomposition-based nature of MOEA/D, it is well feasible to improve the population diversity by reasonably adjusting the weight vectors. Therefore, we design a new weight adjustment strategy. The weight vectors are periodically adjusted in the iteration to provide an accurate direction for the evolution of individuals to sparse areas. First, in the current population, the crowded individuals with high similarity to other individuals are removed. And then the same number of individuals who have high dissimilarity to the individuals of the current population are added to the current population. To ensure the population convergence, the newly-added individuals are selected from an external elite population, which stores nondominated individuals during the iteration. To enable the newly-added individuals to evolve towards a sparser area of the current population in the objective space, we improved the calculation method of its weight vector, which use the objective function value of its neighbor to calculate its weight vectors.

Based on the above analysis, the contributions of this paper can be summarized as follows. 

1) We introduce a bi-objective UAV path planning problem under 3-D scenarios, which aims to minimize path length and the degree of terrain threats simultaneously.

2) To improve population diversity, an adaptive areal weight adjustment strategy is designed. To be specific, a crowded individual in the current population and its weight vector are deleted, while the individual with the maximum sparsity selected from the external elite population and its weight vector calculated by the objective function of neighbors are added. This strategy contributes to navigating the evolution of solutions towards sparse areas on PFs.

3) We conduct extensive experiments compared with other three state of the art methods in twenty synthetic scenarios with different number of obstacles and four realistic scenarios to verify the effectiveness of the proposed method.

The rest of this paper is organized as follows. Section \ref{cha2} reviews the related works. Section \ref{cha3} states the mathematical model of UAV 3-D multi-objective path planning. Section \ref{cha4} presents the proposed algorithm in detail. Section \ref{cha5} reports a series of experiments and comparative results. Finally, conclusions are drawn in Section \ref{cha6}.

\section{Literature review}\label{cha2}

Path planning can be generalized as seeking a collision-free between two positions based on certain evaluation criteria \cite{bortoff2000path}. So far, the path planning methods can be grouped into five categories: 1) methods based mathematical model: linear programming algorithm \cite{radmanesh2016flight}, dynamic programming algorithm \cite{fabiani2007autonomous}, etc; 2) methods based on geometric search: Voronoi diagram \cite{chen2014uav, pehlivanoglu2012new}, probabilistic roadmaps \cite{kavraki1996probabilistic, xu2021autonomous}, rapidly-exploring random tree \cite{karaman2011anytime, zhang2020novel}, elliptic tangent graph method \cite{liu2020autonomous}, etc; 3) methods based on heuristic algorithms: Dijkstra \cite{li2021openstreetmap}, A-Star \cite{wu2010multi, zhang2007path}, etc; 4) methods based on the artificial potential field \cite{fox1997dynamic}; 5) methods based on evolutionary and swarm intelligence algorithms: particle swarm optimization \cite{zhang2007path, cheng20123}, ant colony optimization \cite{you2016chaotic, zhao2012improved}, genetic algorithm \cite{joines1994use, xiao2013flight, roberge2018fast}, differential evolution \cite{yu2020constrained, liu2005fuzzy}, etc.

In recent years, UAV 3-D path planning methods have been attracted much attention. Nevertheless, some of them achieve the purpose of reducing problem complexity by strategies that diminish a dimension of 3-D planning space, that is, converting computationally expensive 3-D path planning problems into more manageable 2-D path planning problems. The most commonly used is altitude fixing strategy. Chen and Gesbert \cite{chen2017optimal} fixed the UAV at a predetermined height in a 3-D urban space to simplify the 3-D path planning problem. This strategy has extended to multiple UAVs. Zhao et al. \cite{zhao2017uav} assumed that all UAVs share the same urban airspace at the same altitude. It might actually cause a high density of air traffic. Dai et al. \cite{dai2020efficient} fixed each UAV at a unique flight height to avoid collisions among UAVs. Although this strategy can simplify the 3-D path planning problem, the solution space will inevitably shrink significantly, resulting in the loss of promising solutions. Besides, the fixed altitude may be hard to maintain due to topography, weather, or task targets. 

The complexity of path planning has significantly increased in a non-simplified 3-D space. First, the diverse range of UAV airborne movements in a 3-D space means an exponential growth in potential paths available. Second, the 3-D space places higher requirements on the constraints of UVA maneuverability and obstacle avoidance \cite{jones2023path}. With considering multiple optimization objectives, the difficulty of solving the problem increases further. The traditional method of solving a MOP is to transform it into a single objective problem by the weighted sum method. For instance, Xu et al. \cite{xu2010chaotic} designed an objective function of the UAV path planning problem by weightedly summing the threat cost and the fuel cost. In order to improve UAV range and avoid detection by enemy radars, Roberge et al. \cite{roberge2018fast} also minimized a weighted-sum objective function of fuel consumption and average flying altitude. Although this weighted sum method has been developed to find achievable paths for path planning, the determination of ideal weights is another consideration. Some researchers tries to find the ideal weights. For example, Masehian and Sedighizadeh \cite{masehian2010multi} constructed an optimization objective function by means of assigning weights to the length and smoothness. The optimal weight of (1, 0.25) was obtained through extensive simulation experiments. Nazarahari et al. \cite{nazarahari2019multi} proposed a method to balance the order of magnitude of three objective function terms to determine a set of optimal weights. The optimization objective is still the weighted sum of the terms.

However, real-world problems commonly involve several conflicting factors, which are difficult to optimize simultaneously. In general, a solution that is optimal for all objectives does not exist. Thus, it is necessary to generate a set of nondominated solutions for decision-makers, known as multi-objective manner. Ahmed and Deb \cite{ahmed2011multi} proposed a path planning method based on nondominated sorting genetic algorithm II (NSGA-II) for discrete grid environments, optimizing both path length and degree of threats. Guo et al. \cite{guo2020global} developed a chaotic and sharing-learning particle swarm optimization (CSPSO) algorithm to optimize path length, smoothness, stranding risk, economic cost, and safety. These two methods only were used in 2-D space. Multi-objective path planning in 3-D space has been considered in several studies. Adhikari et al. \cite{adhikari2017fuzzy} aimed to simultaneously minimize the fuel and the threat cost as well as finding the shortest path. Sun et al. \cite{sun2016path} provided multiple feasible paths for the GEO-UAV bistatic SAR system with different tradeoffs between navigation for UAV and bistatic SAR imaging performance. Peng and Qiu \cite{peng2022decomposition} designed a local infeasibility utilization mechanism to optimize the flight path length and the degree of terrain threats. Wan et al. \cite{wan2022accurate} also optimized these two terms according to an improved multi-objective ant colony algorithm combining the preferred search direction and random neighborhood search mechanism. Although these methods achieve competitive results in the experiment, it is still a significant challenge for multi-objective path planning in 3-D space to obtain a well-distributed and diverse set of solutions.

In brief, we propose an improved MOEA/D named MOEA/D-AAWA for short. The proposed MOEA/D-AAWA can obtain a well-distributed and diverse set of nondominated solutions on complex PFs.

\section{Model of UAV path planning}\label{cha3}

\subsection{Representation of UAV path }\label{cha3.1}

B-spline curve is used to represent the UAV path by given control points \cite{nikolos2003evolutionary}. Suppose that the UAV flies from the starting point $p_{0} =( {x_{0},y_{0},z_{0}} )$ to the target point $p_{n + 1} = ( x_{n + 1},y_{n + 1},z_{n + 1} )$. B-spline curve, i.e., the UAV path, is generated from $(n+2)$ control points $( ( {x_{0},y_{0},z_{0}} ),\ldots,( {x_{i},y_{i},z_{i}} ),\ldots,( {x_{n+1},y_{n+1},z_{n+1}} ) )$. The UAV path is composed of $(s+1)$ path points $( {( {x_{0}^{'},y_{0}^{'},z_{0}^{'}} ),\ldots,( {x_{j}^{'},y_{j}^{'},z_{j}^{'}} ),\ldots,( {x_{s}^{'},y_{s}^{'},z_{s}^{'}} )} )$, which can be calculated as Eq.(\ref{eqn2}):
\begin{align}
\label{eqn2} \left\{\begin{aligned}
&x_{j}^{'} = {\sum\limits_{i = 0}^{n + 1}{x_{i}B_{i,k}( t_{j} )}}\\
&y_{j}^{'} = {\sum\limits_{i = 0}^{n + 1}{y_{i}B_{i,k}( t_{j} )}}\\
&z_{j}^{'} = {\sum\limits_{i = 0}^{n + 1}{z_{i}B_{i,k}( t_{j} )}}
\end{aligned}\right.
,j = 0,1,\ldots,s,
\end{align}
where $k$ is the degree of B-spline curve. The larger value of $k$, the smoother of the curve. $B_{i,k}( t_{j} )$ is the blending function of the curve and defined recursively in terms of a non-decreasing sequence of real numbers by using Eq.(\ref{eqn3})-(\ref{eqn5}). The related parameters are set to $k=3$, $s=150$.
\begin{align}
\label{eqn3} B_{i,1}( t_{j} ) =
\left\{\begin{aligned}
&1,u_{i} \leq t_{j} < u_{i + 1},\\
&0,otherwise,
\end{aligned}\right.
\end{align}
\begin{align}
\label{eqn4} B_{i,k}( t_{j} ) = \frac{t_{j} - u_{i}}{u_{i + k - 1} - u_{i}}B_{i,k - 1}( t_{j} ) + \frac{u_{i + k} - t_{j}}{u_{i + k} - u_{i + 1}}B_{i + 1,k - 1}( t_{j} ),
\end{align}
\begin{align}
\label{eqn5} u_{i} =
\left\{\begin{aligned}
&0,i \leq k,\\
&i - k,k < i \leq s - 1,\\
&s - k,i \geq s,
\end{aligned}\right.
\end{align}
where $t$ is a set of discrete points in the range between 0 to $(n+k+2)$ with a constant step.
\begin{figure}[pos=htb]
    \centering
        \subfigure[3-D View]{\includegraphics[width=4.6cm]{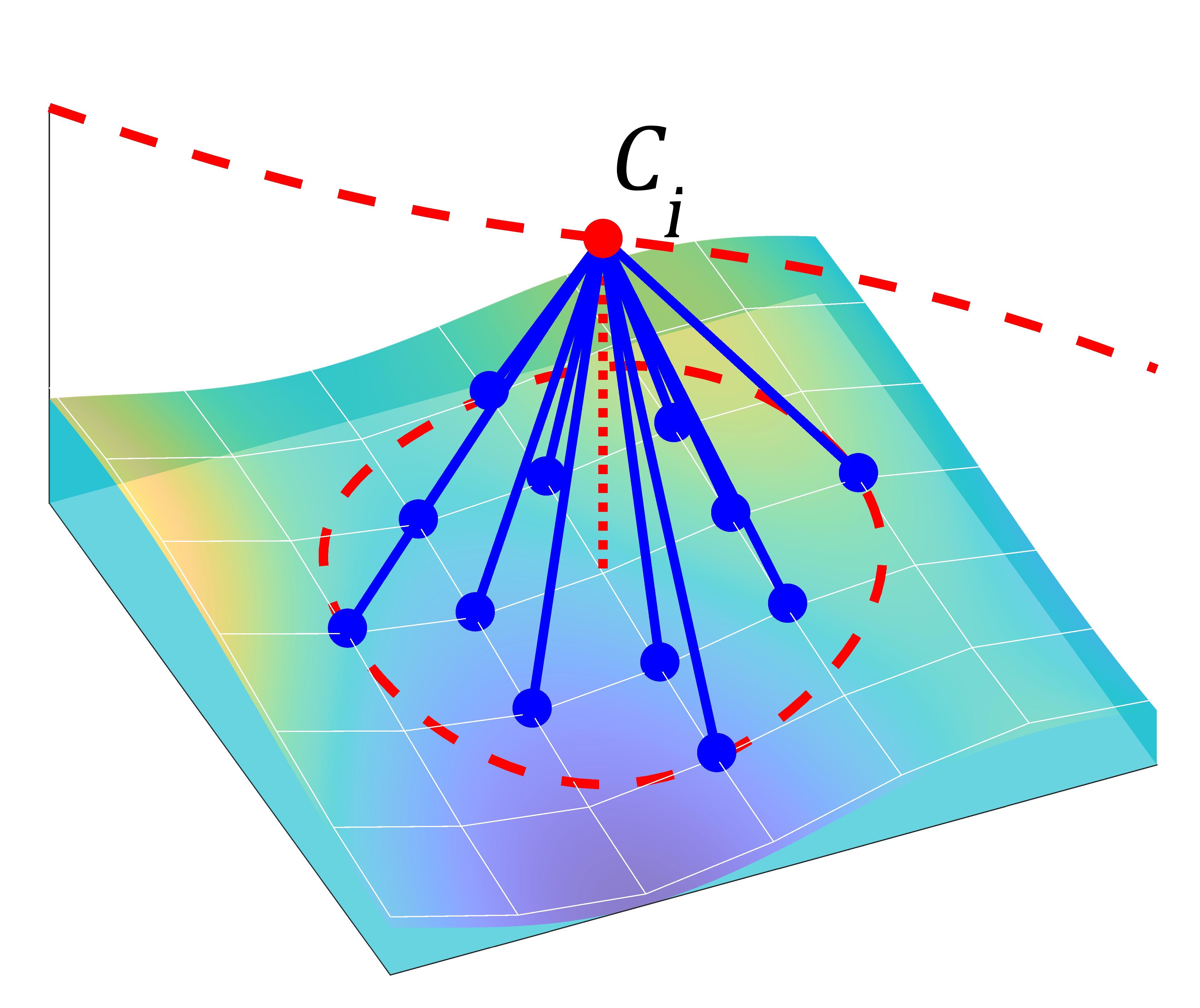}}
        \subfigure[Vertical view]{\includegraphics[width=4.6cm]{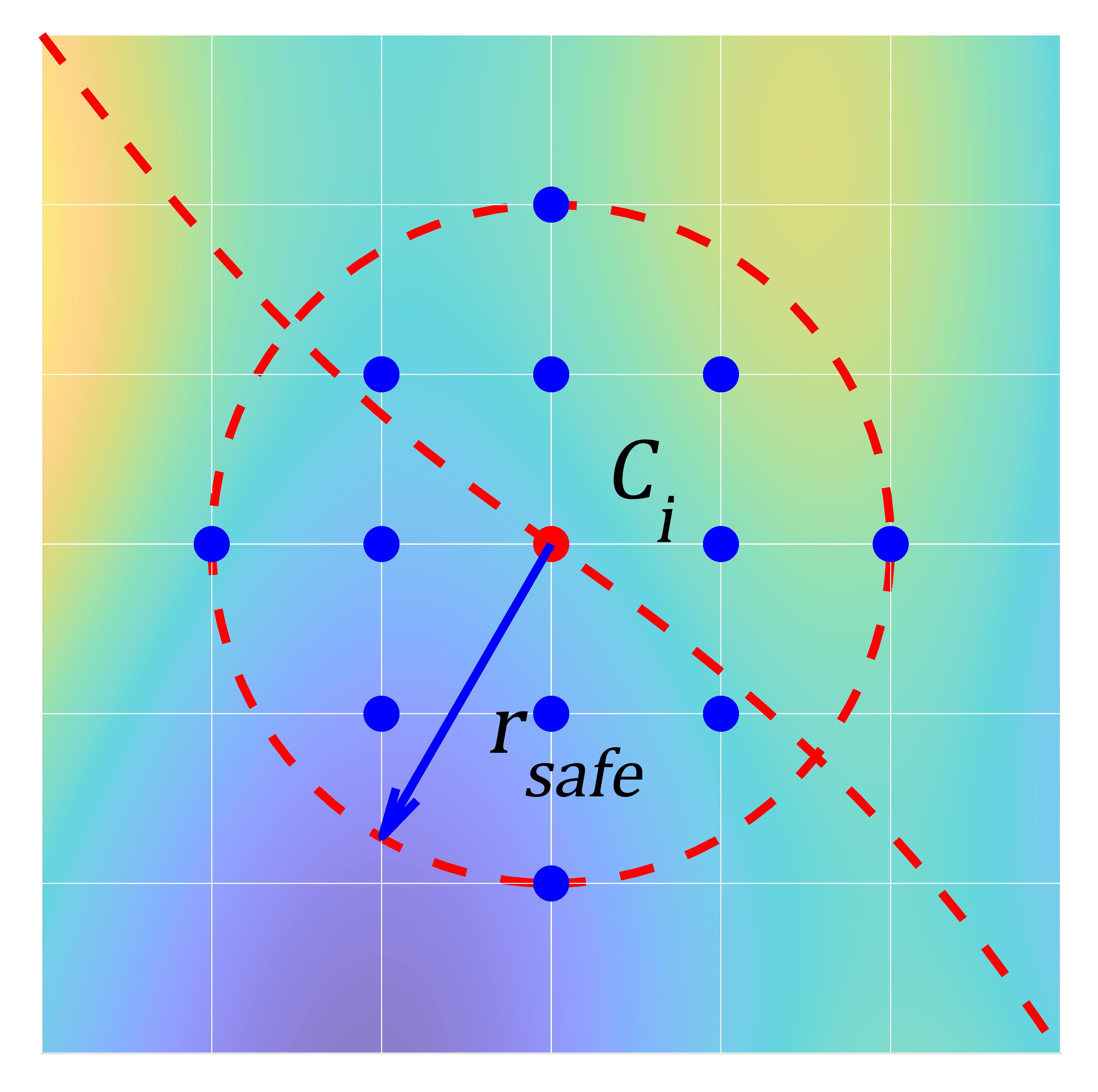}}
    \caption{Terrain grid points within the safe range}
    \label{fig1}
\end{figure}

\subsection{Objective function}\label{cha3.2}

In this paper, UAV path planning problem is considered as a bi-objective optimization problem that minimizes overall path length $f_{1}$ and degree of threats $f_{2}$ as follows.
\begin{align}
\label{eqn6} \min{F} = ( {f_{1},f_{2}} ).
\end{align}

The length $f_{1}$ can be calculated by the Euclidean operator:
\begin{align}
\label{eqn7} f_{1} = {\sum_{i = 0}^{s-1}{\sqrt{{\Delta x_{i}^{'}}^{2} + {\Delta y_{i}^{'}}^{2} + {\Delta z_{i}^{'}}^{2}}}},
\end{align}
where $\Delta x_{i}^{'}=x_{i + 1}^{'} - x_{i}^{'}$,$\Delta y_{i}^{'}=y_{i + 1}^{'} - y_{i}^{'}$,$\Delta z_{i}^{'}=z_{i + 1}^{'} - z_{i}^{'}$.

The $f_{2}$ is used to measure the degree of threats \cite{yu2021knee}:
\begin{align}
\label{eqn8} f_{2} = {\sum_{i = 0}^{s}{\sum_{j = 1}^{n_{g}}( \frac{r_{safe}}{r_{i,j}} )^{2}}},
\end{align}
where $r_{safe}$ is the minimum safe distance between the drone and terrain obstacles. \cite{sun2016path}. $r_{i,j}$ is the Euclidean distance between the path point $C_{i}$ and the terrain grid point $G_{j}$. $n_{g}$ is the number of terrain grid points. As shown in Fig. \ref{fig1}, all terrain grid points $G_{j}$ within a circle with the path point $C_{i}$ as its center and the radius of $r_{safe}$ participate in calculating. The ratio of $r_{safe}$ to $r_{i,j}$ is used as risk assessment of the terrain grid points to the path point.

\subsection{Constraint}\label{cha3.3}

(1) Minimum flight altitude

The UAV path should maintain a safe flight altitude with the terrain. The minimum flight altitude constraint violation of a path point $C_{i}$, i.e., $g_{1}^{i}(p)$, is calculated according to Eq.(\ref{eqn9}). The minimum flight altitude constraint violation of a flight path $g_{1}(p)$ is obtained according to Eq.(\ref{eqn10}).
\begin{align}
\label{eqn9} g_{1}^{i}(p) = 
\left\{\begin{aligned}
&0,z_{i}^{'} - z\left( {x_{i}^{'},y_{i}^{'}} \right) > r_{safe}, \\
&1,otherwise,
\end{aligned}\right.
\end{align}
\begin{align}
\label{eqn10} g_{1}(p) = {\sum_{i = 0}^{s}{g_{1}^{i}(p)}},
\end{align}
where $z\left( {x_{i}^{'},y_{i}^{'}} \right)$ is the topographic elevation of $C_{i}$.

(2) Maximum turning angle

Limited by the inherent maneuverability of the UAV, it can only turn within a certain angle range in the horizontal direction. The constraint violation of maximum turning angle is calculated as shown in Eq.(\ref{eqn11}) and (\ref{eqn12}).
\begin{align}
\label{eqn11} g_{2}^{i}(p) =
\left\{\begin{aligned}
&0,0 \leq \propto_{i} \leq \propto_{max}, \\
&1,otherwise,
\end{aligned}\right.
\end{align}
\begin{align}
\label{eqn12} g_{2}(p) = {\sum_{i = 1}^{s - 1}{g_{2}^{i}(p)}},
\end{align}
where $g_{2}^{i}(p)$ is the constraint violation of the $i$-th turning angle on the path. $g_{2}(p)$ is the maximum turning angle constraint of the path; $\propto_{i}$ is the $i$-th turning angle on the path, which calculated by Eq.(\ref{eqn13}). $\propto_{max}$ is the maximum turning angle, which is 45° in this paper.
\begin{align}
\label{eqn13} \displaystyle{\propto_{i} = \arccos{\frac{\Delta x_{i-1}^{'} \Delta x_{i}^{'} + \Delta y_{i-1}^{'} \Delta y_{i}^{'}}{\sqrt{{\Delta x_{i-1}^{'}}^{2} + {\Delta y_{i-1}^{'}}^{2}} \sqrt{{\Delta x_{i}^{'}}^{2} + {\Delta y_{i}^{'}}^{2}}}}}.
\end{align}

(3) Maximum climbing slope

There should also be a limit on the UAV climbing slope in the vertical direction, i.e., maximum climbing slope. The constraint violation of maximum climbing slope is calculated by Eq.(\ref{eqn14}).
\begin{align}
\label{eqn14} g_{3}^{i}(p) =
\left\{\begin{aligned}
&0,0 \leq \beta_{i} \leq \beta_{max}, \\
&1,otherwise,
\end{aligned}\right.
\end{align}
\begin{align}
\label{eqn15} g_{3}(p) = {\sum_{i = 1}^{s}{g_{3}^{i}(p)}},
\end{align}
where $g_{3}^{i}(p)$ is the constraint violation of the $i$-th climbing slope on the path. $g_{3}(p)$ is the maximum climbing slope constraint of the path. $\beta_{i}$ is the $i$-th climbing slope on the path, which calculated by Eq.(\ref{eqn36}). $\beta_{max}$ is the maximum climbing slope, which is 45° in this paper.
\begin{align}
\label{eqn36}
\beta_{i} = \displaystyle\arctan{\frac{| z_{i}^{'} - z_{i - 1}^{'} |}{\sqrt{\left( {x_{i}^{'} - x_{i - 1}^{'}} \right)^{2} + \left( {y_{i}^{'} - y_{i - 1}^{'}} \right)^{2}}}}.
\end{align}

Finally, the constraint violation $g_{k}(p)$ is normalized to $g_{k}^{'}(p)$, and then calculate the total constraint violation $g(p)$ according to Eq.(\ref{eqn16}).
\begin{align}
\label{eqn16} g(p) = {\sum_{k = 1}^{3}{g_{k}^{'}(p)}}.
\end{align}

\section{Design of MOEA/D-AAWA}\label{cha4}
\subsection{Basic idea}\label{cha4.1}
Zhang and Li \cite{zhang2007moea} proposed MOEA/D, which decomposes a MOP into a number of scalar optimization subproblems. All subproblems are optimized by collaboratively utilizing their neighborhood information and then collectively approach the PF. A subproblem is usually defined by a weight vector and the neighborhood relationship is usually determined by the Euclidean distance between weight vectors.

However, a set of uniformly distributed weight vectors employed in the classic MOEA/D cannot guarantee the uniformity and diversity of the optimal solutions on the PF, especially for complex PFs, such as discontinuous PFs and PFs with a sharp peak and a low tail. Thus, our main goal is to obtain a uniformly distributed and diverse optimal solution set on the PF of the multi-objective path planning problem.

Qi et al. \cite{qi2014moea} proposed an adaptive wight adjustment (AWA) strategy, which periodically removes and adds weight vectors. The newly-added weight vector is calculated by the objective function value of the newly-added individual. In our proposed AAWA, we adopt the same subproblem deletion strategy as AWA, and propose a new subproblem adding strategy (i.e., areal subproblem adding strategy). For the newly-added individual, it weight vector is calculated by the objective function value of its neighbors in the current population rather than itself, and thus the newly-added individual can evolve towards for the sparser area on PFs.

\subsection{WS-transformation}\label{cha4.2}

In this section, we introduce WS-transformation to construct new weight vectors.

Common aggregation methods include weighted sum method, Tchebycheff method, and boundary intersection approach, among which the aggregation function of Tchebycheff method for minimization problem is defined as follows:
\begin{align}
\label{eqn17} \mathop{\min}_{\boldsymbol{x} \in \Omega}g^{tc}(\boldsymbol{x}|\bm{\lambda},\boldsymbol{z}^{\ast}) = \mathop{\min}_{\boldsymbol{x} \in \Omega} \mathop{\max}_{1 \leq i \leq m}{\left\{ \lambda_{i} \left( f_{i} \left( \boldsymbol{x} \right) - z_{i}^{\ast} \right) \right\}},
\end{align}
where $\Omega \in \mathbb{R}^{n}$ is the decision space and $\boldsymbol{x} = (x_{1},\ldots,x_{n})^{T} \in \Omega$ is a decision variable which represents a solution to the target MOP. $g^{tc}(\boldsymbol{x}|\bm{\lambda},\boldsymbol{z}^{\ast})$ is Tchebycheff value. $\boldsymbol{z}^{\ast}=\left( z_{1}^{\ast},\ldots,z_{m}^{\ast} \right)^{T}$ is the reference point, i.e., $z_{i}^{\ast}=\min{\left\{ f_{i} \left( \boldsymbol{x} \right) | \boldsymbol{x} \in \Omega \right\}}$, $i=1,\ldots,m$. $f_{i}(\boldsymbol{x})$ is the objective function of $\boldsymbol{x}$ in $i$-th dimension of objective space for all $i=1,\ldots,m$. $\bm{\lambda} = \left( \lambda_{1},\ldots,\lambda_{m} \right)^{T}$ is the weight vector, $\lambda_{i} \geq 0$ and ${\sum_{i = 1}^{m}\lambda_{i}} = 1, i = 1,2,\ldots,m$. $m$ is the dimension of the objective space.

\begin{figure}[pos=htb]
	\centering
		\includegraphics[width=8cm]{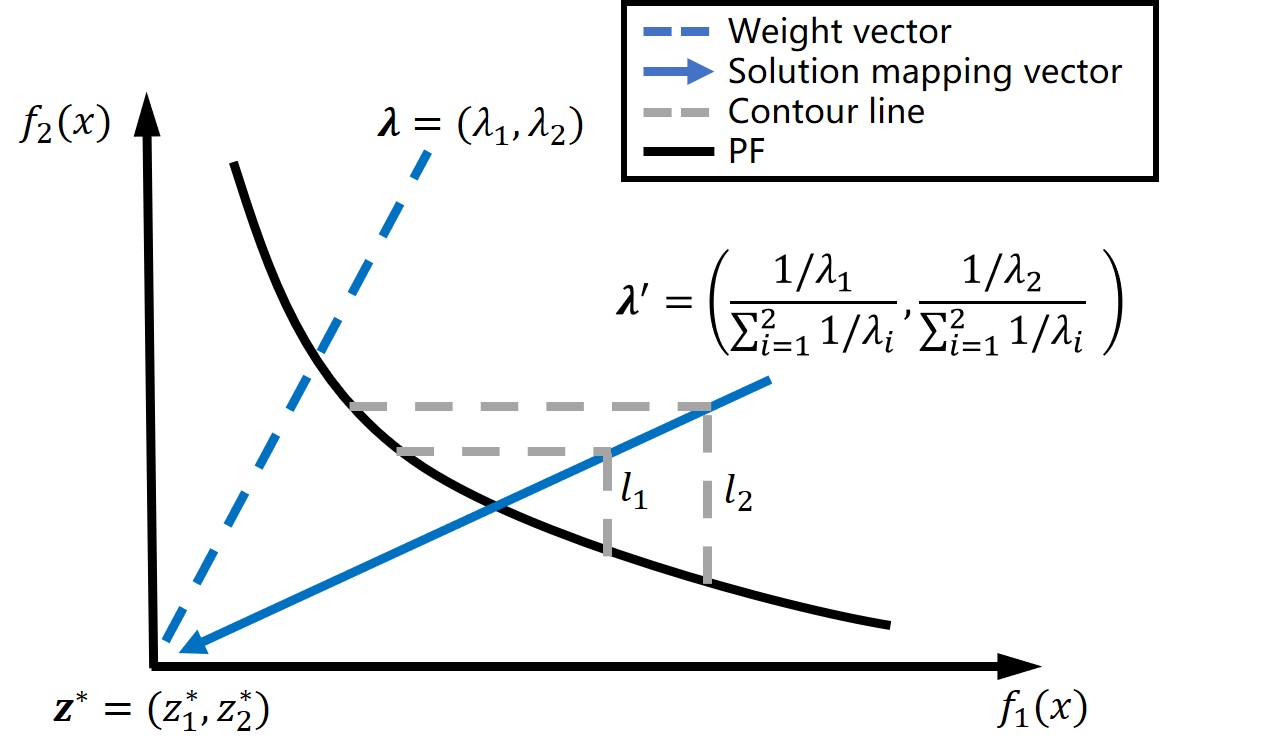}
	  \caption{Individual convergence in a bi-objective space under the Tchebycheff approach}
   \label{fig2}
\end{figure}

Fig. \ref{fig2} shows the individual convergence in a bi-objective space under the Tchebycheff approach. $\bm{\lambda}$ is the weight vector. The individual approaches the PF along the solution mapping vector $\bm{\lambda}^{'}$. $l_{1}$ and $l_{2}$ are two different contour lines of $\bm{\lambda}^{'}$. In the objective space, the Tchebycheff contour line is right-angled serrated, implying that Tchebycheff values of all points on $l_{1}$ are equal and $l_{2}$ is the same. Besides, the Tchebycheff values of all points on $l_{1}$ are less than $l_{2}$.

Based on the MOEA/D using the Tchebycheff method, WS-transformation is introduced to describe how the weight vector and solution mapping vector are transformed. The relevant formula is as follows:
\begin{align}
\label{eqn18} \bm{\lambda} = WS\left( \bm{\lambda}^{'} \right) = WS\left( F-\boldsymbol{z}^{\ast} \right) = \left( \frac{\frac{1}{f_{1} - z_{1}^{\ast}}}{\sum_{i = 1}^{m}\frac{1}{f_{i} - z_{i}^{\ast}}},\ldots,\frac{\frac{1}{f_{m} - z_{m}^{\ast}}}{\sum_{i = 1}^{m}\frac{1}{f_{i} - z_{i}^{\ast}}} \right),
\end{align}
where $\bm{\lambda}^{'}$ is the solution mapping vector, which is the mapping of the weight vector in the objective space. $F$ is the objective function value. For an individual with a known objective function value $F$, its solution mapping vector can be calculated by $\bm{\lambda}^{'} = F-\boldsymbol{z}^{\ast}$, and then the corresponding weight vector $\bm{\lambda}$ can be obtained by applying WS-transformation on $\bm{\lambda}^{'}$.

\subsection{Adaptive areal weight adjustment}\label{cha4.3}

AAWA includes a subproblem deletion strategy and an areal subproblem adding strategy. In the objective space, the subproblem deletion strategy deletes overcrowded subproblems of the current population $Pop$. Meanwhile, the areal subproblem adding strategy selects sparse individuals from the external elite population $EP$, which store a specific number of nondominated individuals during the iteration process. The objective function value of selected sparse individuals' neighbors is used to calculate their solution mapping vectors. Then the corresponding weight vector can be obtained by the WS-transformation.

(1) Subproblem deletion strategy

The sparsity level of individuals in $Pop$ among $Pop$ is evaluated by vicinity distance \cite{kukkonen2006fast} as follows:
\begin{align}
\label{eqn19} SL\left( {x^{i},Pop} \right) = {\prod\limits_{j \in N_{k}^{i}}L_{i,j}},x^{i} \in Pop,
\end{align}
where $L_{i,j}$ is Euclidean distance between the individual $\boldsymbol{x}^{i}$ and its neighbor $\boldsymbol{x}^{j}$. $N_{k}^{i}$ consists of indexes of the $k$-nearest individuals of $\boldsymbol{x}^{i}$ among $Pop$, and $k$ is usually set to the objective function dimension. The smaller sparsity level implies the greater similarity between $\boldsymbol{x}^{i}$ and its neighbors.

The process of deleting subproblems is as follows. First, the sparsity level of each individual in $Pop$  among $Pop$ is calculated by Eq.(\ref{eqn19}). Then, the individual with the minimum sparsity level form $Pop$ is removed and its weight vector is deleted.

\begin{figure}[pos=htb]
    \centering
        \subfigure[Subproblem deletion strategy]{\includegraphics[width=5.4cm]{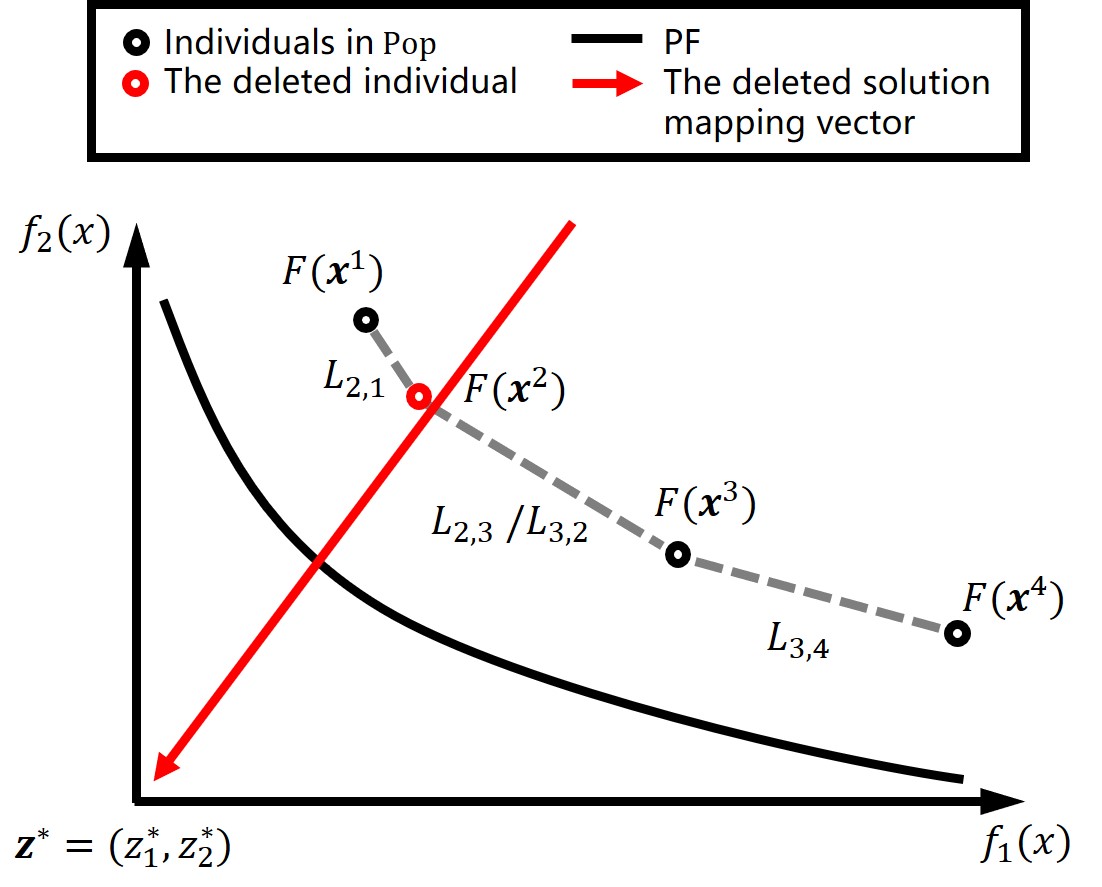}}
        \subfigure[Subproblem adding strategy]{\includegraphics[width=5.4cm]{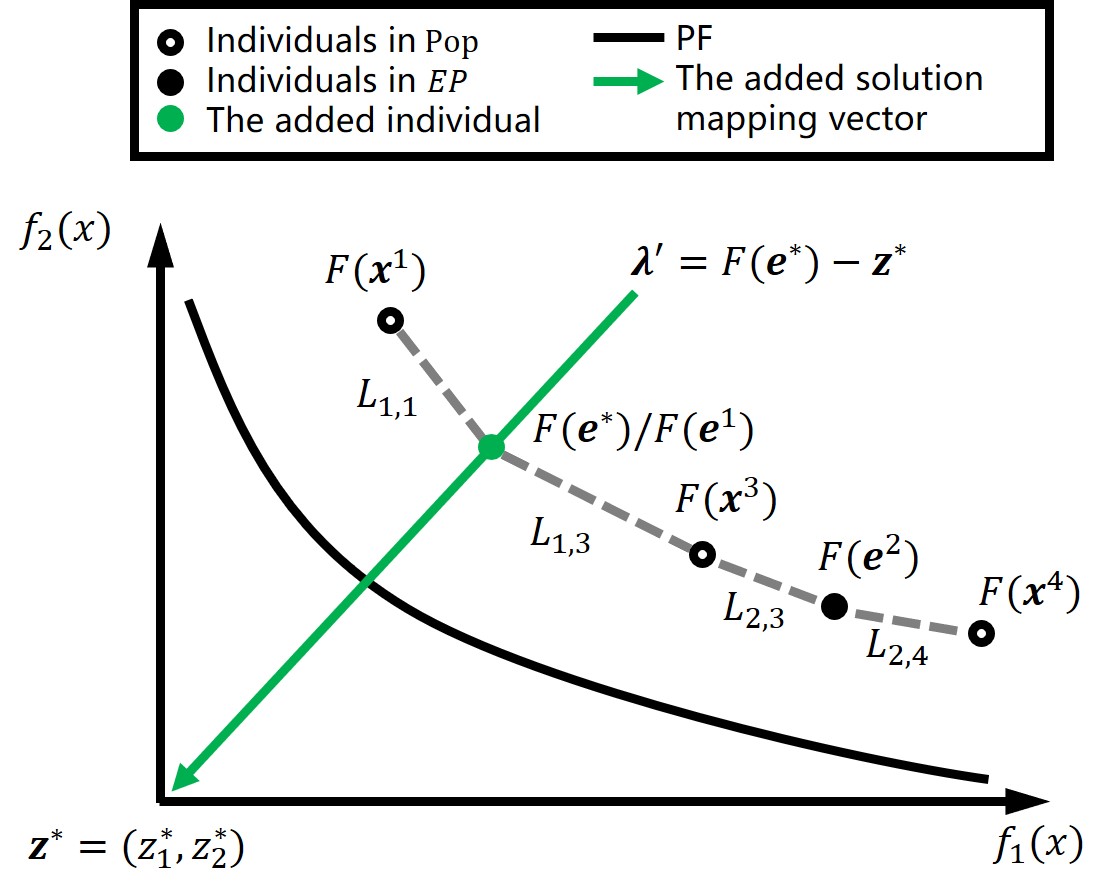}}
        \subfigure[Areal subproblem adding strategy]{\includegraphics[width=5.4cm]{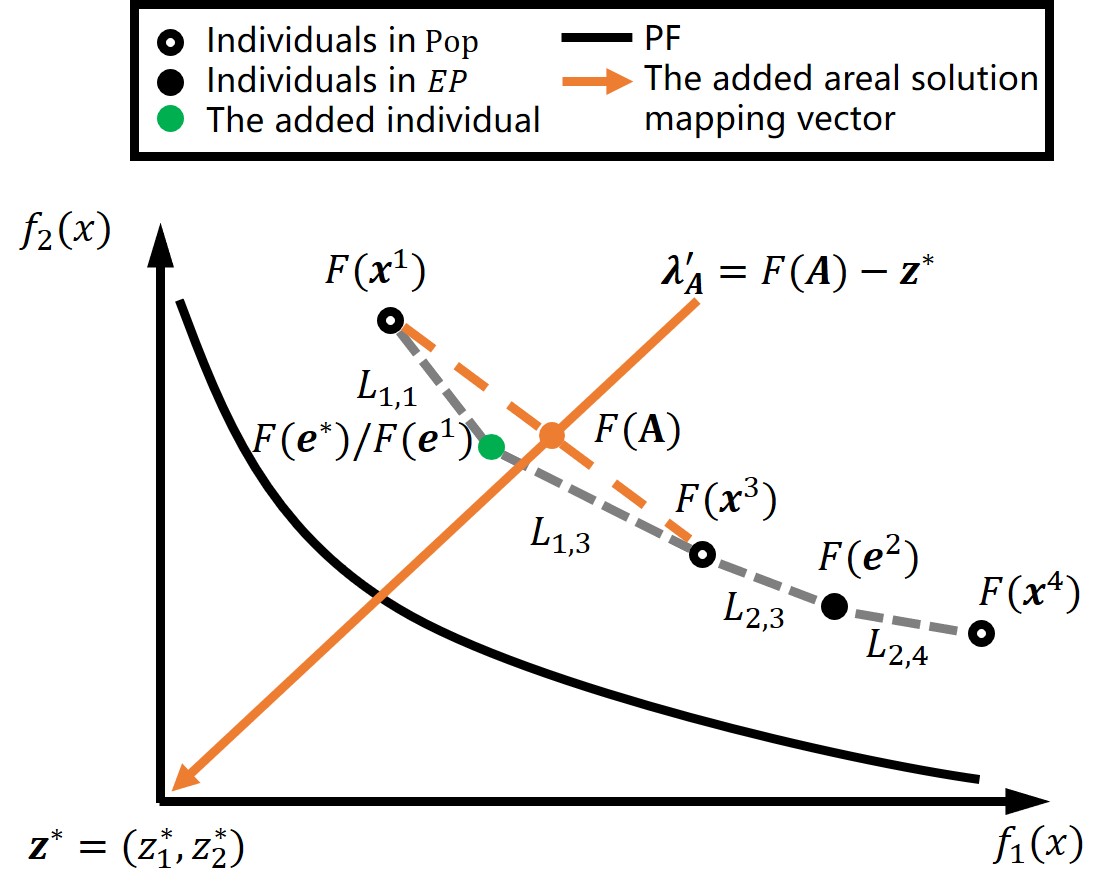}}
    \caption{Adaptive weight adjustment.}
    \label{fig3}
\end{figure}

As shown in Fig. \ref{fig3}(a), there is a population $Pop = \left\{ \boldsymbol{x}^{1},\boldsymbol{x}^{2},\boldsymbol{x}^{3},\boldsymbol{x}^{4} \right\}$. Since the distance between $\boldsymbol{x}^{1}$ and $\boldsymbol{x}^{3}$ is always greater than the distance between $\boldsymbol{x}^{2}$ and $\boldsymbol{x}^{3}$ in the objective space, the $SL\left( {\boldsymbol{x}^{1},Pop} \right)$ is always greater than $SL\left( {\boldsymbol{x}^{2},Pop} \right)$. Similarly, the $SL\left( {\boldsymbol{x}^{4},Pop} \right)$ is always greater than $SL\left( {\boldsymbol{x}^{3},Pop} \right)$. Thus, boundary points as valuable individuals in the population will not be deleted. We just need to compare the sparsity level of $\boldsymbol{x}^{2}$ and $\boldsymbol{x}^{3}$. Apparently, $SL\left( {\boldsymbol{x}^{2},Pop} \right) = L_{2,1}L_{2,3} < SL\left( {\boldsymbol{x}^{3},Pop} \right) = L_{3,2}L_{3,4}$, thus delete $\boldsymbol{x}^{2}$ and its weight vector.

(2) Areal subproblem adding strategy

After deleting subproblems from $Pop$, the same number of new subproblems, which are related to new individuals and their weight vectors, are added to $Pop$. According to the sparsity, the newly-added individuals are selected from $EP$, which is used to store elite individuals (nondominated solutions). The size of $EP$ is set to $1.5 \times |Pop|$. More specifically, the sparsity level of each individual in $EP$ among $Pop$ is firstly calculate according to Eq.(\ref{eqn20}). The elite individual with higher sparsity level, has greater dissimilarity from its neighbors in $Pop$. Then the elite individual with the maximum sparsity level is added to $Pop$ and its weight vector will be constructed.
\begin{align}
\label{eqn20} SL\left( {\boldsymbol{e}^{i},Pop} \right) = {\prod\limits_{j \in N_{k}^{i}}L_{i,j}},\boldsymbol{e}^{i} \in EP,
\end{align}
where $L_{i,j}$ is Euclidean distance between the individual $\boldsymbol{e}^{i}$ and its neighbor $\boldsymbol{x}^{j}$. $N_{k}^{i}$ consists of indexes of the $k$-nearest individuals of $\boldsymbol{e}^{i}$ among $Pop$, and $k$ is usually set to the objective function dimension.

As shown in Fig. \ref{fig3}(b), after deleting $\boldsymbol{x}^{2}$ from $Pop$, $Pop = \left\{ \boldsymbol{x}^{1},\boldsymbol{x}^{3},\boldsymbol{x}^{4} \right\}$. Supposing $EP = \left\{ \boldsymbol{e}^{1},\boldsymbol{e}^{2} \right\}$, $\boldsymbol{x}^{1}$ and $\boldsymbol{x}^{3}$ are neighbors of $\boldsymbol{e}^{1}$, $\boldsymbol{x}^{3}$ and $\boldsymbol{x}^{4}$ are neighbors of $\boldsymbol{e}^{2}$. Apparently, $SL\left( {\boldsymbol{e}^{1},Pop} \right) = L_{1,1}L_{1,3} > SL\left( {\boldsymbol{e}^{2},Pop} \right) = L_{2,3}L_{2,4}$, thus $\boldsymbol{e}^{1}$ is selected as added individual, i.e., $\boldsymbol{e}^{\ast}$.

AWA constructs a new weight vector based on the objective function value of the added individual. The following calculation method can be obtained by substituting the objective function value $F\left( \boldsymbol{e}^{\ast} \right)$ into Eq.(\ref{eqn18}).
\begin{align}
\label{eqn21} \bm{\lambda} = WS\left( \bm{\lambda}^{'} \right) = WS\left( {F\left( \boldsymbol{e}^{\ast} \right) - \boldsymbol{z}^{\ast}} \right) = \left( {\frac{\frac{1}{f_{1}\left( \boldsymbol{e}^{\ast} \right) - z_{1}^{\ast}}}{\sum_{i = 1}^{m}\frac{1}{f_{i}\left( \boldsymbol{e}^{\ast} \right) - z_{i}^{\ast}}},\ldots,\frac{\frac{1}{f_{m}\left( \boldsymbol{e}^{\ast} \right) - z_{m}^{\ast}}}{\sum_{i = 1}^{m}\frac{1}{f_{i}\left( \boldsymbol{e}^{\ast} \right) - z_{i}^{\ast}}}} \right).
\end{align}

\begin{figure}[pos=htb]
	\centering
		\includegraphics[width=7cm]{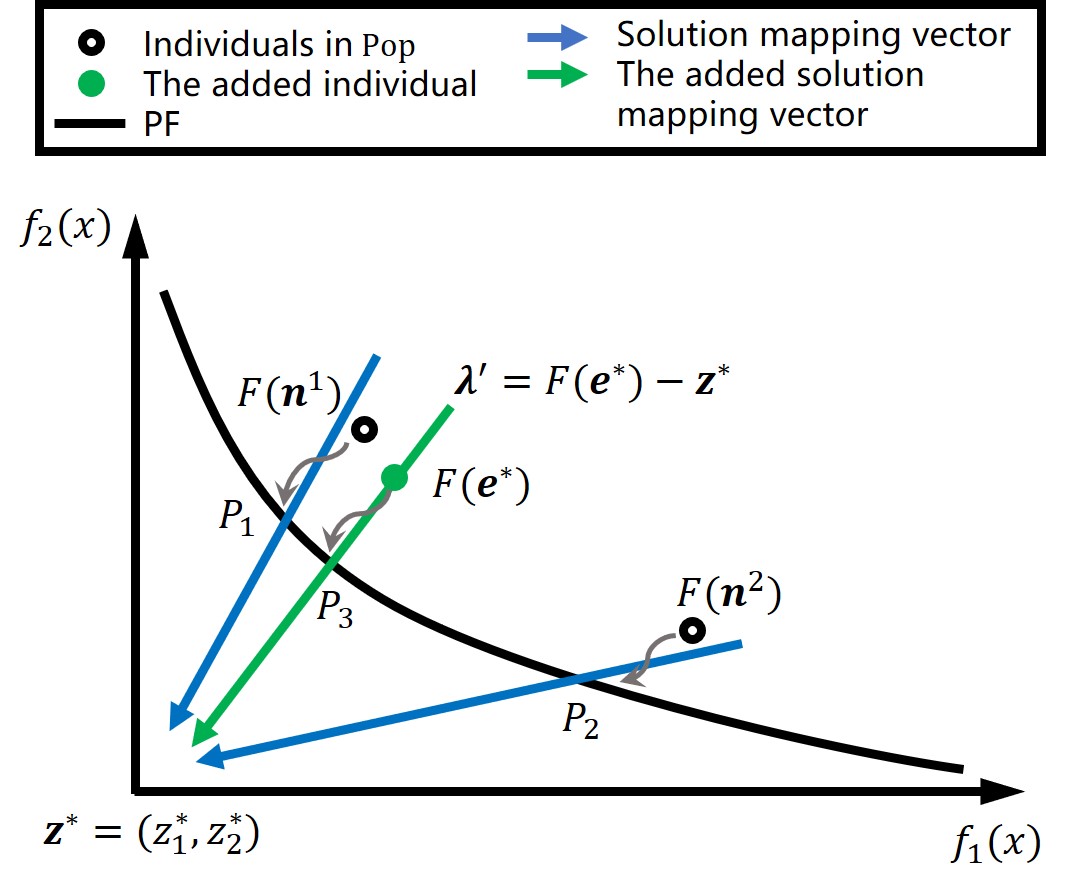}
	  \caption{The deficiency of subproblem adding strategy of AWA}
   \label{fig4}
\end{figure}

However, when the added individual $\boldsymbol{e}^{\ast}$ has a high similarity with one of its neighbors, the added subproblem also has a high similarity with the neighbor subproblem. This cannot guarantee the population diversity. As shown in Fig. \ref{fig4}, suppose that $\boldsymbol{n}^{1}$ and $\boldsymbol{n}^{2}$ are neighbors of $\boldsymbol{e}^{\ast}$, and $F \left( \boldsymbol{n}^{1} \right)$ and $F \left( \boldsymbol{n}^{2} \right)$ are objective function point of the neighbors respectively. $\boldsymbol{n}^{1}$ and $\boldsymbol{n}^{2}$ evolve towards $P_{1}$ and $P_{2}$ on the PF respectively. When $\boldsymbol{e}^{\ast}$ and $\boldsymbol{n}^{1}$ are close in the objective space, that is, there is a high similarity between $\boldsymbol{e}^{\ast}$ and $\boldsymbol{n}^{1}$, and the newly constructed solution mapping vector $\bm{\lambda}^{'}$ also has a high similarity with the solution mapping vector of $\boldsymbol{n}^{1}$. Then that $\boldsymbol{e}^{\ast}$ evolves towards $P_{3}$ which is close to $P_{1}$ is adverse to population diversity.
\begin{algorithm}
\caption{Adaptive areal weight adjustment}
\label{alg1}
\begin{algorithmic}[1]
\REQUIRE current population $Pop$; current weight vector set; external elite population $EP$; the number of adjusting subproblems $nus$      
\ENSURE the adjusted population $Pop$; adjusted weight vector set   

\FOR{$i \gets 1$ to $|Pop|$}
    \IF{$\left. g^{te}\left( \boldsymbol{x}^{j} \middle| {\bm{\lambda}^{i},\boldsymbol{z}^{\ast}} \right) < g^{te}\left( \boldsymbol{x}^{i} \middle| {\bm{\lambda}^{i},\boldsymbol{z}^{\ast}} \right), j = 1,2,\ldots, | Pop | + | EP | \right.$}
        \STATE $\boldsymbol{x}^{i} \gets \boldsymbol{x}^{j},F\left( \boldsymbol{x}^{i} \right) \gets F\left( \boldsymbol{x}^{j} \right)$
    \ENDIF
\ENDFOR

\FOR{$k \gets 1$ to $nus$}
    \STATE Calculate sparsity $SL\left( {\boldsymbol{x},Pop} \right), \boldsymbol{x} \in Pop$ by Eq.(\ref{eqn19});
    \STATE Remove the individual and its weight vector with the minimum sparsity $SL\left( {\boldsymbol{x},Pop} \right)$;
\ENDFOR

\STATE Remove the individuals in $EP$ which are dominated by the individuals in $Pop$;

\FOR{$k \gets 1$ to $nus$}
    \STATE Calculate sparsity $SL\left( {\boldsymbol{e},Pop} \right), \boldsymbol{e} \in EP$ by Eq.(\ref{eqn20});
    \STATE Add the individual with the maximum sparsity $SL\left( {\boldsymbol{e},Pop} \right)$ to the $Pop$;
    \STATE Calculate the areal weight vector by Eq.(\ref{eqn22})-(\ref{eqn23}) and add it to the weight vector set;
\ENDFOR

\end{algorithmic}
\end{algorithm}

To solve the above issue, we propose an areal subproblem adding strategy, which constructs a new weight vector based on the objective function values of the added individual's neighbors. In the $m$-dimensional objective space, $k$ neighbors form an area, i.e., $A$. The centroid coordinate value of $A$ can construct a new weight vector, which is called "the areal weight vector". To be specific, according to Eq.(\ref{eqn22}), we firstly calculate the centroid coordinate of area $A$ composed of $k$ neighbors. Then, we can obtain the corresponding solution mapping vector $\bm{\lambda}_{A}^{'} = F\left( A \right) - \boldsymbol{z}^{\ast}$. Finally, we can construct the new weight vector based on WS-transformation, which can be defined as:
\begin{align}
\label{eqn22} F(A) = \frac{1}{n} {\sum\limits_{i = 1}^{m}{F\left( n^{i} \right)}},
\end{align}
\begin{align}
\label{eqn23} \bm{\lambda}_{A} = WS\left( \bm{\lambda}^{'}_{A} \right) = WS\left( {F\left( A \right) - \boldsymbol{z}^{\ast}} \right) = \left( {\frac{\frac{1}{f_{1}\left( A \right) - z_{1}^{\ast}}}{\sum_{i = 1}^{m}\frac{1}{f_{i}\left( A \right) - z_{i}^{\ast}}},\ldots,\frac{\frac{1}{f_{m}\left( A \right) - z_{m}^{\ast}}}{\sum_{i = 1}^{m}\frac{1}{f_{i}\left( A \right) - z_{i}^{\ast}}}} \right),
\end{align}
where $n^{i}$ is $i$-th neighbor of the added individual among $Pop$. $F \left( n^{i} \right)$ is the objective function values of the $i$-th ($i=1,\ldots,m$) neighbor of the added individual $\boldsymbol{e}^{\ast}$. $F \left( A \right)$ is the centroid coordinate of $A$, which is formed by $k$ neighbors in $m$-dimensional objective space. In this paper, set $k=m$. $\bm{\lambda}_{A}$ is the areal weight vector.
\begin{figure}[pos=htb]
	\centering
		\includegraphics[width=7cm]{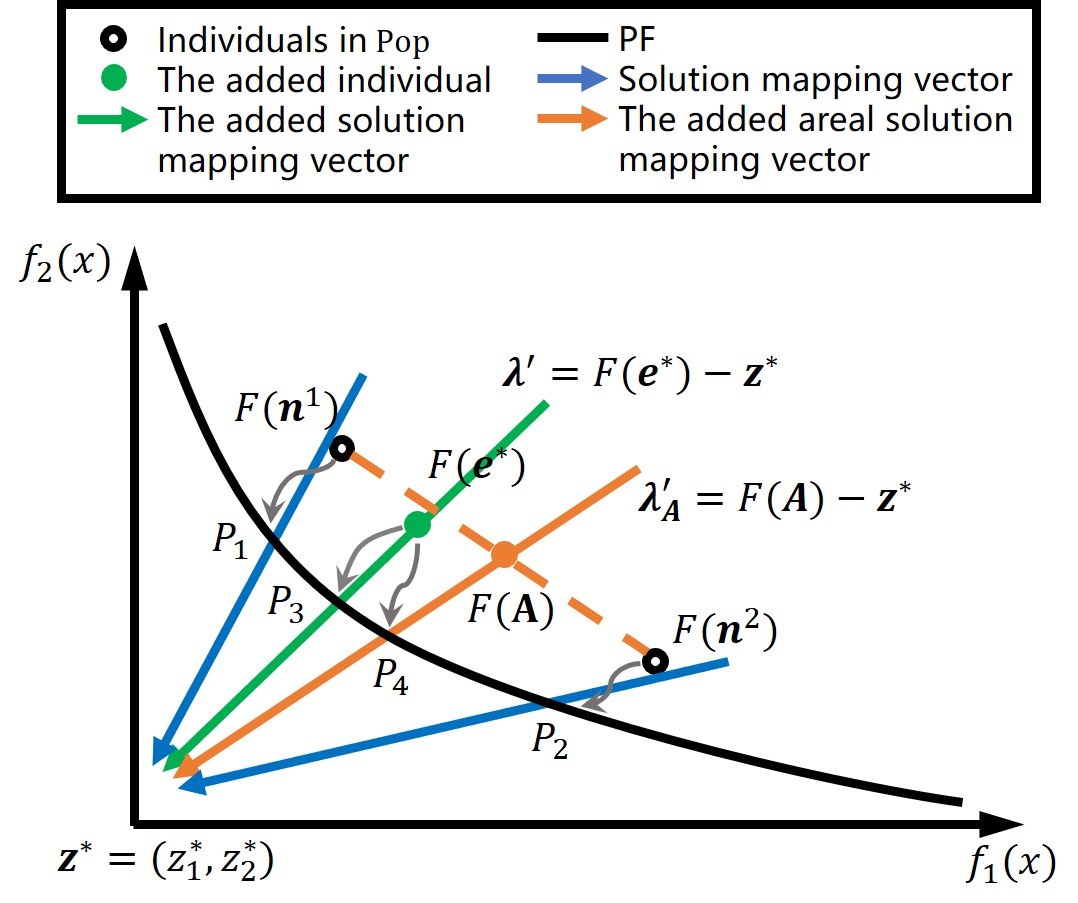}
	  \caption{Comparison of two subproblem adding strategies in terms of the added individuals' evolution}
   \label{fig5}
\end{figure}

As shown in Fig. \ref{fig3}(c), after electing the added individual $\boldsymbol{e}^{\ast}$, its two adjacent neighbors $\boldsymbol{x}^{1}$ and $\boldsymbol{x}^{3}$ in the $Pop$ form an area $A$ without any other individuals. The areal centroid $ F \left( A \right)$ is represented as the midpoint of the line segment composed of the two neighbors. $\bm{\lambda}_{A}$ points from $ F \left( A \right)$ to the reference point, providing evolutionary direction for the selected individual. Finally, the new weight vector can be calculated according to WS-transformation.

To facilitate a great understanding of the method, we take a bi-objective minimization problem as an example as shown in Fig. \ref{fig5}. $\boldsymbol{n}^{1}$ and $\boldsymbol{n}^{2}$ are the two-nearest neighbors of the added individual $\boldsymbol{e}^{\ast}$. The midpoint of the line segment composed of points $F \left( \boldsymbol{n}^{1} \right)$ and $F \left( \boldsymbol{n}^{2} \right)$ is the centroid $F \left( A \right)$. $\bm{\lambda}_{A}^{'}$ is the solution mapping vector by our areal subproblem adding strategy, guiding the added individual $\boldsymbol{e}^{\ast}$ to evolve towards $P_{4}$ on the PF. And the solution mapping vector $\bm{\lambda}^{'}$ constructed by the objective function value of the added individual $\boldsymbol{e}^{\ast}$ guides $\boldsymbol{e}^{\ast}$ to evolve towards $P_{3}$. The uniformity and diversity of solutions $\{P_{1},P_{2},P_{4}\}$ on the PF is better than that of solutions $\{P_{1},P_{2},P_{3}\}$. The reason is that the centroid of the area formed by neighbors is closer to the sparsest area of the current population, so that the constructed areal weight vector or its solution mapping vector can more accurately guide the evolution of the added individual to the sparse area on the PF. In addition, the areal weight vector is constructed by the centroids of sparse area. Regardless of where the newly-added individual $\boldsymbol{e}^{\ast}$ is located in the area, the evolutionary direction provided by the areal weight vector navigates to the sparsest part of the area. The pseudocode of the adaptive areal weight adjustment strategy is shown in Algorithm \ref{alg1}.

\subsection{MOEA/D-AAWA}\label{cha4.4}

Based on the above analysis, the pseudocode of the MOEA/D-AAWA is shown in Algorithm \ref{alg2} and the main steps are as follows.

\textit{Step 1 (Initialization):} initialize $Pop$, $EP$, neighborhood list $B$ and related parameters (line 1, lines 2-5). Generate a set of uniformly distributed solution mapping vectors. Obtain initial weight vectors according to WS-transformation (line 2).

\textit{Step 2 (Allocate computing resources):} the pseudocode of the computing resources allocation strategy is shown in Algorithm \ref{alg3} \cite{zhang2009performance}. A utility function $U$ is defined and computed for each subproblem. The indexes of evolutionary individuals $I$ are selected from $Pop$ according its utility function (line 8).
\begin{algorithm}
\caption{MOEA/D-AAWA} 
\label{alg2}
\begin{algorithmic}[1]
\REQUIRE the size of current population $N$; objective function dimension $m$; the neighborhood size $T$; the probability of selecting mate subproblem from its neighborhood $\delta$; the ratio of iteration times to evolve with only MOEA/D $r_{evol}$; individual evolution ratio $r_{pop}$; update number $n_{r}$; the maximal number of subproblems needed to be adjusted $nus$; the iteration interval of utilizing computing resource allocation strategy $g_{r}$; the iteration interval of utilizing the adaptive areal weight vector adjustment strategy $g_{w}$
\ENSURE population $Pop$ and its objective function values

\STATE Initialize population randomly $Pop = \{\boldsymbol{x^{i}}, i = 1,\ldots,N\}$; for each $\boldsymbol{x^{i}}$, evaluate $F\left( \boldsymbol{x^{i}} \right) = \left( f_{1}\left( \boldsymbol{x}^{i} \right), \ldots, f_{m}\left( \boldsymbol{x}^{i} \right) \right)^{T}$;
\STATE Initialize the weight vectors $\{\bm{\lambda}^{i},i = 1,\ldots,N\}$ by applying the WS-transformation on uniformly distributed solution mapping vectors;
\STATE Calculate the Euclidean distances between any two weight vectors. For each $i = 1,\ldots,N$, set the neighborhood list of the $i$-th subproblem as $B(i) = \left( i_{1},i_{2},\ldots,i_{T} \right)^{T}$, where $i_{1},i_{2},\ldots,i_{T}$ are the indexes of $T$ closest weight vectors to $\bm{\lambda}^{i}$;
\STATE Initialize the reference $\boldsymbol{z}^{\ast} = \left( z_{1}^{\ast}, \ldots, z_{m}^{\ast} \right)^{T}$ by setting $z_{i}^{\ast} = {\mathit{\min}\left\{ {f_{i}\left( \boldsymbol{x}^{1} \right),\ldots,f_{i}\left( \boldsymbol{x}^{N} \right)} \right\}} - 10^{- 7}$;
\STATE Set $g=1$, $EP=\emptyset$ and $U^{i} = 1, i = 1, \ldots, N$;

\WHILE{termination criteria is not satisfied}
    \IF{mod($g$,$g_{r}$)==0}
        \STATE Allocate computing resources by Algorithm \ref{alg3} and obtain indexes list of evolutionary individuals $I$;
    \ENDIF
    
    \FOR{$i \in I$}
        \STATE Select mating scope: $P = \left\{\begin{aligned} &B(i), rand < \delta \\ &\left\{ {1,2,\ldots,N} \right\},otherwise \end{aligned}\right.$, where $rand$ is a uniformly random number from (0, 1);
        \STATE Reproduce: randomly select two indexes $r_{1}$ and $r_{2}$ from $P$, and then generate a new individual $\boldsymbol{y}$ from $\boldsymbol{x}^{r_{1}}$ and $\boldsymbol{x}^{r_{2}}$ by the SBX operator and the polynomial mutation operator;
        \STATE Repair $\boldsymbol{y}$: if an element of $\boldsymbol{y}$ is out of the boundary, randomly select its value inside the boundary;
        \STATE Update the reference $\boldsymbol{z}^{\ast}$: for each $j = 1, \ldots, m$, if $z_{j}^{\ast} > f_{j}\left( \boldsymbol{y} \right)$, then set $z_{j}^{\ast} = f_{j}\left( \boldsymbol{y} \right) - 10^{-7}$;
        \STATE Update the population $Pop$ by Algorithm \ref{alg4};
    \ENDFOR

    \IF{$g \geq r_{evol} \times {g_{max}}$}
        \STATE Update the external population $EP$ by the new generation of offspring according to the vicinity distance based nondominated sorting;
        \IF{mod($g$,$g_{r}$)==0}
            \STATE Adjust weight vectors by Algorithm \ref{alg1};
            \STATE Update neighborhood: for each $i = 1,\ldots,N$, find the $T$ closest weights to $\bm{\lambda}^{i}$ and build the new $B(i)$.
        \ENDIF
    \ENDIF
    \STATE Set $g = g + 1$.
\ENDWHILE

\end{algorithmic}
\end{algorithm}

\textit{Step 3 (Update):} for each evolutionary individual in $I$, its offspring $\boldsymbol{y}$ is generated by the SBX operator and the polynomial mutation operator \cite{deb2001self} shown as Eq.(\ref{eqn24})-(\ref{eqn27}) (lines 11-13). Update the reference $\boldsymbol{z}^{\ast}$ (line 14). Update the $Pop$ by constraint-domination principle (CDP) shown in Algorithm \ref{alg4} \cite{deb2002fast} (line 15). The definition of constraint-domination is given as follows. A solution $\boldsymbol{x}$ is said to constraint-dominate another solution $\boldsymbol{y}$, if any of the following criterion is satisfied: 1) $\boldsymbol{x}$ is feasible and $\boldsymbol{y}$ is infeasible; 2) Both $\boldsymbol{x}$ and $\boldsymbol{y}$ are infeasible and $\boldsymbol{x}$ has a smaller degree of constraint violation than $\boldsymbol{y}$; 3) Both $\boldsymbol{x}$ and $\boldsymbol{y}$ are feasible and $\boldsymbol{x}$ dominates $\boldsymbol{y}$.
\begin{align}
\label{eqn24} \boldsymbol{y}_{k}^{'} = 0.5 \times \left\lbrack \left( {1 + \beta} \right) \boldsymbol{x}^{r_{1}}_{k} + \left( {1 - \beta} \right) \boldsymbol{x}^{r_{2}}_{k} \right\rbrack,
\end{align}
\begin{align}
\label{eqn25} \beta = 
\left\{\begin{aligned}
&(rand \times 2)^{1/(1 + \eta_{c})},rand \leq 0.5, \\
&\left\lbrack 1/(2 - rand \times 2) \right\rbrack^{1/(1 + \eta_{c})},otherwise,
\end{aligned}\right.
\end{align}
\begin{align}
\label{eqn26} \boldsymbol{y}_{k}= \boldsymbol{y}_{k}^{'} + \delta_{k},
\end{align}
\begin{align}
\label{eqn27} \textstyle\delta_{k} =
\left\{\begin{aligned}
&(2 \times rand)^{1/(\eta_{m} + 1)} - 1,if~rand \leq 0.5, \\
&1 - \left\lbrack 2\left( {1 - rand} \right) \right\rbrack^{1/(\eta_{m} + 1)},otherwise,
\end{aligned}\right.
\end{align}
where $rand$ is a uniformly random number from (0, 1); $\eta_{c}$ and $\eta_{m}$ are the distribution index of the SBX operator and the polynomial mutation operator.
\begin{algorithm}
\caption{Allocation of computing resources} 
\label{alg3}
\begin{algorithmic}[1]
\REQUIRE The objective function values in the last allocation of computing resources $f_{old}$; Current objective function values $f$; Individual evolution ratio $r_{pop}$
\ENSURE $I$

\STATE Calculate the increment of the utility function: $\Delta^{i} = \frac{f_{old}\left( \boldsymbol{x}^{i} \right) - f\left( \boldsymbol{x}^{i} \right)}{f_{old}\left( \boldsymbol{x}^{i} \right)}$;
\STATE Update the utility function: $U^{i} = \left\{\begin{aligned} &1,\Delta^{i} > 0.001 \\ &\left( {0.95 + 0.05*\frac{\Delta^{i}}{0.001}} \right)U^{i},otherwise \end{aligned}\right.$
\STATE Select the subproblems: set $I=\emptyset$ and select all indexes of the $m$ boundary weight vectors. Choose other $\lfloor N \times r_{pop} \rfloor - m$ indexes using 2-tournament selection \cite{miller1995genetic} according to $U^{i}$, and add them to $I$.

\end{algorithmic}
\end{algorithm}

\begin{algorithm}
\caption{Update scheme of population} 
\label{alg4}
\begin{algorithmic}[1]
\STATE Set $c=0$;
\WHILE{$c < n_{r}$ or $P \neq \emptyset$}
    \STATE Randomly pick an index $j$ from $P$;
    \IF{$V(\boldsymbol{y})  < V\left( \boldsymbol{x}^{j} \right)$}
        \STATE $\boldsymbol{x}^{j} = \boldsymbol{y},f\left( \boldsymbol{x}^{j} \right) = f(\boldsymbol{y}),V\left( \boldsymbol{x}^{j} \right) = V(\boldsymbol{y}),c = c + 1$
    \ELSIF{$V(\boldsymbol{y}) = V\left( \boldsymbol{x}^{j} \right) = 0$}
        \IF{$g^{te}\left( \boldsymbol{y} \middle| {\bm{\lambda}^{j},\boldsymbol{z}^{\ast}} \right) < g^{te}\left( \boldsymbol{x}^{j} \middle| \bm{\lambda}^{j},\boldsymbol{z}^{\ast} \right)$}
            \STATE $\boldsymbol{x}^{j} = \boldsymbol{y},f\left( \boldsymbol{x}^{j} \right) = f(\boldsymbol{y}),V\left( \boldsymbol{x}^{j} \right) = V(\boldsymbol{y}),c = c + 1$
        \ENDIF
    \ENDIF
\ENDWHILE
\end{algorithmic}
\end{algorithm}

\textit{Step 4 (Adaptive areal weight adjustment):} update $EP$ by the new generation of offspring according to the vicinity distance based nondominated sorting (line 18). Periodically delete crowded subproblems and add new subproblems according to Algorithm \ref{alg1} (line 20). Update $B$ after adjusting Weight vectors (line 21).

\textit{Step 5 (Stop criteria):} go to Step 2 and repeat the above steps until termination criteria is satisfied (lines 6-25).

\section{Computational experiments}\label{cha5}

In this section, we evaluate the performance of MOEA/D-AAWA in synthetic (mountain and urban environments) and realistic scenarios. The proposed algorithm and the three compared algorithms are coded in MATLAB, and run on a PC computer with Core i5-10400F 2.90GHz CPU, 16G memory, and Windows 10 operating system.

\subsection{Experimental setting}\label{cha5.1}

Twenty-four experimental cases are designed in this paper. The first environment is mountain environments. Ten cases (C1-C10) are generated under the mountain environment in the range of 200 km × 200 km, whose numbers of mountain obstacles are 2-20 with a step of 2. 

The mountain environment consists of original terrain and mountain obstacles. The elevation matrix of original terrain is constructed by Eq.(\ref{eqn28}) \cite{jiang2016towards}.
\begin{align}
\label{eqn28} z_{1}\left( x,y \right) = {\mathit{\sin}}\left( y + a \right) + b{\mathit{\sin}}\left( x \right) + c{\mathit{\cos}}\left( y \right) + d{\mathit{\cos}}\left( e\sqrt{x^{2} + y^{2}} \right) + f{\mathit{\sin}}\left( f\sqrt{x^{2} + y^{2}} \right),
\end{align}
where $(x,y)$ is the coordinate of a point on the horizontal plane. $z_{1}(x,y)$ is the corresponding terrain height. $a$, $b$, $c$, $d$, $e$, $f$ are the terrain coefficients, and a combination of them can simulate different topographic features. In this paper, the terrain coefficients are set to $a=3\pi$, $b=0.1$, $c=0.3$, $d=0.9$, $e=0.5$ and $f=0.5$.

The elevation matrix of 3-D mountain obstacles is constructed by Eq.(\ref{eqn29}).
\begin{align}
\label{eqn29} z_{2}(x,y) = {\sum_{i = 1}^{k}{h(i) * \exp\left( - \frac{x - x_{0}(i)}{a(i)} - \frac{y - y_{0}(i)}{b(i)} \right)}},
\end{align}
where $z_{2}(x,y)$ is the corresponding mountain height. $k$ is the number of mountain obstacles. $h(i)$ is the height of $i$-th mountain obstacle. $(x_{0}(i),y_{0}(i))$ is the coordinate of the center of the $i$-th mountain obstacle on the horizontal plane. $a(i)$ and $b(i)$ are the corresponding slope parameters along the $x$ axis and $y$ axis directions. By adjusting $h(i)$, $a(i)$ and $b(i)$, mountain obstacles with various contours can be constructed.

Then, the elevation matrix of mountain environments can be formed by merging the elevation matrix of original terrain and mountain obstacles according to Eq.(\ref{eqn30}).
\begin{align}
\label{eqn30} z_{m}(x,y) = \max\left\{ z_{1}(x,y),z_{2}(x,y) \right\}.
\end{align}

The second environment is urban environments. Ten cases (C11-C20) are generated under the urban environment in the range of 2000 m × 2000 m, whose numbers of urban obstacles are 2-20 with a step of 2. 

We model 3-D urban obstacles as prisms. To further keep the safety of the UAV during the flight, we set the protected zone for each obstacle, whose width is define as a safe distance $r_{safe}$. The relevant elevation matrix of urban environments can be calculated as follows. For each urban obstacle, if $(x, y)$ is within $k$-th urban obstacle, set $z_{u}^{k}\left( {x,y} \right) = h_{k}$, otherwise 0. $h_{k}$ is the height of $i$-th urban obstacle and $z_{u}^{k}(x,y)$ is elevation matrix of $k$-th urban obstacle. And then the elevation matrix of urban environments can be calculated as $z_{u}\left( {x,y} \right) = \max\limits_{k}\left\{ z_{u}^{k}\left( {x,y} \right) \right\}$. 

To make the experiment more convincing, four cases (C21-C24) are generated under the realistic scenarios in the range of 10 km × 10 km, and the spatial resolution is 5 m. The latitude and longitude ranges of the four cases are shown in Table \ref{tbl0}. 
\begin{table}[pos=htb, width=0.7\textwidth]
    \caption{The latitude and longitude ranges of C21-C24}
    \label{tbl0}
    \begin{tabular*}{\tblwidth}{@{}LL@{}}
        \toprule
        case & the range of longitude and latitude \\
        \midrule
        C21 & (78.680000$^{\circ}$E-78.608046$^{\circ}$E, 34.300000$^{\circ}$N-34.390212$^{\circ}$N) \\
        C22 & (78.500000$^{\circ}$E-78.608046$^{\circ}$E, 33.670000$^{\circ}$N-33.760212$^{\circ}$N) \\
        C23 & (78.410000$^{\circ}$E-78.518135$^{\circ}$E, 33.760000$^{\circ}$N-33.850212$^{\circ}$N) \\
        C24 & (78.190000$^{\circ}$E-78.297817$^{\circ}$E, 33.500000$^{\circ}$N-33.590212$^{\circ}$N) \\
        \bottomrule
    \end{tabular*}
\end{table}

NSGA-III \cite{jain2013evolutionary}, C-MOEA/D \cite{jain2013evolutionary} and MOEA/D-AWA \cite{qi2014moea} are used for the comparison. The parameters of this paper are set as follows:

1) Each algorithm runs over 30 independent times, and stops after $k$ function evaluations. The times of function evaluation $k$ and number of path control points (excluding starting and target points) $DOP$ in all cases are given in Table \ref{tbl1}.

2) The safe distance $r_{safe}$ is set to 200 m, 10 m and 20 m in mountain environments (C1-C10), urban environments (C11-C20) and realistic scenarios (C21-C24) respectively;

3) The population size $N$ is set to 20, the neighborhood size $T$ is set to $0.1 \times N$;

4) The proposed MOEA/D-AAWA has the same parameters as MOEA/D-AWA: $\eta_{c}=20$, $\eta_{m}=1$, $n_{r}=2$, $r_{evol}=0.8$, $r_{pop}=0.2$, $nus=0.1 \times N$, $g_{r}=50$, $g_{w}=100$;

5) The rest of the parameters in NSGA-III and C-MOEA/D remain the same as in the original papers.
\begin{table}[pos=htb, width=0.7\textwidth]
    \caption{The values of times of function evaluation $k$ and numbers of path control points}
    \label{tbl1}
    \begin{tabular*}{\tblwidth}{@{}LLL@{}}
        \toprule
        case & $k$ & $DOP$ \\
        \midrule
        C1-C3,C11-C13 & 10000 & 2 \\
        C4-C6,C14-C16 & 12000 & 3 \\
        C7-C9,C17-C19 & 14000 & 4 \\
        C10,C20       & 16000 & 5 \\
        C21-C24       & 10000 & 4 \\
        \bottomrule
    \end{tabular*}
\end{table}

\subsection{Performance metrics}\label{cha5.3}

Due to the unknown PFs of path planning problems, hypervolume (HV) \cite{shang2020new} and pure diversity (PD) \cite{wang2016diversity} are used to evaluate the performance of the four algorithms in this paper, as shown in Eq.(\ref{eqn33})-(\ref{eqn35}). In addition, the larger HV and PD implies the better convergence and diversity of PFs.
\begin{align}
\label{eqn33} {\rm HV} = {\rm hypervolume}\left( {\bigcup\limits_{i = 1}^{|{PF}|}s_{i}} \right),
\end{align}
\begin{align}
\label{eqn34} {\rm PD(X)} &= \max{\left\{ {\rm PD} \left( {X - s_{i}} \right) + d\left( s_{i},X - s_{i} \right) \right\}},
\end{align}
\begin{align}
\label{eqn35} d\left( {s,X} \right) &= {\min\limits_{s_{i} \in X}\left\{ {\rm dissimilarity}\left( s,s_{i} \right) \right\}},
\end{align}
where $s_{i}$ is the solution in the Pareto optimal solution set. $d(s,X)$ represents the dissimilarity between the solution  $s$ and a community $X$. When calculating the HV, first normalize the values of multiple objective functions, and then calculate the area of the polygon formed by the values and the reference point (1,1). Moreover, when there are no feasible solutions in the population, set HV and PD to 0. 

\subsection{Results and analyses}\label{cha5.4}

\subsubsection{Experiments under mountain environments}\label{cha5.4.1}

\begin{table}[pos=htb, width=\textwidth]
    \caption{Mean and Std values of HV and PD in mountain environments}
    \label{tbl2}
    \begin{tabular*}{\tblwidth}{@{}LLLLLLL@{}}
        \toprule
        \multirow{2}{*}[-2pt]{Case}&\multirow{2}{*}[-2pt]{\makecell{Number \\of threats}} &\multirow{2}{*}[-2pt]{Metric}&\multicolumn{4}{c}{Algorithm}\\
		\cmidrule(lr){4-7}
		&&&NSGA-III&C-MOEA/D&MOEA/D-AWA&MOEA/D-AAWA\\\midrule
  
        \multirow{2}{*}[-2pt]{C1}&\multirow{2}{*}[-2pt]{2}
        &HV &0.9799±0.0039 (3) &0.7968±0.0226 (4)&0.9799±0.0091 (2) &\textbf{0.9828}±0.004 \textbf{(1)}\\
        &&PD($\times 10^{4}$) &1.5202±0.1658 (3)&0.5881±0.144 (4)&2.2678±0.7973 (2)&\textbf{2.4318}±0.7354 \textbf{(1)} \\
        \multirow{2}{*}[-2pt]{C2}&\multirow{2}{*}[-2pt]{4}
        &HV &0.9663±0.0154 (3) &0.7895±0.0297 (4) &0.9723±0.0137 (2) &\textbf{0.9763}±0.0092 \textbf{(1)}\\
        &&PD($\times 10^{4}$) &1.4563±0.336 (3)&0.5424±0.1416 (4)&2.1858±1.024 (2)&\textbf{2.1922}±0.6543 \textbf{(1)}\\
        \multirow{2}{*}[-2pt]{C3}&\multirow{2}{*}[-2pt]{6}
        &HV &0.9663±0.0071 (2) &0.7919±0.0512 (4) &\textbf{0.9694}±0.0139 \textbf{(1)} &0.9656±0.0165 (3)\\
        &&PD($\times 10^{4}$) &1.8452±0.3764 (3)&0.4038±0.1045 (4)&\textbf{2.6521}±1.3831 \textbf{(1)}&2.4123±1.0745 (2)\\
        \multirow{2}{*}[-2pt]{C4}&\multirow{2}{*}[-2pt]{8}
        &HV &0.9729±0.0161 (3) &0.7696±0.042 (4) &0.977±0.0069 (2) &\textbf{0.9789}±0.0074 \textbf{(1)}\\
        &&PD($\times 10^{4}$) &1.5276±0.2884 (3)&0.5397±0.2363 (4)&2.2935±0.7193 (2)&\textbf{2.5778}±1.223 \textbf{(1)}\\
        \multirow{2}{*}[-2pt]{C5}&\multirow{2}{*}[-2pt]{10}
        &HV &0.94±0.1782 (3) &0.7779±0.0578 (4) &0.9749±0.0119 (2) &\textbf{0.9793}±0.0068 \textbf{(1)}\\
        &&PD($\times 10^{4}$) &1.6526±0.4933 (3)&0.5266±0.2524 (4)&\textbf{2.8152}±0.8421 \textbf{(1)}&2.6343±0.6803 (2)\\
        \multirow{2}{*}[-2pt]{C6}&\multirow{2}{*}[-2pt]{12}
        &HV &0.9736±0.0087 (3) &0.7773±0.06 (4) &0.9774±0.0086 (2) &\textbf{0.9791}±0.0063 \textbf{(1)}\\
        &&PD($\times 10^{4}$) &1.7189±0.4173 (3)&0.5836±0.2936 (4)&\textbf{2.5545}±0.7485 \textbf{(1)}&2.4437±0.7408 (2)\\
        \multirow{2}{*}[-2pt]{C7}&\multirow{2}{*}[-2pt]{14}
        &HV &0.7768±0.3957 (3) &0.7384±0.0438 (4) &\textbf{0.9799}±0.0073 \textbf{(1)} &0.976±0.0259 (2)\\
        &&PD($\times 10^{4}$) &1.3276±0.7345 (3)&0.5986±0.2677 (4)&2.7111±0.8724 (2)&\textbf{2.7573}±1.3857 \textbf{(1)}\\
        \multirow{2}{*}[-2pt]{C8}&\multirow{2}{*}[-2pt]{16}
        &HV &0.7737±0.3937 (3) &0.7045±0.0978 (4) &\textbf{0.9715}±0.012 \textbf{(1)} &0.9664±0.0227 (2)\\
        &&PD($\times 10^{4}$) &1.6206±0.9349 (3)&0.8282±0.3471 (4)&2.9873±1.0119 (2)&\textbf{3.3547}±1.4623 \textbf{(1)}\\
        \multirow{2}{*}[-2pt]{C9}&\multirow{2}{*}[-2pt]{18}
        &HV &0.7918±0.3625 (3) &0.6986±0.1362 (4) &0.9593±0.0308 (2) &\textbf{0.9629}±0.0299 \textbf{(1)}\\
        &&PD($\times 10^{4}$) &1.6448±0.8734 (3)&1.0799±0.5171 (4)&3.1508±1.2755 (2)&\textbf{3.3412}±0.9261 \textbf{(1)}\\
        \multirow{2}{*}[-2pt]{C10}&\multirow{2}{*}[-2pt]{20}
        &HV &0.4291±0.4698 (4) &0.6873±0.1092 (3) &0.8468±0.2903 (2) &\textbf{0.9423}±0.0598 \textbf{(1)}\\
        &&PD($\times 10^{4}$) &1.0252±1.4117 (3)&1.0026±0.5005 (4)&2.6693±1.4752 (2)&\textbf{3.2183}±1.4812 \textbf{(1)}\\\midrule

        \multicolumn{3}{c}{Overall ranking}&(60)&(79)&(34)&\textbf{(27)}\\
        \bottomrule
    \end{tabular*}
\end{table}
\begin{figure}[pos=htb]
    \centering
        \subfigure[]{\includegraphics[width=4cm]{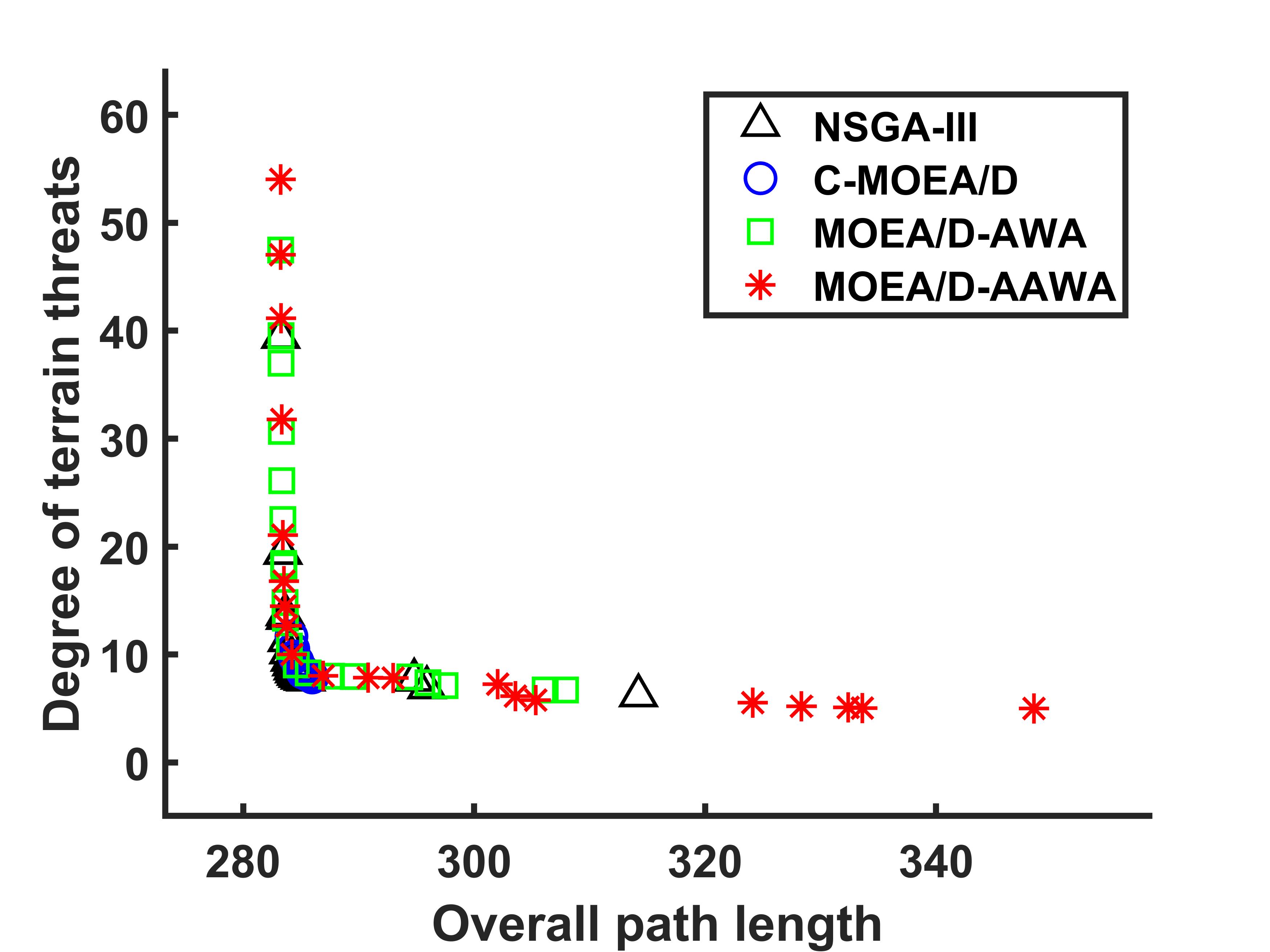}}
        \subfigure[]{\includegraphics[width=4cm]{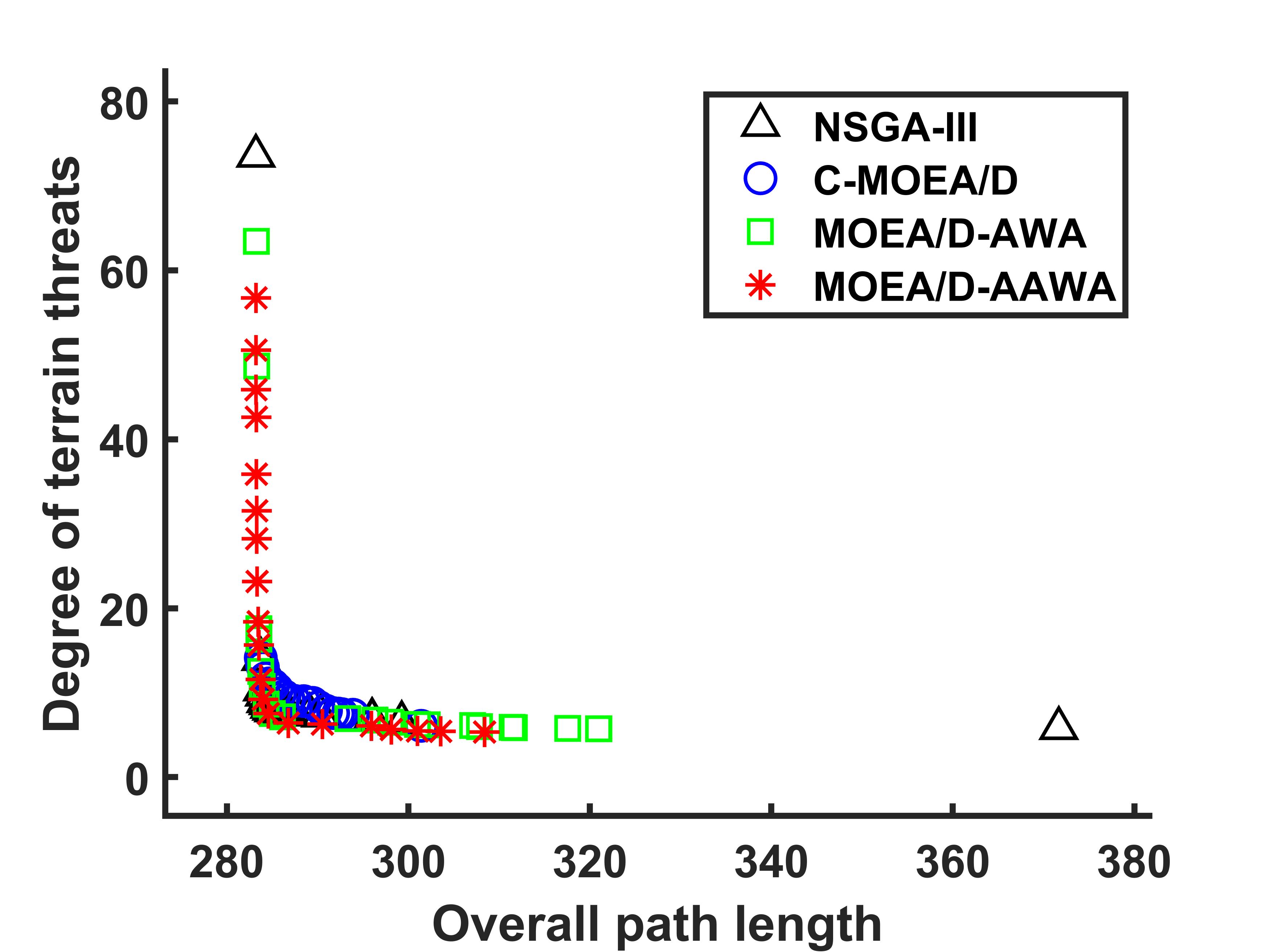}}
        \subfigure[]{\includegraphics[width=4cm]{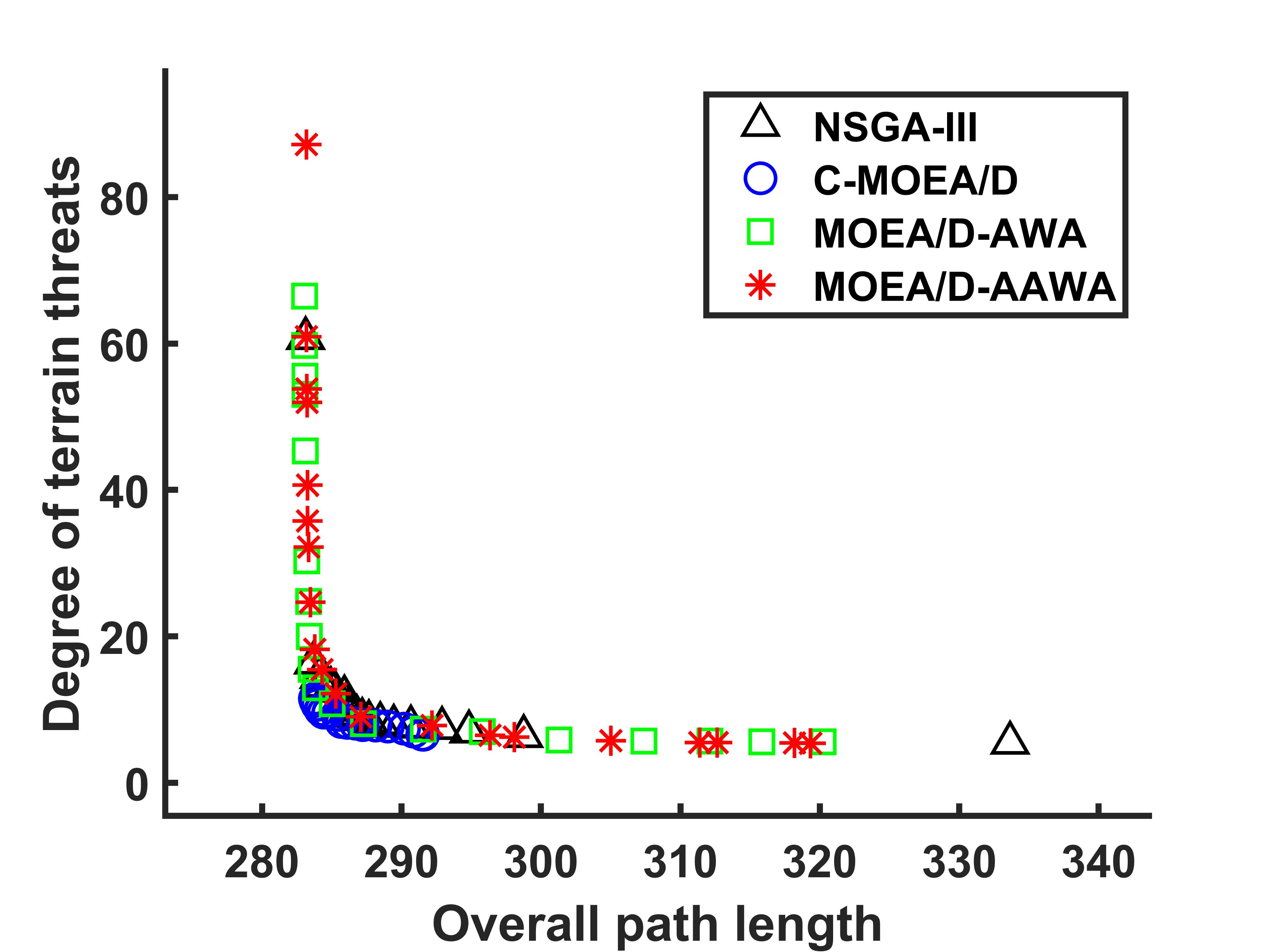}}
        \subfigure[]{\includegraphics[width=4cm]{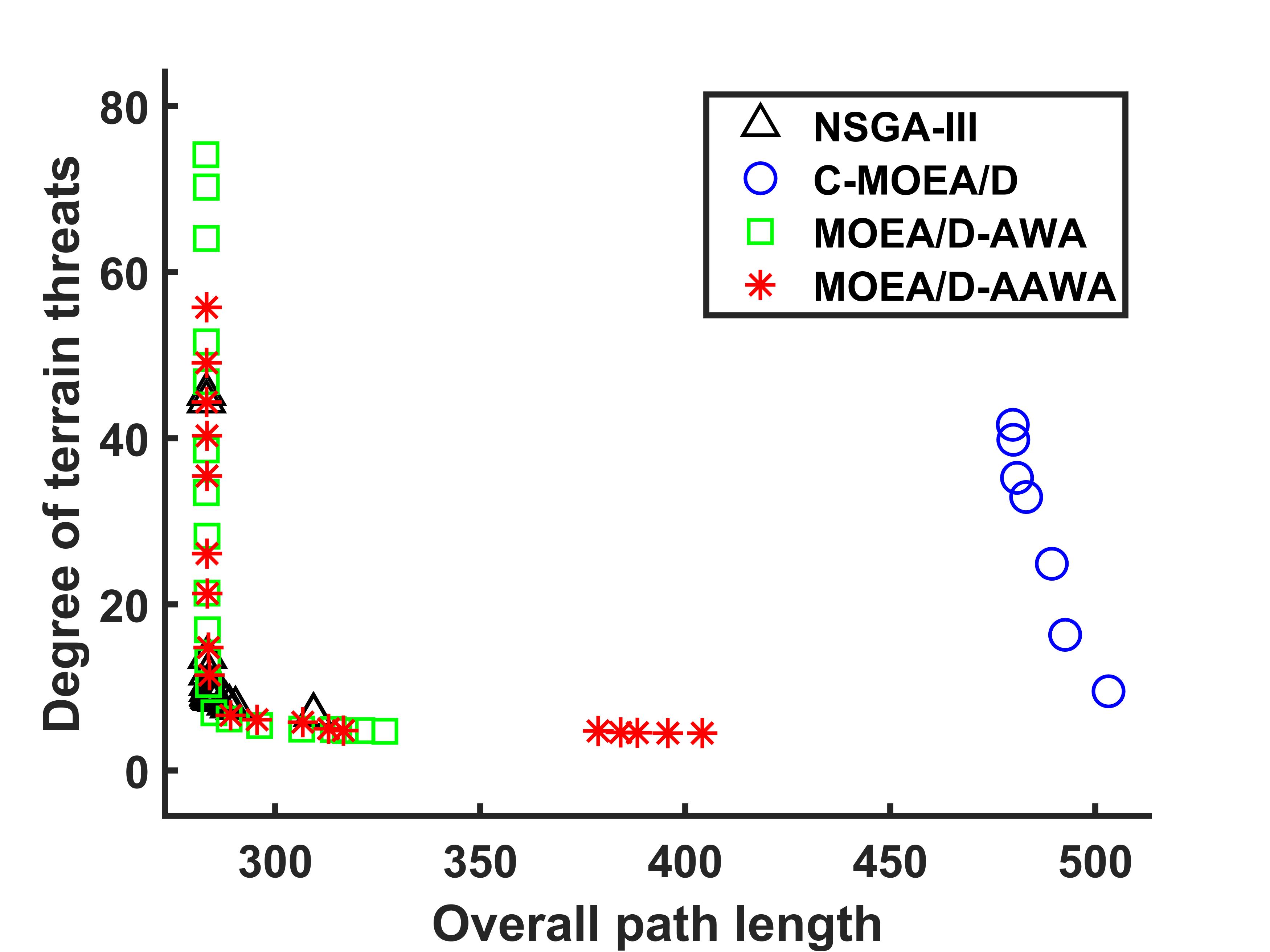}}
    \caption{Final nondominated solutions in mountain environments. (a): C3; (b): C6; (c): C9; (d): C10.}
    \label{fig6}
\end{figure}
\begin{figure}[pos=htb]
	\begin{minipage}{0.24\linewidth}
		\vspace{3pt}\centerline{\includegraphics[width=4.5cm]{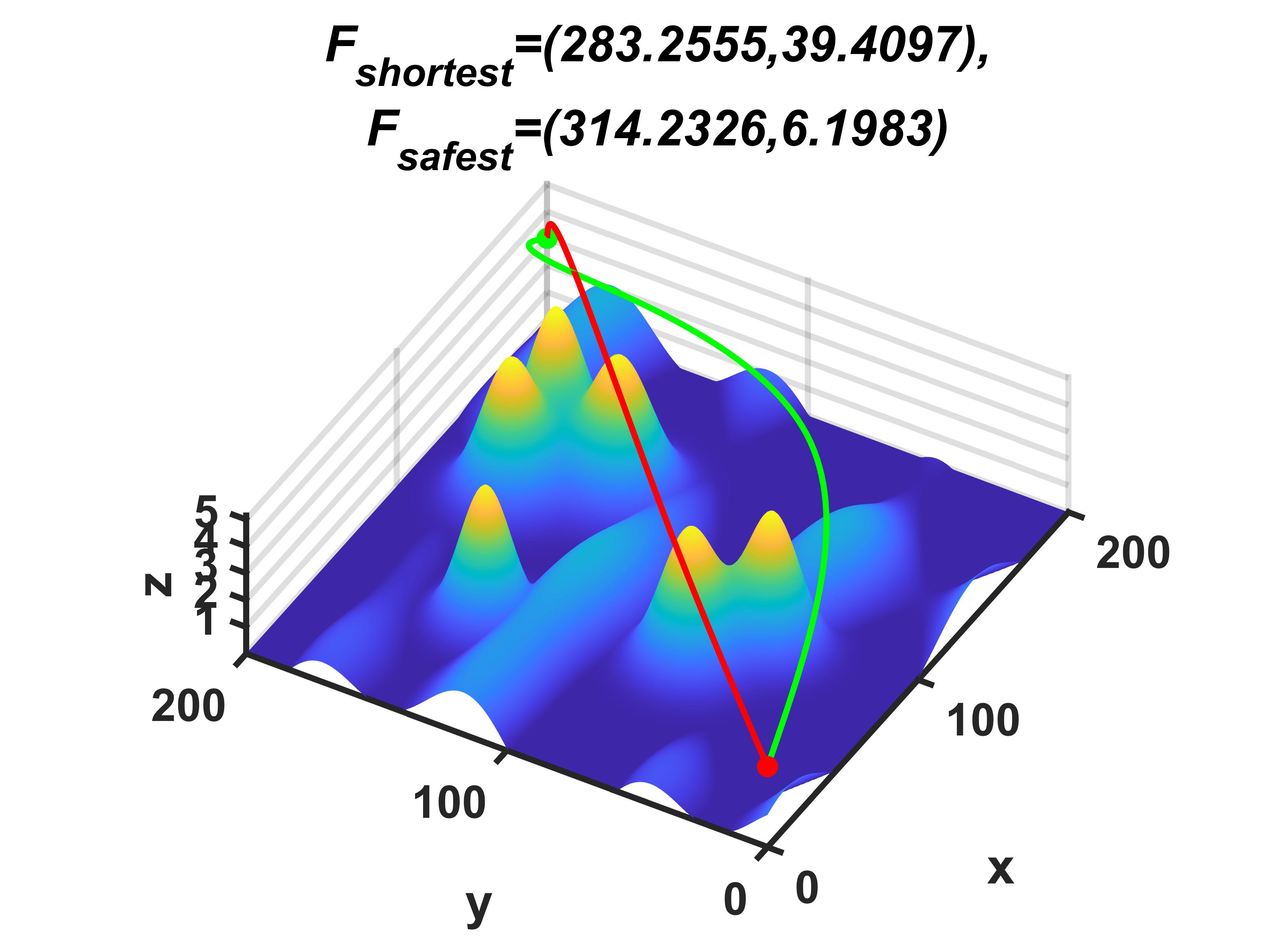}}\centerline{\footnotesize{(a1)}}
	\end{minipage}
    \begin{minipage}{0.24\linewidth}
		\vspace{3pt}\centerline{\includegraphics[width=4.5cm]{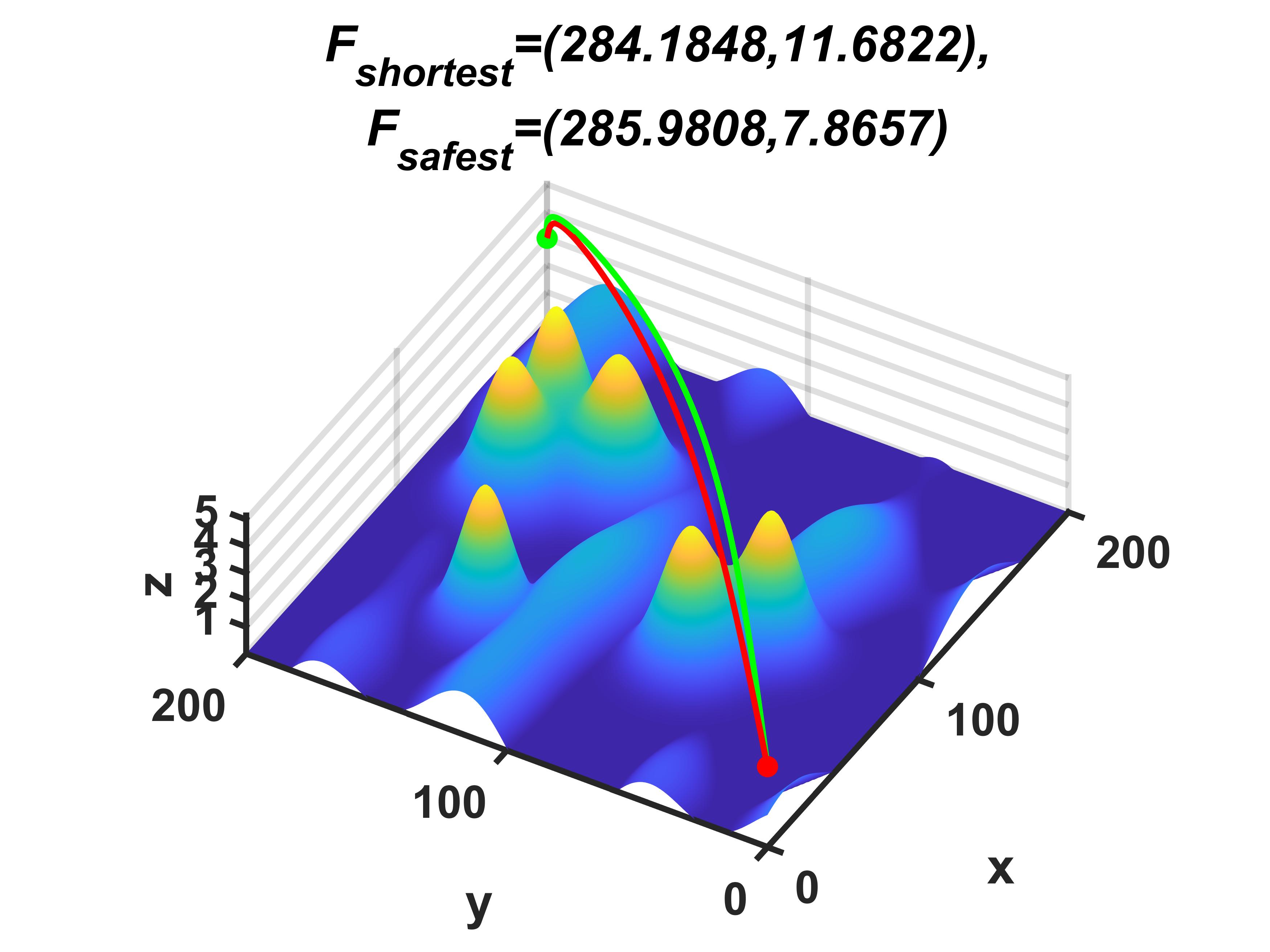}}\centerline{\footnotesize{(a2)}}
	\end{minipage}
    \begin{minipage}{0.24\linewidth}
		\vspace{3pt}\centerline{\includegraphics[width=4.5cm]{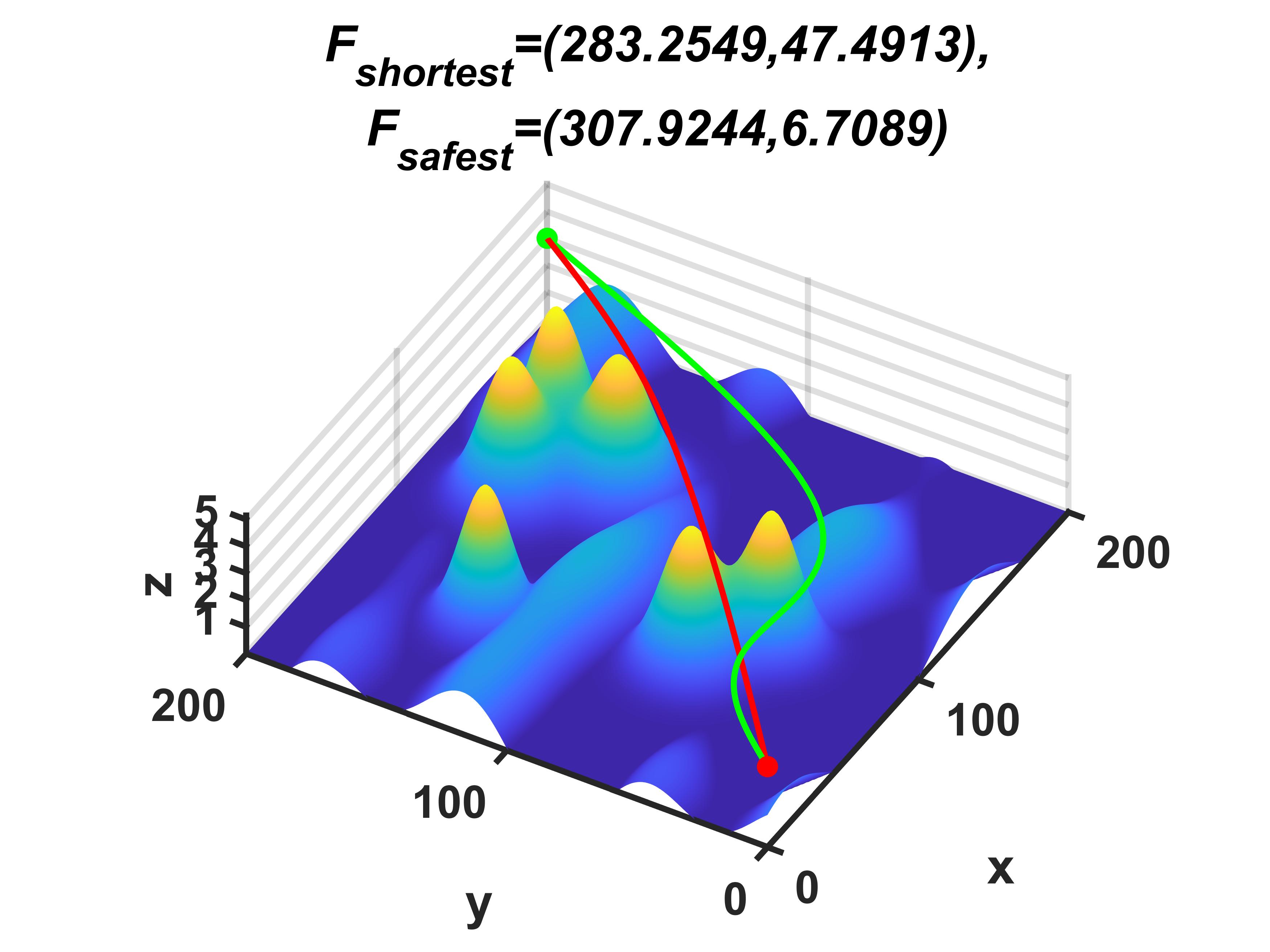}}\centerline{\footnotesize{(a3)}}
	\end{minipage}
    \begin{minipage}{0.24\linewidth}
		\vspace{3pt}\centerline{\includegraphics[width=4.5cm]{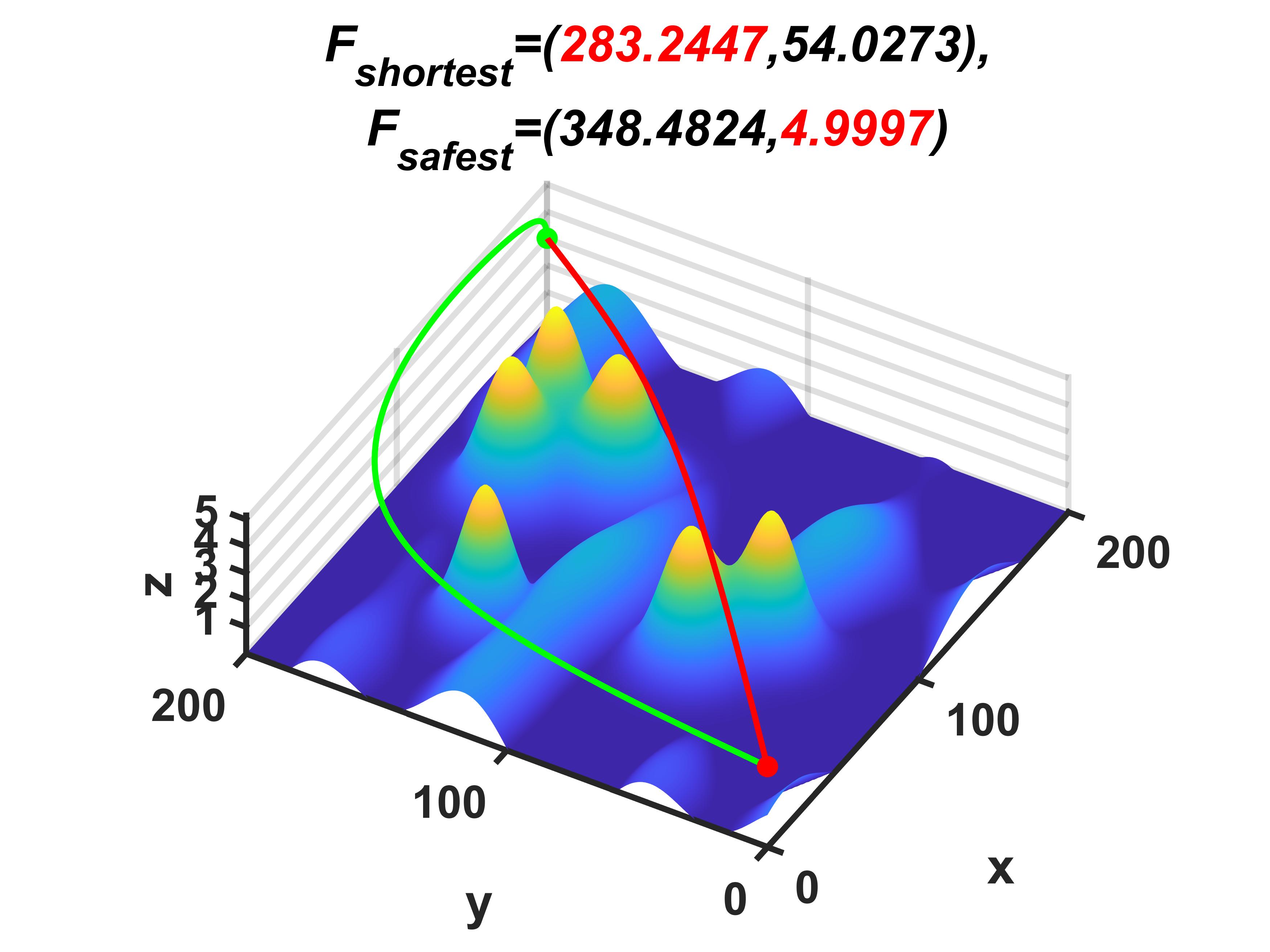}}\centerline{\footnotesize{(a4)}}
	\end{minipage}
 
    \begin{minipage}{0.24\linewidth}
		\vspace{3pt}\centerline{\includegraphics[width=4.5cm]{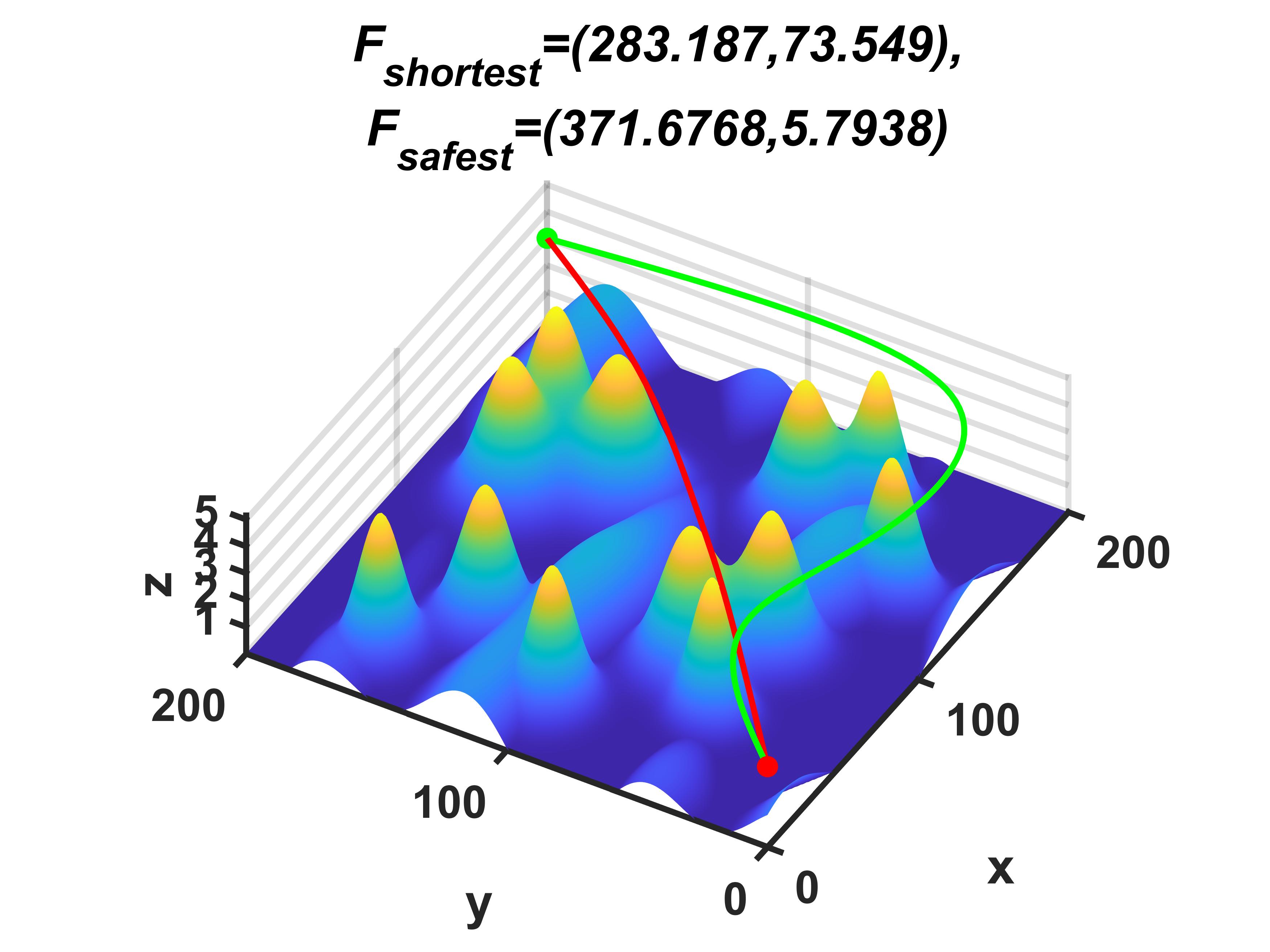}}\centerline{\footnotesize{(b1)}}
	\end{minipage}
     \begin{minipage}{0.24\linewidth}
		\vspace{3pt}\centerline{\includegraphics[width=4.5cm]{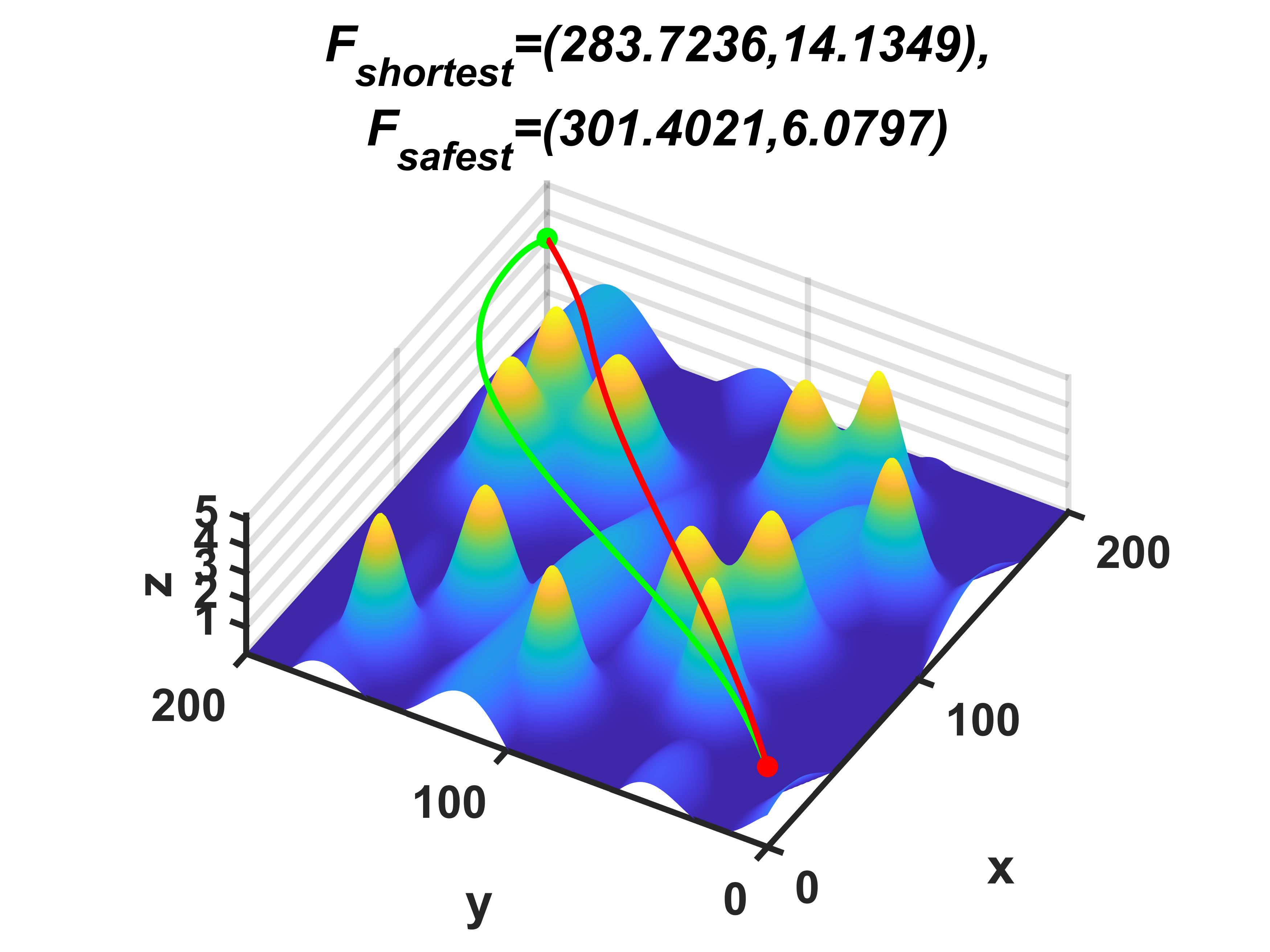}}\centerline{\footnotesize{(b2)}}
	\end{minipage}
     \begin{minipage}{0.24\linewidth}
		\vspace{3pt}\centerline{\includegraphics[width=4.5cm]{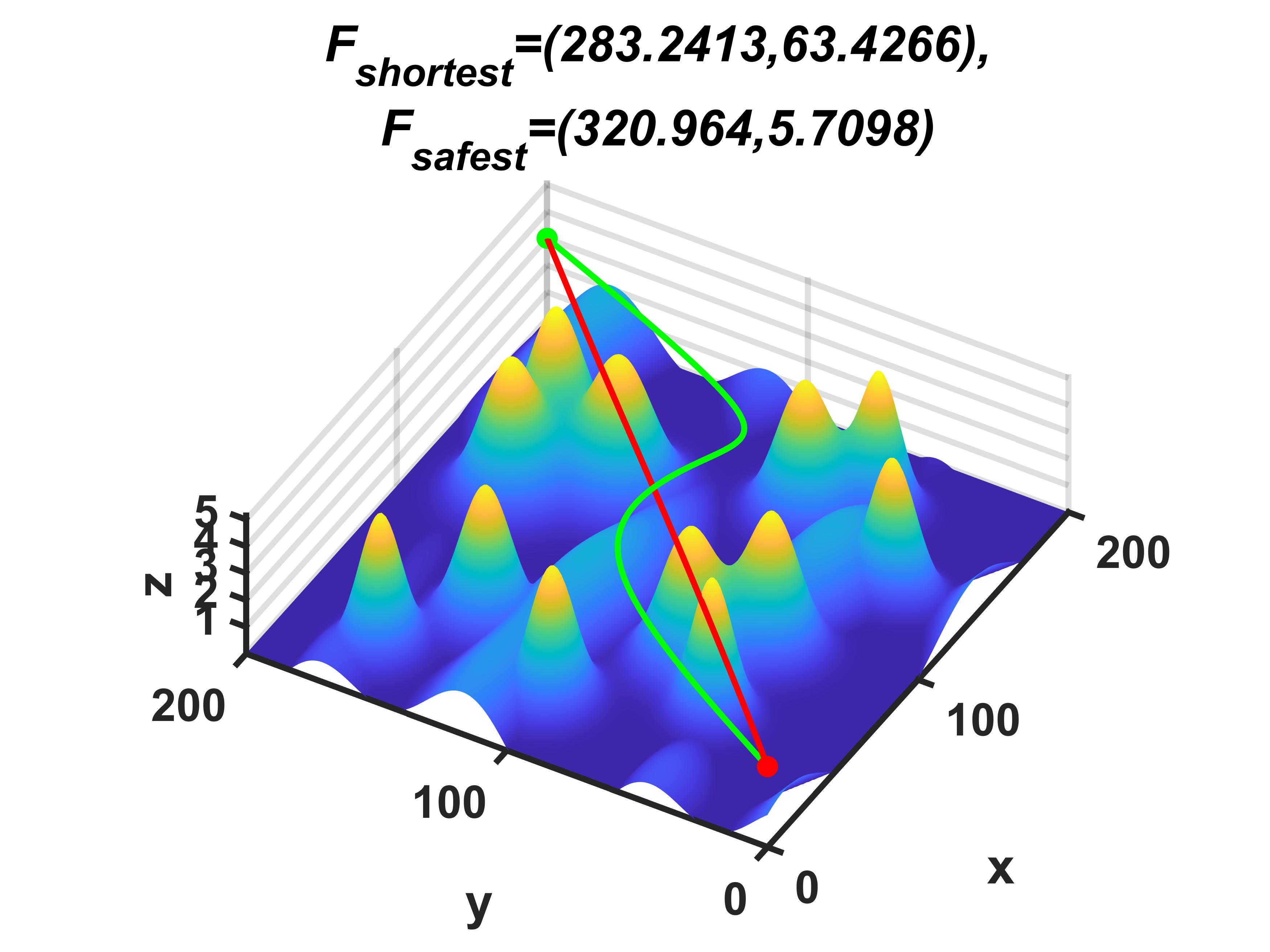}}\centerline{\footnotesize{(b3)}}
	\end{minipage}
     \begin{minipage}{0.24\linewidth}
		\vspace{3pt}\centerline{\includegraphics[width=4.5cm]{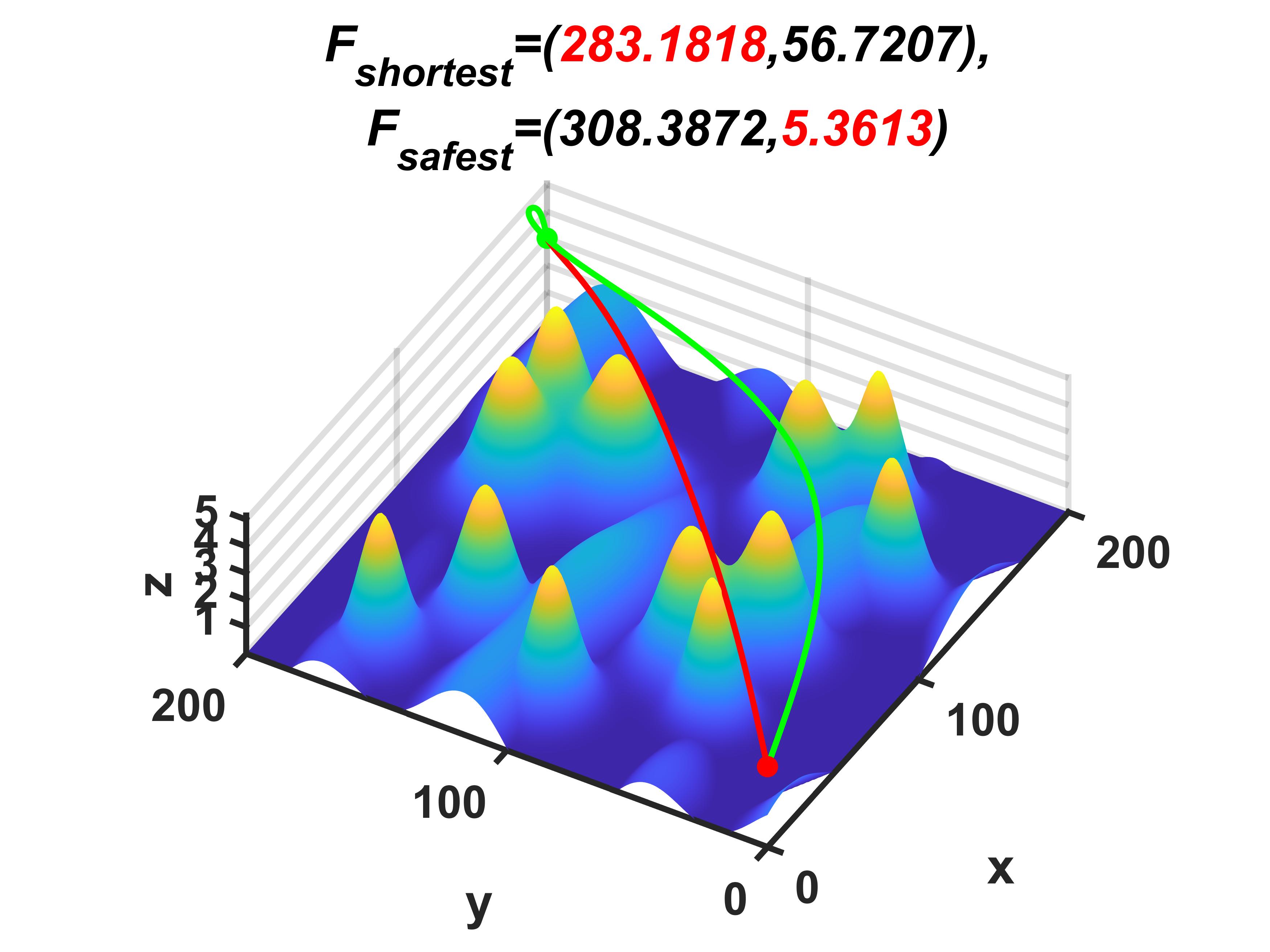}}\centerline{\footnotesize{(b4)}}
	\end{minipage}
 
    \begin{minipage}{0.24\linewidth}
		\vspace{3pt}\centerline{\includegraphics[width=4.5cm]{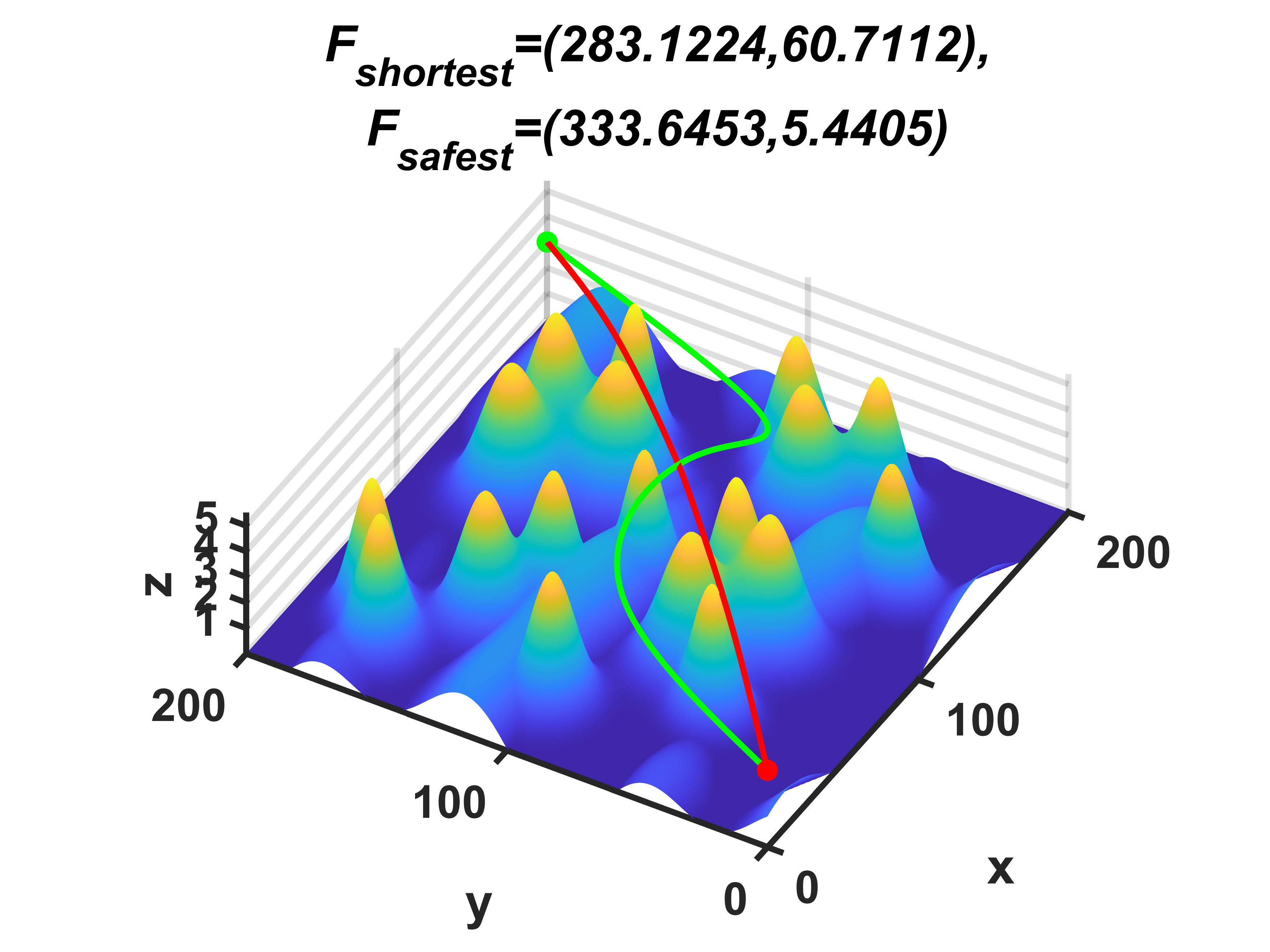}}\centerline{\footnotesize{(c1)}}
	\end{minipage}
     \begin{minipage}{0.24\linewidth}
		\vspace{3pt}\centerline{\includegraphics[width=4.5cm]{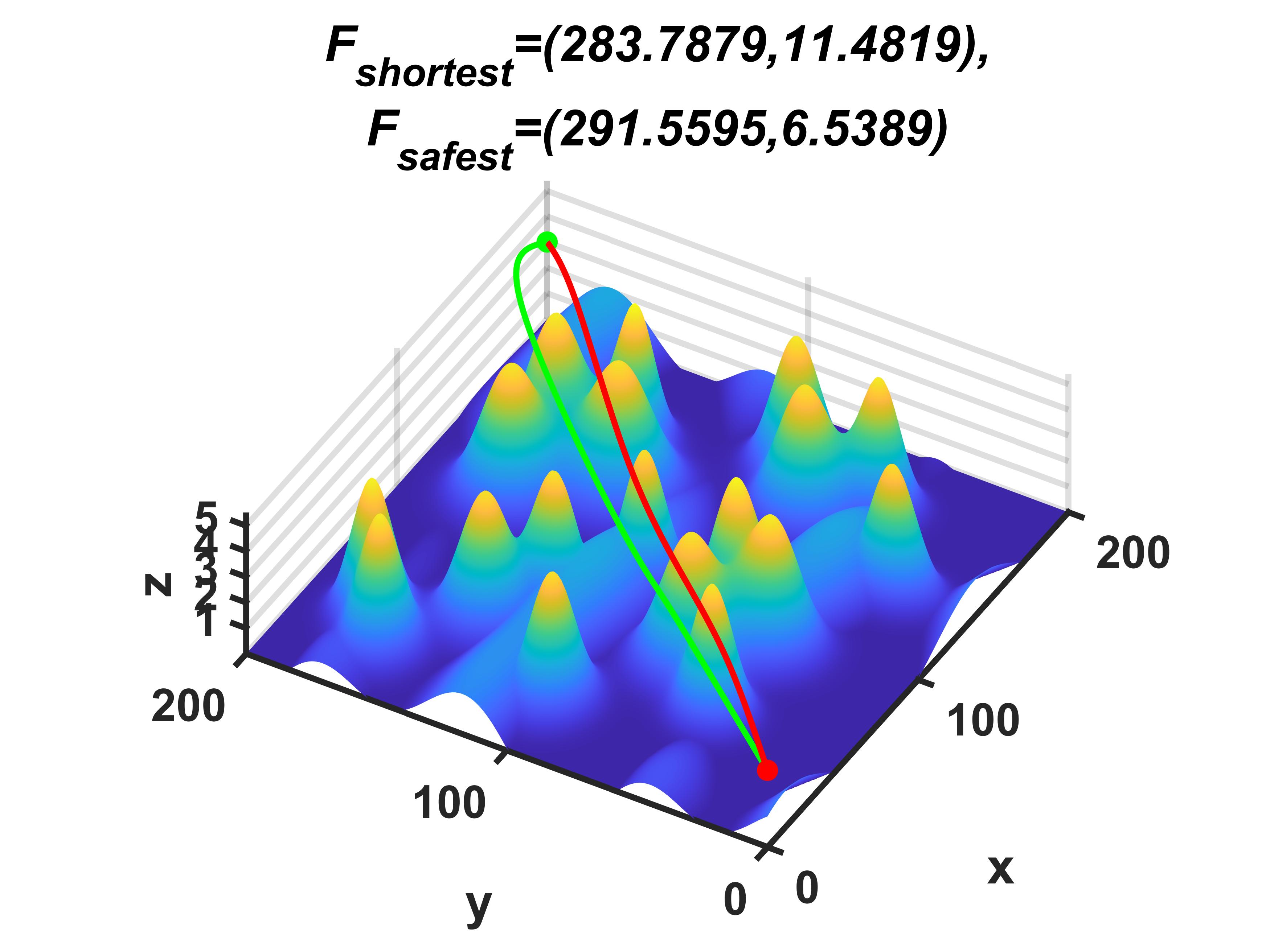}}\centerline{\footnotesize{(c2)}}
	\end{minipage}
     \begin{minipage}{0.24\linewidth}
		\vspace{3pt}\centerline{\includegraphics[width=4.5cm]{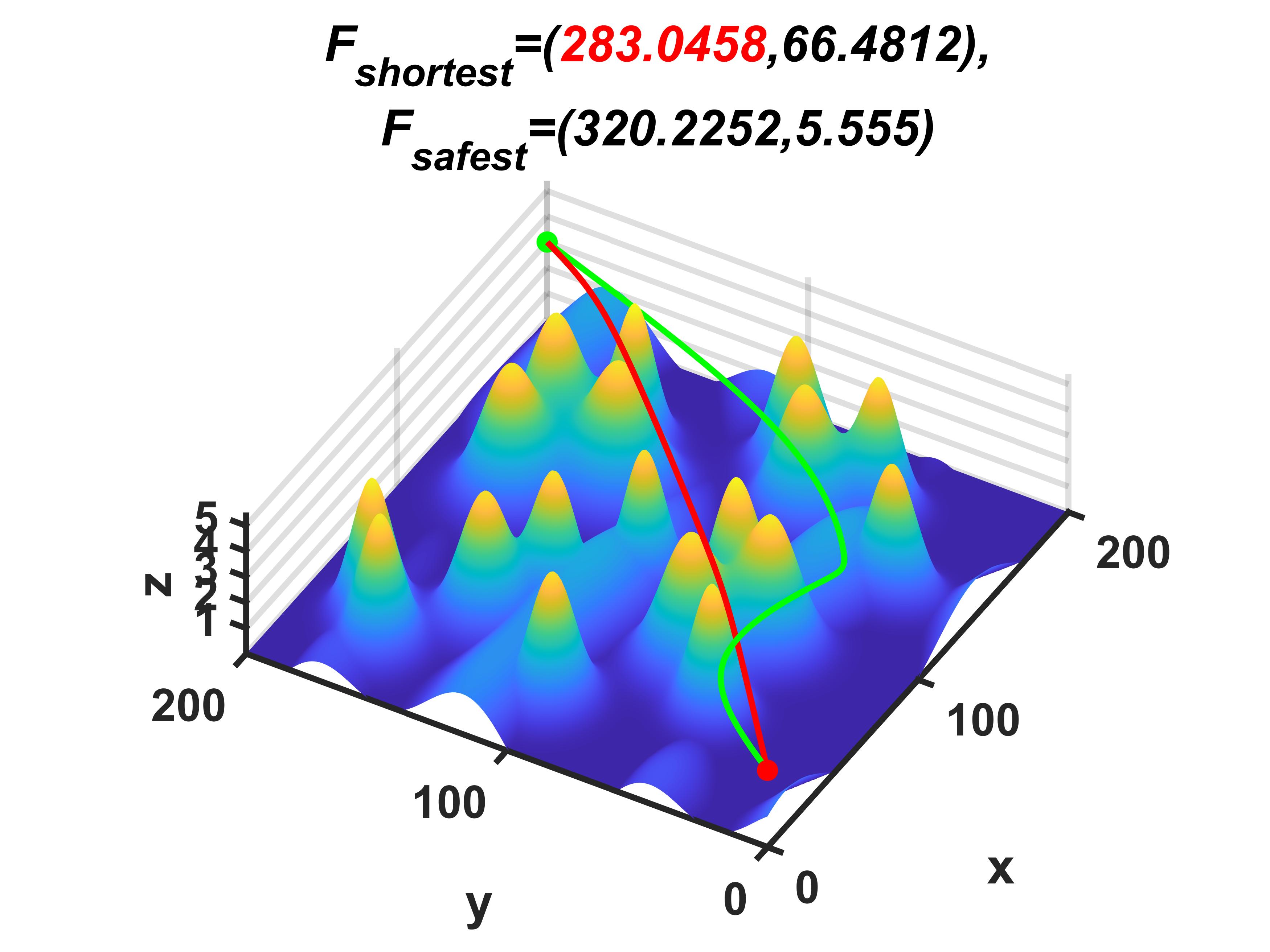}}\centerline{\footnotesize{(c3)}}
	\end{minipage}
     \begin{minipage}{0.24\linewidth}
		\vspace{3pt}\centerline{\includegraphics[width=4.5cm]{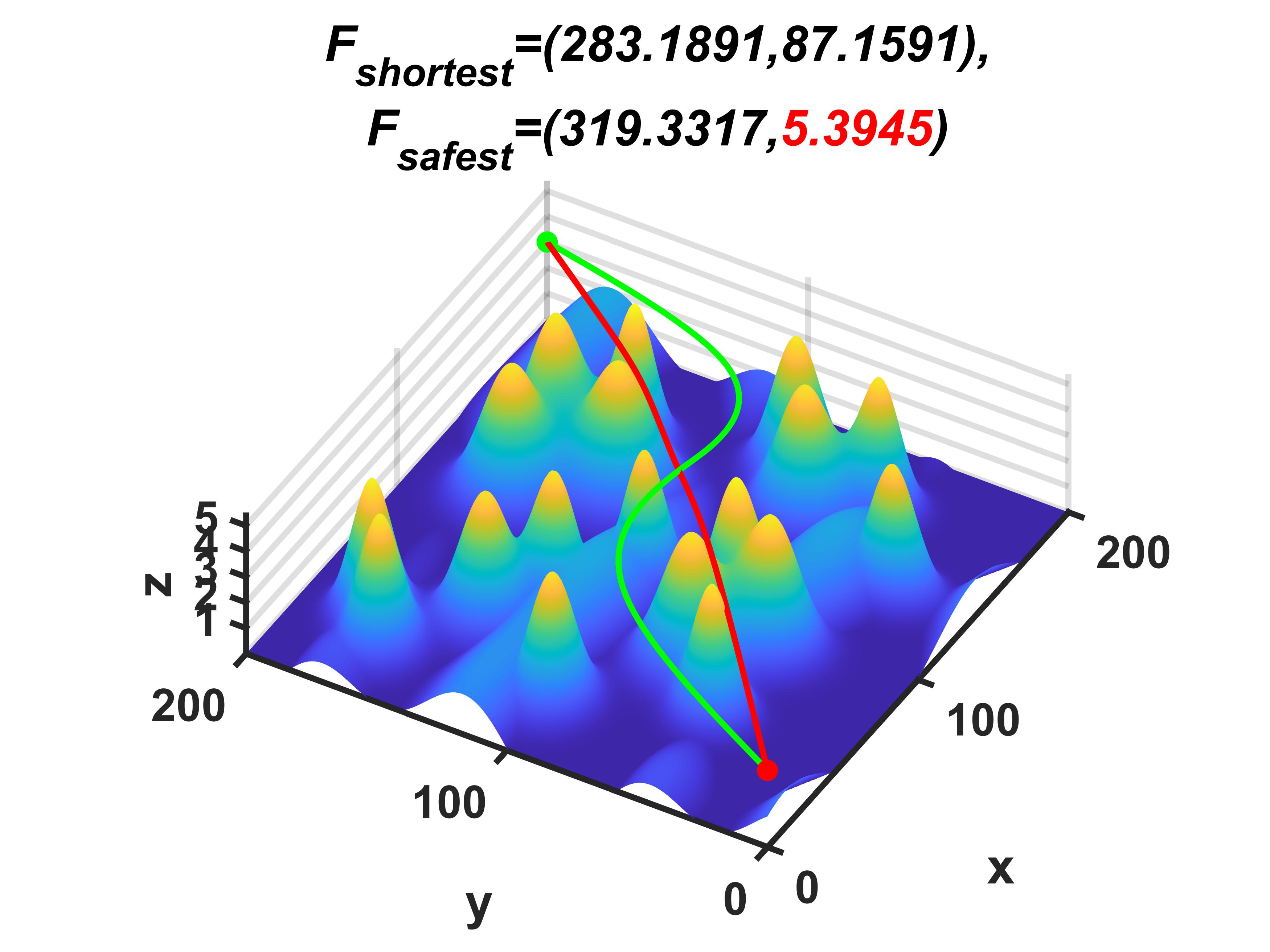}}\centerline{\footnotesize{(c4)}}
	\end{minipage}
 
	\begin{minipage}{0.24\linewidth}
		\vspace{3pt}\centerline{\includegraphics[width=4.5cm]{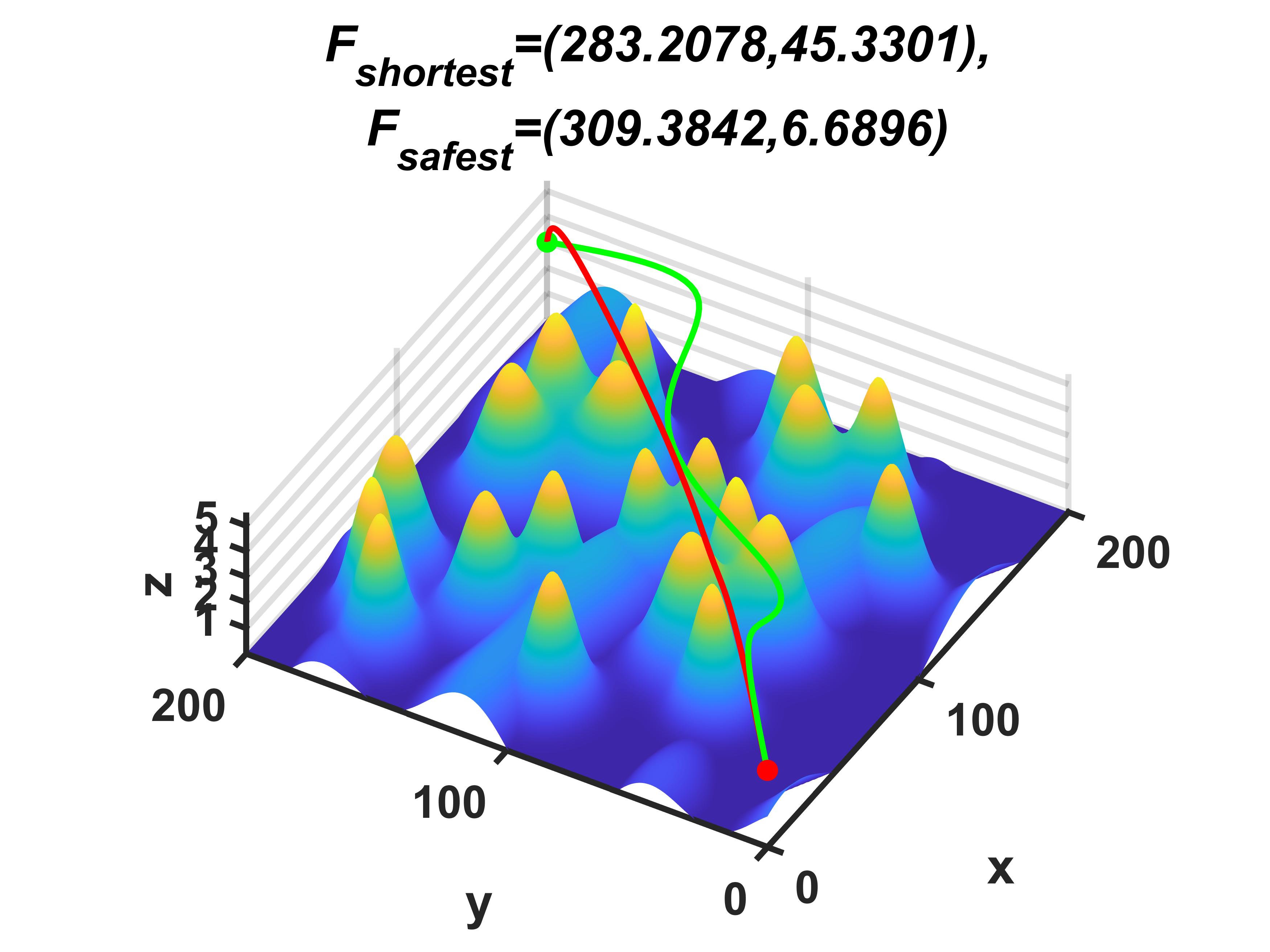}}\centerline{\footnotesize{(d1)}}
	\end{minipage}
    \begin{minipage}{0.24\linewidth}
		\vspace{3pt}\centerline{\includegraphics[width=4.5cm]{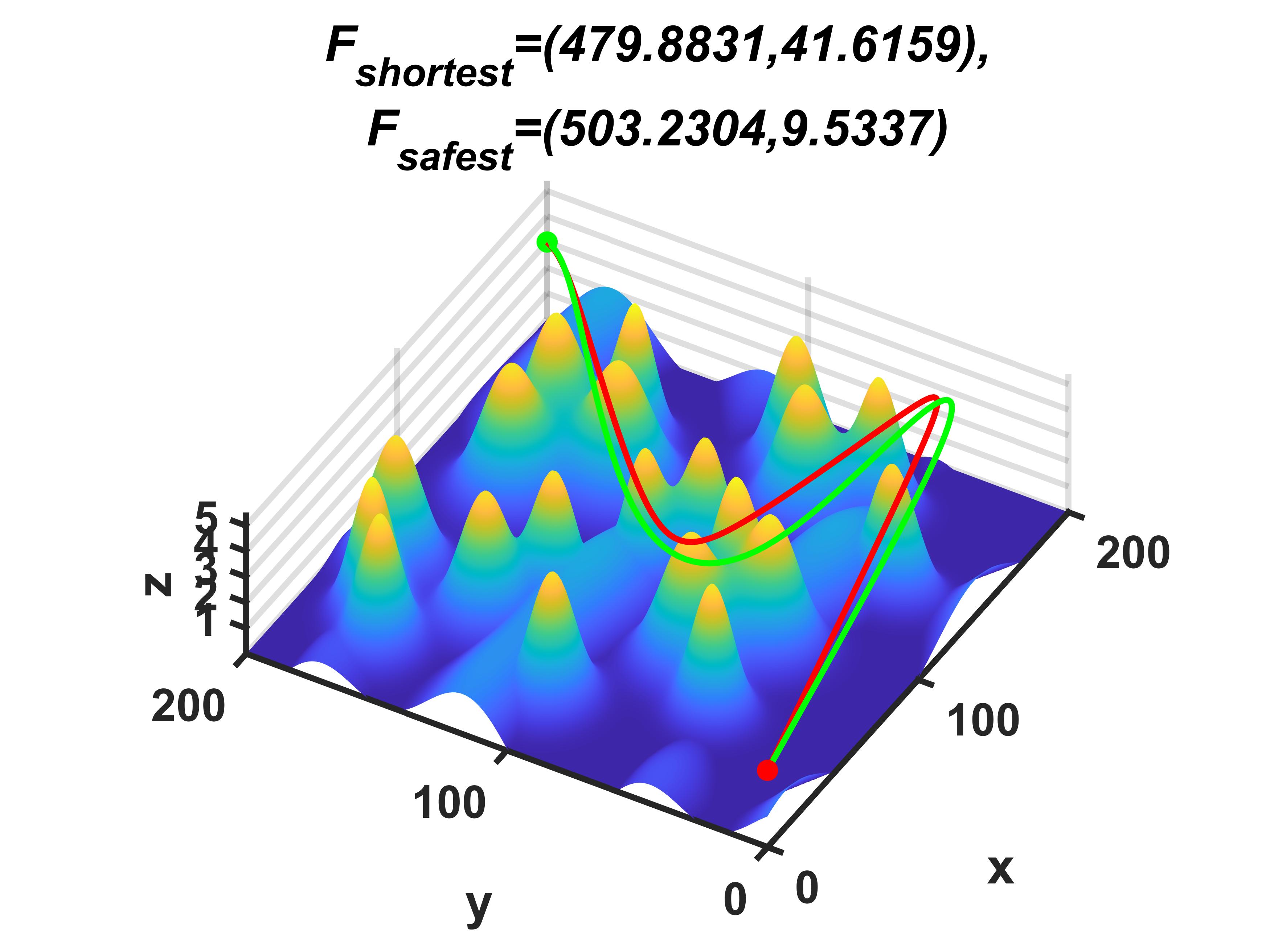}}\centerline{\footnotesize{(d2)}}
	\end{minipage}
    \begin{minipage}{0.24\linewidth}
		\vspace{3pt}\centerline{\includegraphics[width=4.5cm]{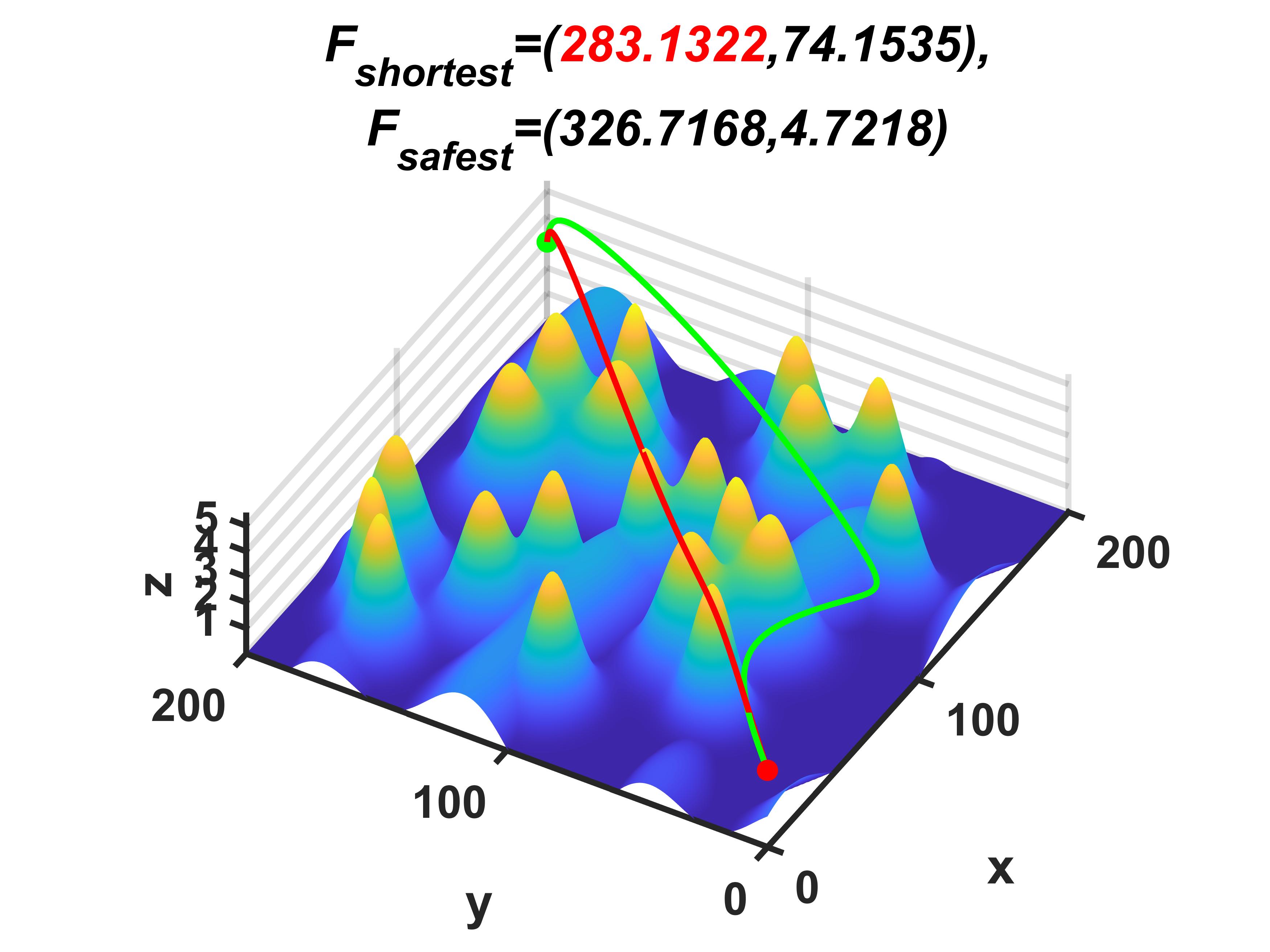}}\centerline{\footnotesize{(d3)}}
	\end{minipage}
    \begin{minipage}{0.24\linewidth}
		\vspace{3pt}\centerline{\includegraphics[width=4.5cm]{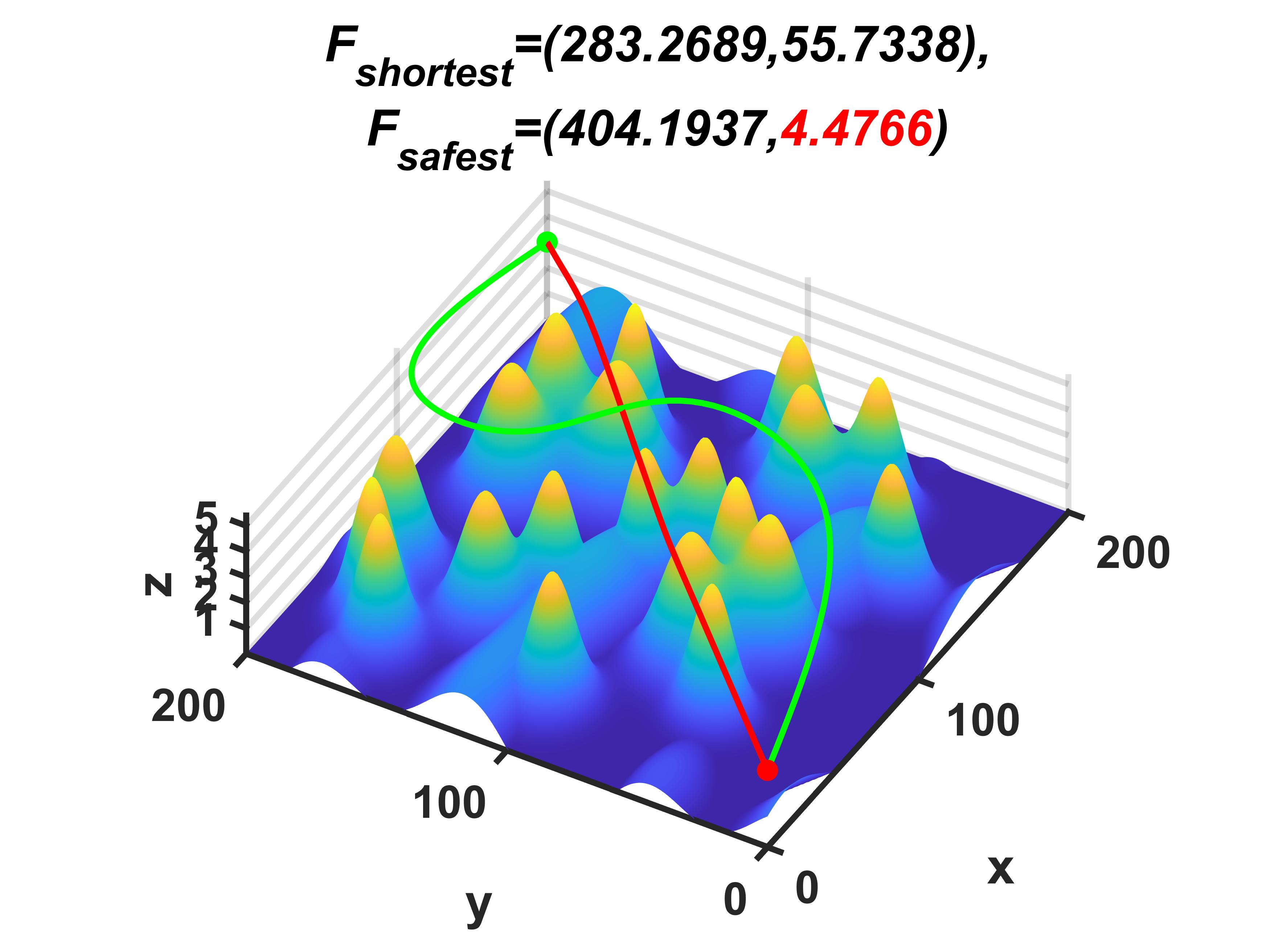}}\centerline{\footnotesize{(d4)}\vspace{1.5ex}}
	\end{minipage}
 
    \caption{The shortest path (red path) and the safest path (green path) in mountain environments. (a1)-(a4): C3; (b1)-(b4): C6; (c1)-(c4): C9; (d1)-(d4): C10. (a1)-(d1): NSGA-III; (a2)-(d2): C-MOEA/D; (a3)-(d3): MOEA/D-AWA; (a4)-(d4): MOEA/D-AAWA.}
    \label{fig7}
\end{figure}
Under the mountain environment, we generate ten cases to verify the effectiveness of our proposed method compared with NSGA-III, C-MOEA/D and MOEA/D-AWA as shown in Table \ref{tbl2}. Four algorithms are conducted on the above-mentioned ten cases over 30 independent runs. The mean and standard deviation (Std) of HV and PD are represented in Table \ref{tbl2}. The number within a bracket denotes the ranking of each algorithm for each case. For each case, the best result is in boldface. For each algorithm, its overall ranking is equal to the sum of rankings for all cases. A smaller overall ranking indicates a better result. In the ten cases (i.e., C1-C10), our proposed algorithm achieves the highest overall ranking, slightly higher than MOEA/D-AWA, followed by NSGA-III and C-MOEA/D, indicating that weight adjustment strategies can effectively improve algorithm performance. The proposed algorithm obtains the highest mean HV values on seven out of ten cases. Meanwhile, the proposed algorithm also achieves the highest mean PD values on seven out of ten cases. It manifests that the proposed algorithm can obtain the optimal convergence and diversity in most mountain environments.

For each run, the score of each algorithm is equal to the sum of the rankings of HV and PD. Fig. \ref{fig6} presents the final nondominated solutions of the four algorithms at the run with the median score. Four sets of final nondominated solutions in cases C3, C6, C9 and C10 are selected to visually compare the performance of the four algorithms. It can be observed that the final nondominated solutions in the objective space have a sharp peak and a low tail. Thus, it is difficult to achieve the shortest and safest path under the mountain environments. From Fig. \ref{fig6}, NSGA-III and C-MOEA/D can hardly reach these sharp peak and low tail areas . Conversely, the proposed algorithm can find more solutions in these areas. It further indicates that the proposed algorithm has a good performance in solution diversity. The reasons are given as follows. On the one hand, the proposed AAWA can guide individuals to approach such sparse areas according to the current population distribution in the objective space. On the other hand, the core idea of MOEA/D is decomposition and cooperation. Individuals evolve cooperatively based on neighborhood relationships. And diverse neighborhood with our strategy provides more valuable information for evolution.

Fig. \ref{fig7} presents the shortest path (red path) and the safest path (green path) at the run with the median score in cases C3, C6, C9 and C10. For each algorithm, its objective function values of the two paths are displayed at the top of the figure, where $F_{shortest}$ is the objective function value of the shortest path and $F_{safest}$ is that of the safest path. For each case, the shortest length and the lowest threat degree among the four algorithms are marked in red. From Fig. \ref{fig7}, in most path planning cases, the proposed algorithm can provide decision-makers with the shorter or safer paths, even both at the same time. It implies the superiority of the proposed algorithm again.

\subsubsection{Experiments under urban environments}\label{cha5.4.2}

\begin{table}[pos=htb, width=\textwidth]
    \caption{Mean and Std values of HV and PD in urban environments}
    \label{tbl3}
    \begin{tabular*}{\tblwidth}{@{}LLLLLLL@{}}
        \toprule
        \multirow{2}{*}[-2pt]{Case}&\multirow{2}{*}[-2pt]{\makecell{Number \\of threats}} &\multirow{2}{*}[-2pt]{Metric}&\multicolumn{4}{c}{Algorithm}\\
		\cmidrule(lr){4-7}
		&&&NSGA-III&C-MOEA/D&MOEA/D-AWA&MOEA/D-AAWA\\\midrule
  
        \multirow{2}{*}[-2pt]{C11}&\multirow{2}{*}[-2pt]{2}
        &HV&0.884±0.0332 (3)&0.8464±0.028 (4)&0.8962±0.0161 (2)&\textbf{0.8966}±0.0134 \textbf{(1)}\\
        &&PD($\times 10^{4}$)&5.0693±0.8773 (3)&4.9734±0.7184 (4)&7.405±1.4496 (2)&\textbf{7.6929}±1.7622 \textbf{(1)}\\
        \multirow{2}{*}[-2pt]{C12}&\multirow{2}{*}[-2pt]{4}
        &HV&0.8915±0.027 (3)&0.8394±0.0254 (4)&0.8969±0.0151 (2)&\textbf{0.9004}±0.0169 \textbf{(1)}\\
        &&PD($\times 10^{4}$)&5.0002±0.5642 (4)&5.4638±1.307 (3)&\textbf{7.4205}±1.3327 \textbf{(1})&7.3192±1.485 (2)\\
        \multirow{2}{*}[-2pt]{C13}&\multirow{2}{*}[-2pt]{6}
        &HV&0.9013±0.0214 (2)&0.835±0.0384 (4)&0.8994±0.0169 (3)&\textbf{0.904}±0.0131 \textbf{(1)}\\
        &&PD($\times 10^{4}$)&5.3669±0.6157 (3)&4.7638±0.6612 (4)&7.7798±1.4584 (2)&\textbf{8.0438}±2.0033 \textbf{(1)}\\
        \multirow{2}{*}[-2pt]{C14}&\multirow{2}{*}[-2pt]{8}
        &HV&0.8615±0.0681 (3)&0.8087±0.0607 (4)&0.894±0.0385 (2)&\textbf{0.9046}±0.0226 \textbf{(1)}\\
        &&PD($\times 10^{4}$)&4.9181±1.3454 (3)&4.7567±0.9142 (4)&8.2094±1.766 (2)&\textbf{8.5427}±1.7508 \textbf{(1)}\\
        \multirow{2}{*}[-2pt]{C15}&\multirow{2}{*}[-2pt]{10}
        &HV&0.8322±0.1627 (3)&0.7964±0.1025 (4)&\textbf{0.9032}±0.0162 \textbf{(1)}&0.8994±0.03 (2)\\
        &&PD($\times 10^{4}$)&4.7939±1.5326 (3)&4.7188±1.1545 (4)&8.8987±1.8765 (2)&\textbf{9.4026}±3.8884 \textbf{(1)}\\
        \multirow{2}{*}[-2pt]{C16}&\multirow{2}{*}[-2pt]{12}
        &HV&0.8139±0.2264 (3)&0.7893±0.0566 (4)&0.8914±0.0317 (2)&\textbf{0.9063}±0.0221 \textbf{(1)}\\
        &&PD($\times 10^{4}$)&4.578±1.6819 (3)&4.0444±0.7341 (4)&\textbf{8.5236}±2.043 \textbf{(1)}&8.3838±2.1779 (2)\\
        \multirow{2}{*}[-2pt]{C17}&\multirow{2}{*}[-2pt]{14}
        &HV&0.4013±0.4098 (4)&0.6122±0.257 (3)&0.6203±0.4141 (2)&\textbf{0.6874}±0.387 \textbf{(1)}\\
        &&PD($\times 10^{4}$)&1.6159±1.8464 (4)&3.7112±4.1819 (3)&6.0677±4.4531 (2)&\textbf{6.4354}±4.0017 \textbf{(1)}\\
        \multirow{2}{*}[-2pt]{C18}&\multirow{2}{*}[-2pt]{16}
        &HV&0.2856±0.4119 (4)&0.5329±0.3396 (3)&0.5373±0.4467 (2)&\textbf{0.7378}±0.3379 \textbf{(1)}\\
        &&PD($\times 10^{4}$)&1.3799±2.2004 (4)&2.6108±1.7486 (3)&6.0971±5.5594 (2)&\textbf{7.8408}±4.775 \textbf{(1)}\\
        \multirow{2}{*}[-2pt]{C19}&\multirow{2}{*}[-2pt]{18}
        &HV&0.3079±0.4123 (4)&\textbf{0.5004}±0.3232 \textbf{(1)}&0.4594±0.4404 (3)&0.4674±0.4471 (2)\\
        &&PD($\times 10^{4}$)&1.5621±2.301 (4)&2.3766±1.6115 (3)&4.8437±4.9738 (2)&\textbf{4.9161}±5.0176 \textbf{(1)}\\
        \multirow{2}{*}[-2pt]{C20}&\multirow{2}{*}[-2pt]{20}
        &HV&0.1331±0.3036 (4)&\textbf{0.3784}±0.3426 \textbf{(1)}&0.3739±0.4364 (2)&0.2566±0.4 (3)\\
        &&PD($\times 10^{4}$)&0.6663±1.6027 (4)&2.4107±3.2936 (2)&\textbf{4.151}±5.057 \textbf{(1)}&2.2967±3.6426 (3)\\\midrule

        \multicolumn{3}{c}{Overall ranking}&(68)&(66)&(38)&\textbf{(28)}\\
        \bottomrule
    \end{tabular*}
\end{table}
\begin{figure}[pos=htb]
    \centering
        \subfigure[]{\includegraphics[width=4cm]{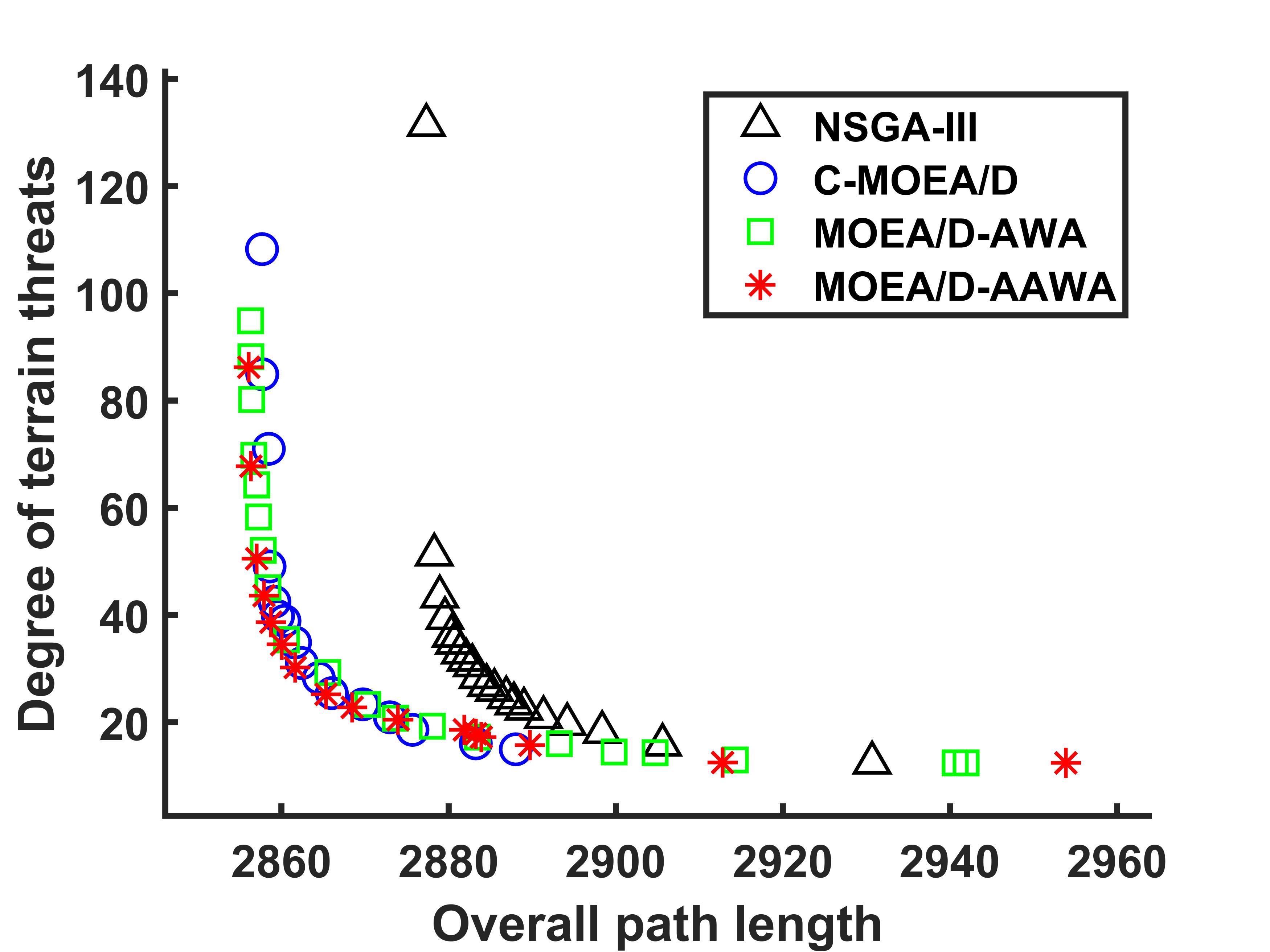}}
        \subfigure[]{\includegraphics[width=4cm]{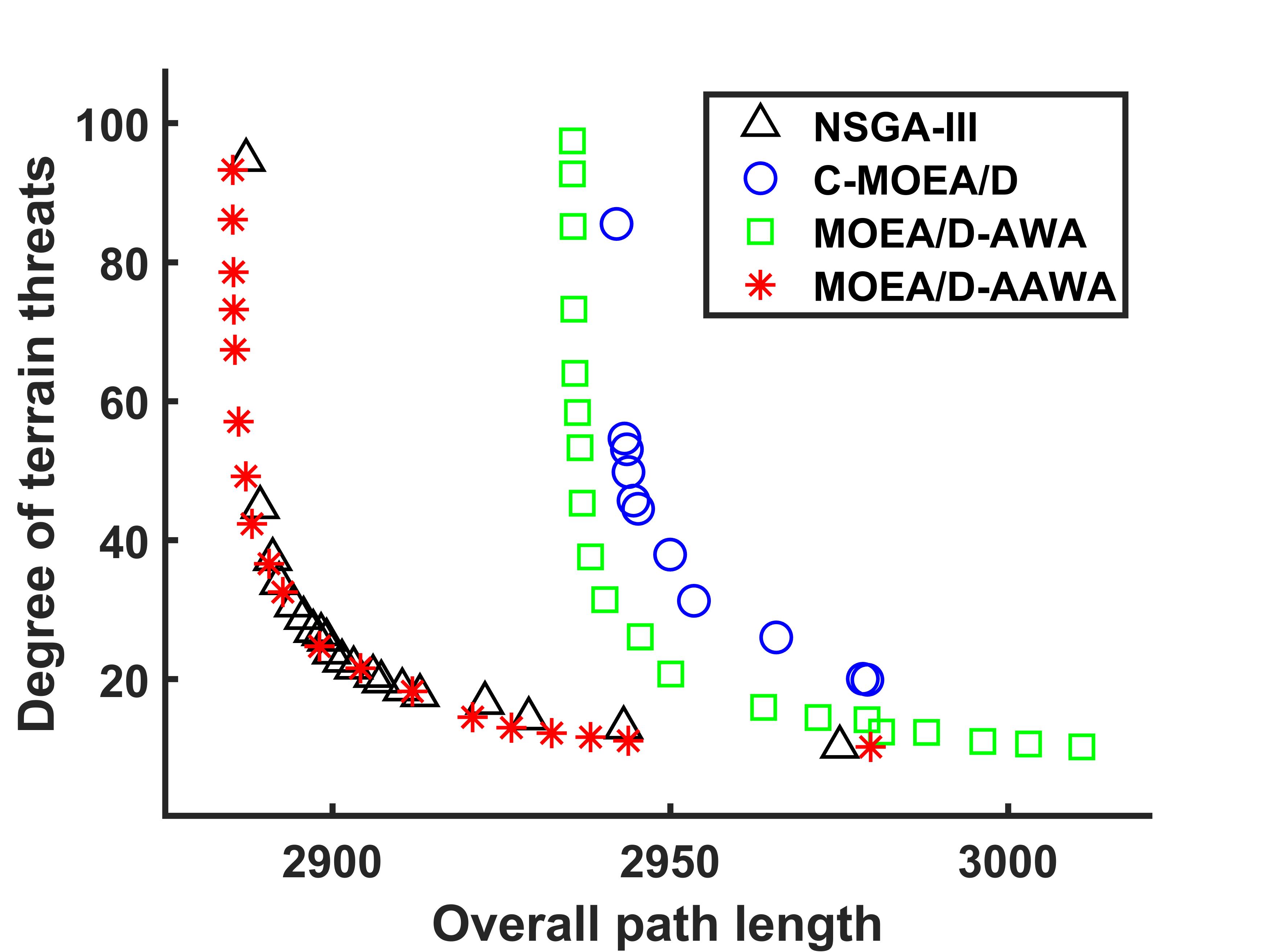}}
        \subfigure[]{\includegraphics[width=4cm]{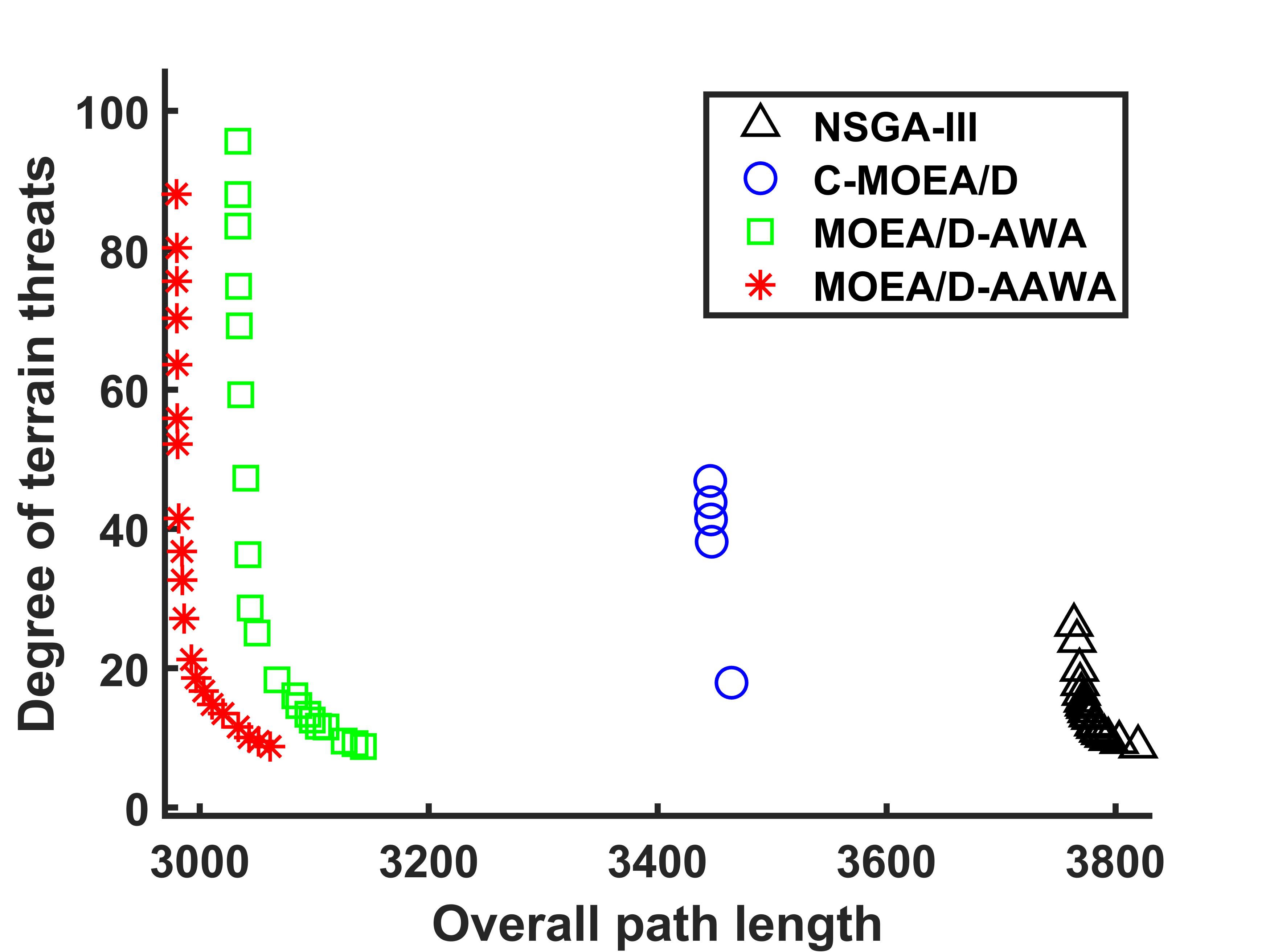}}
        \subfigure[]{\includegraphics[width=4cm]{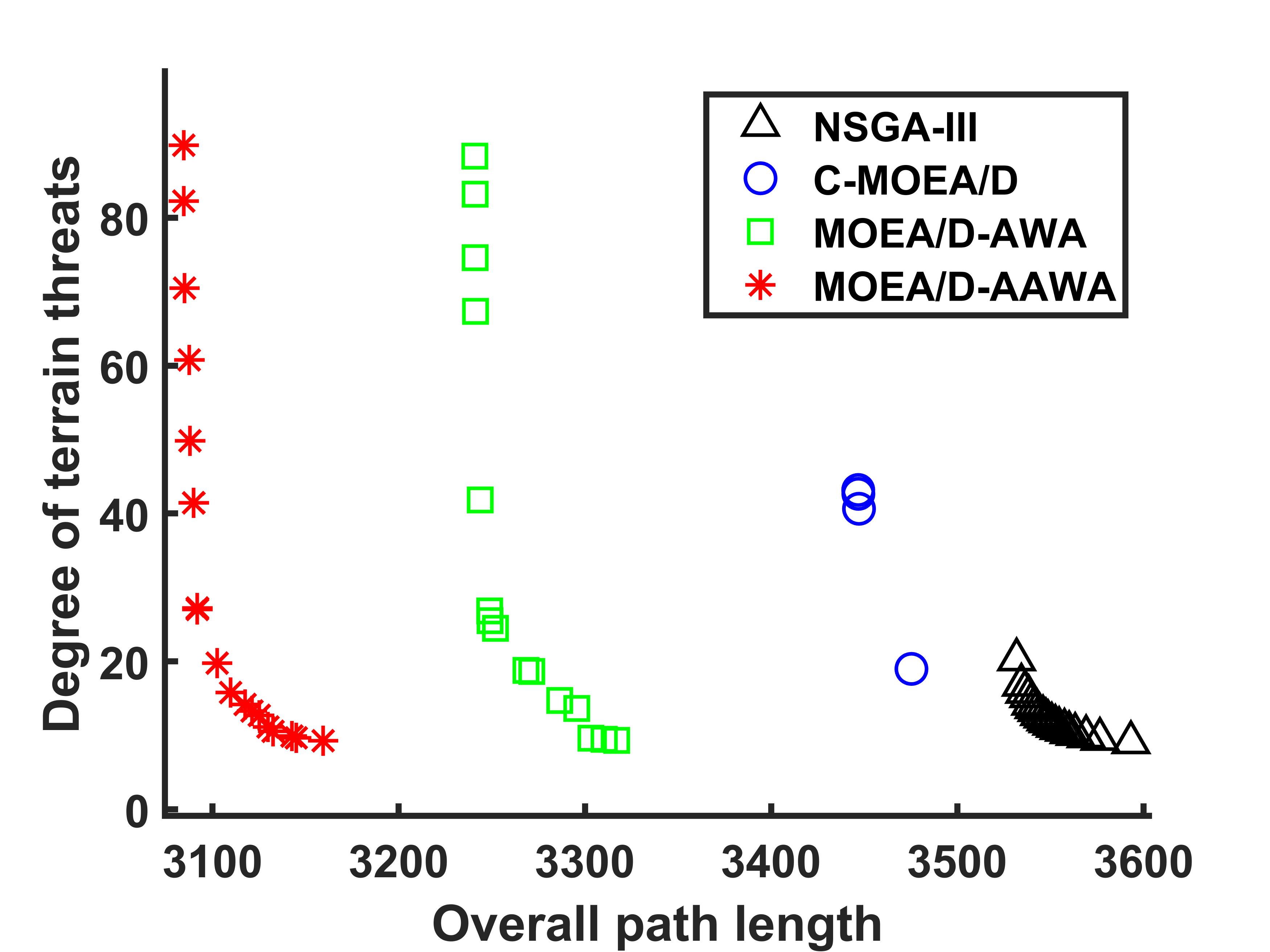}}
    \caption{Final nondominated solutions in urban environments. (a): C11; (b): C14; (c): C18; (d): C19.}
    \label{fig8}
\end{figure}
\begin{figure}[pos=htb]
	\begin{minipage}{0.24\linewidth}
		\vspace{3pt}\centerline{\includegraphics[width=4.5cm]{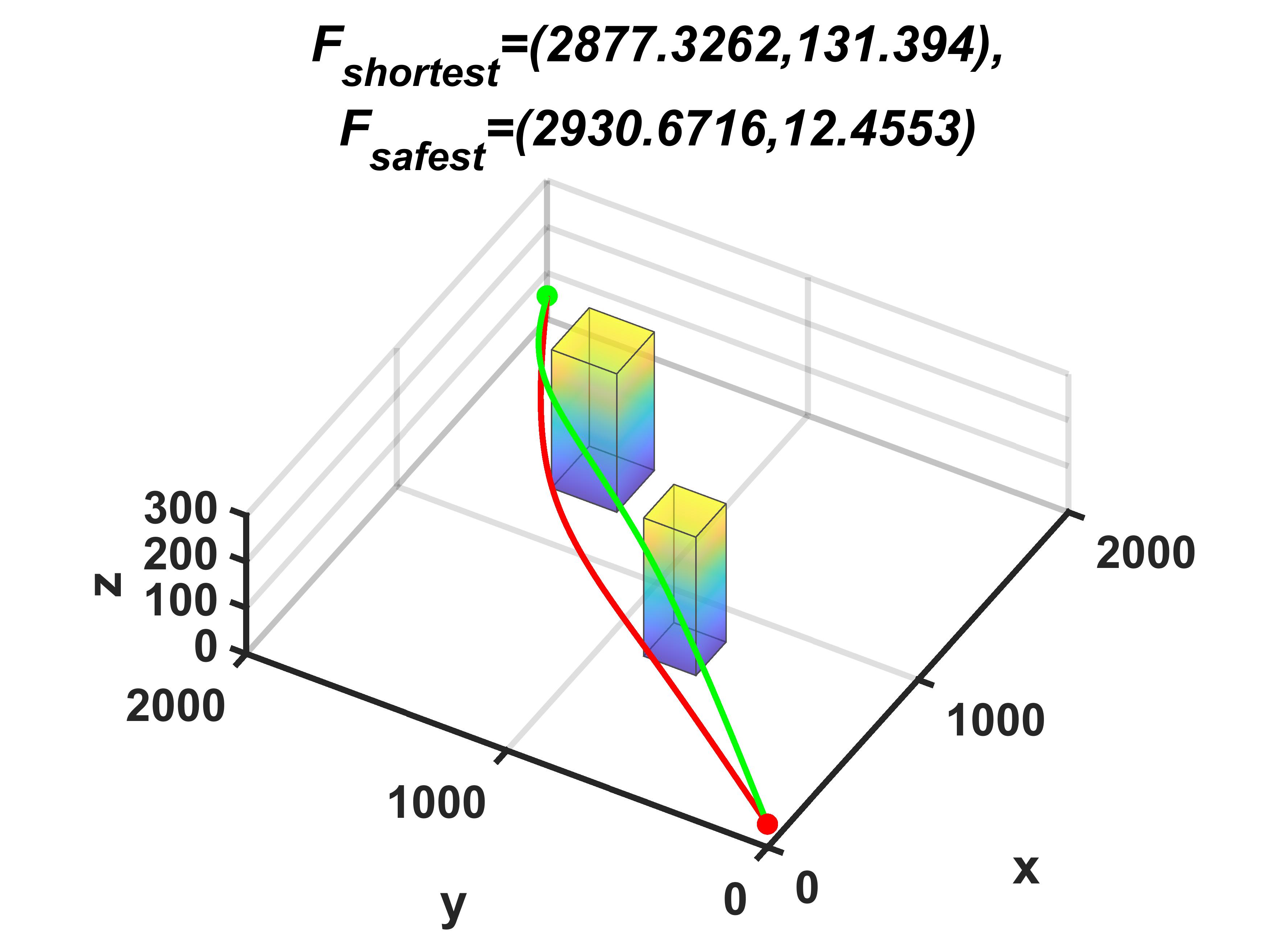}}\centerline{\footnotesize{(a1)}}
	\end{minipage}
    \begin{minipage}{0.24\linewidth}
		\vspace{3pt}\centerline{\includegraphics[width=4.5cm]{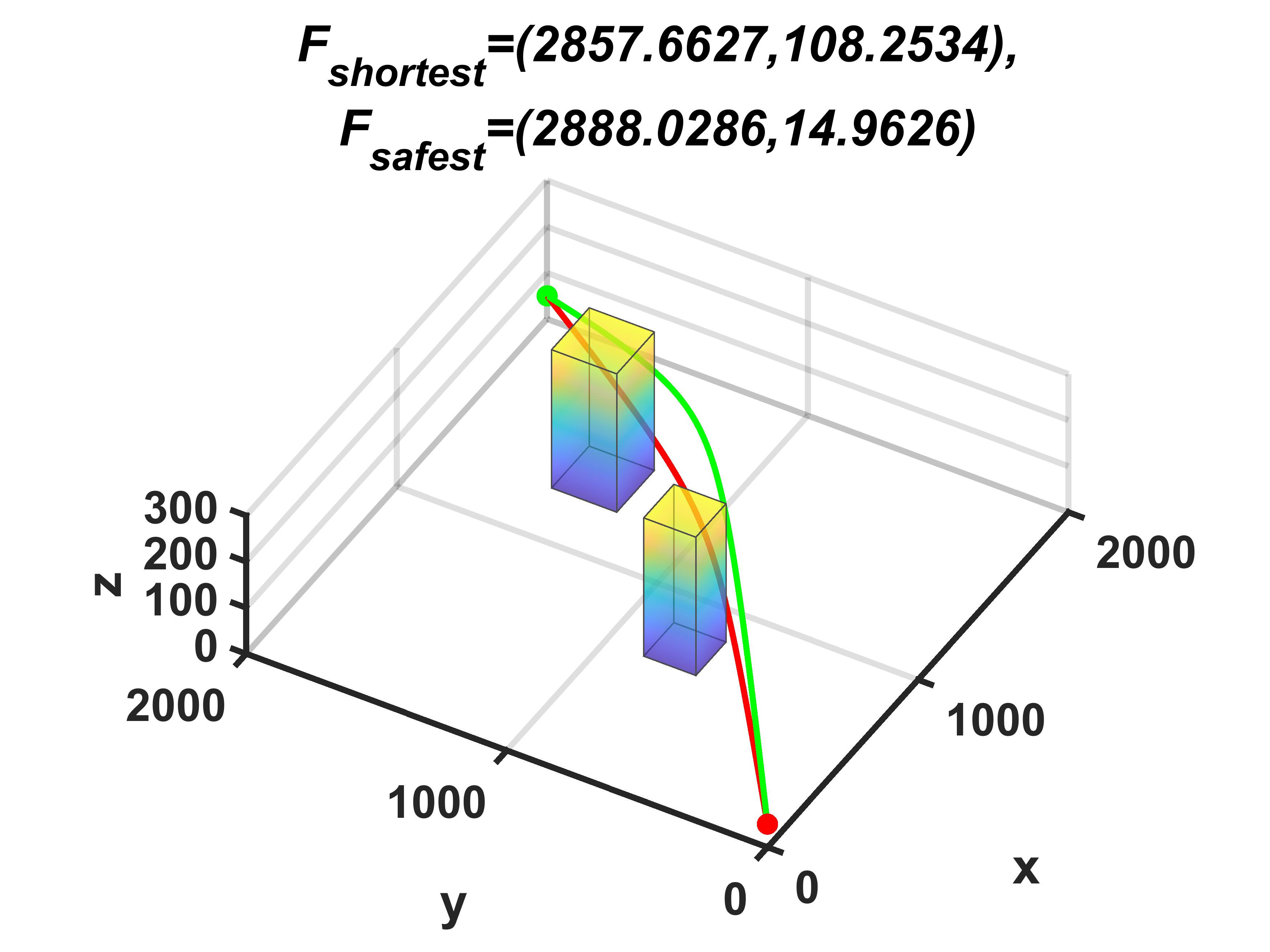}}\centerline{\footnotesize{(a2)}}
	\end{minipage}
    \begin{minipage}{0.24\linewidth}
		\vspace{3pt}\centerline{\includegraphics[width=4.5cm]{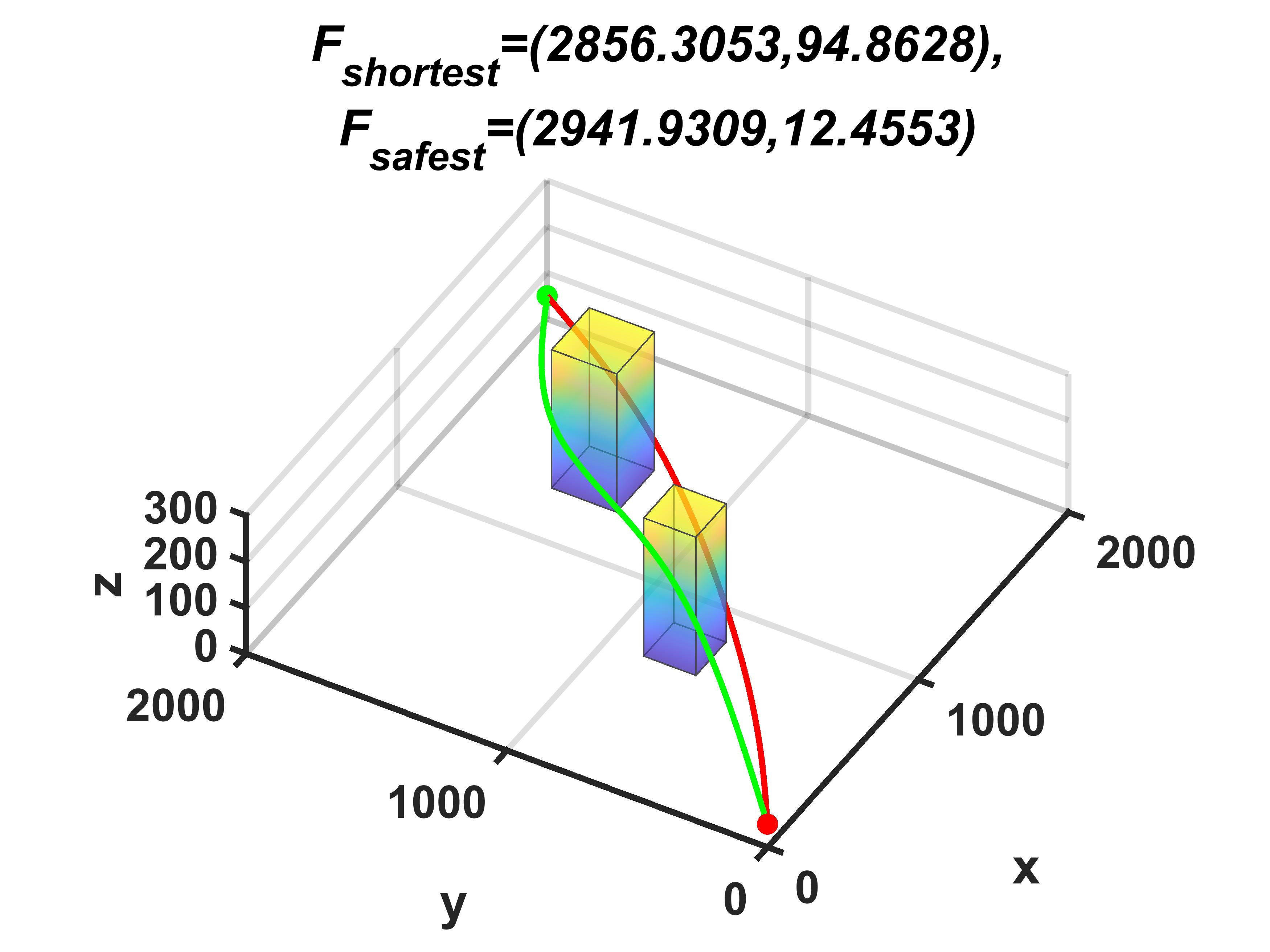}}\centerline{\footnotesize{(a3)}}
	\end{minipage}
    \begin{minipage}{0.24\linewidth}
		\vspace{3pt}\centerline{\includegraphics[width=4.5cm]{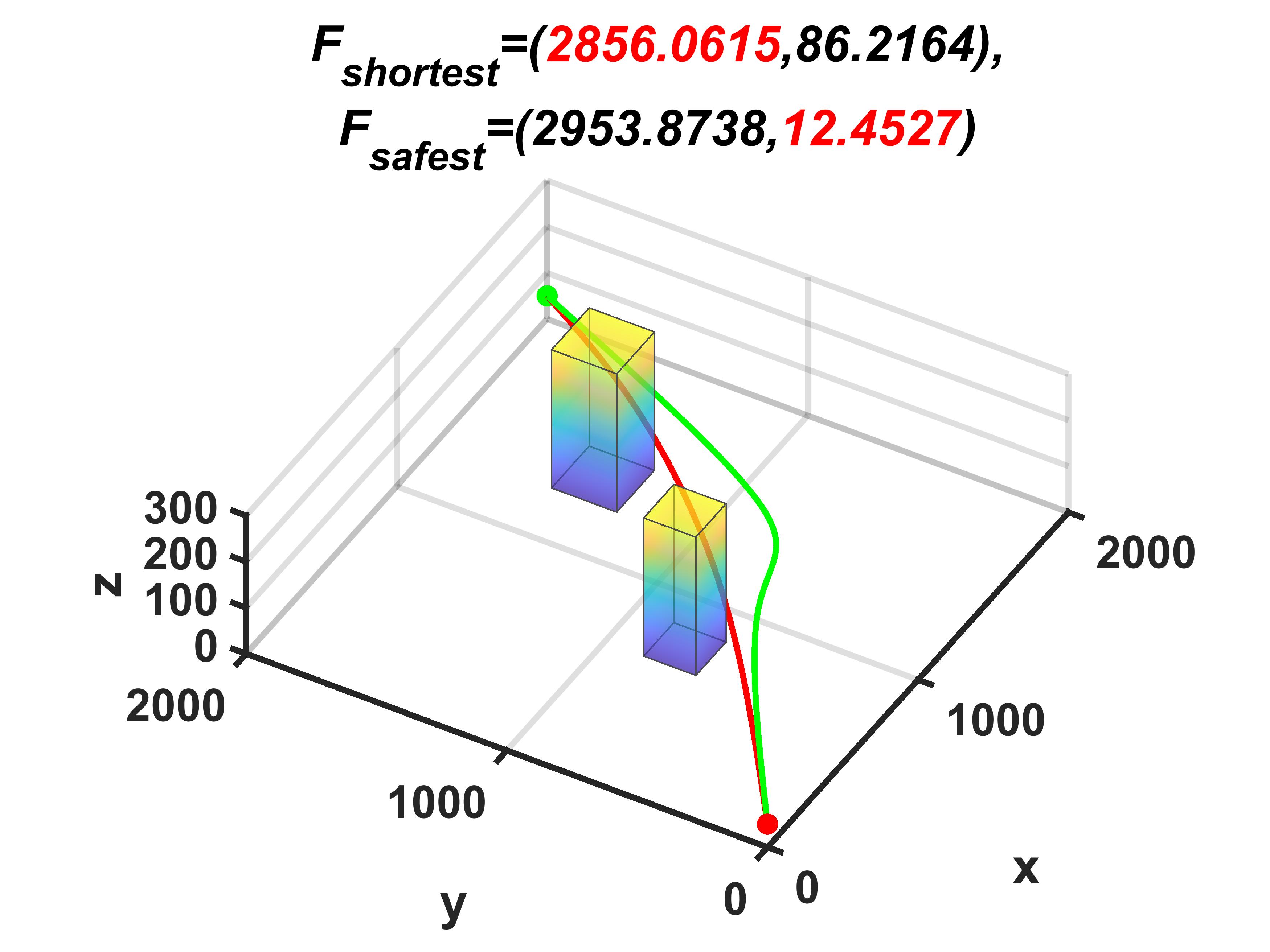}}\centerline{\footnotesize{(a4)}}
	\end{minipage}
 
    \begin{minipage}{0.24\linewidth}
		\vspace{3pt}\centerline{\includegraphics[width=4.5cm]{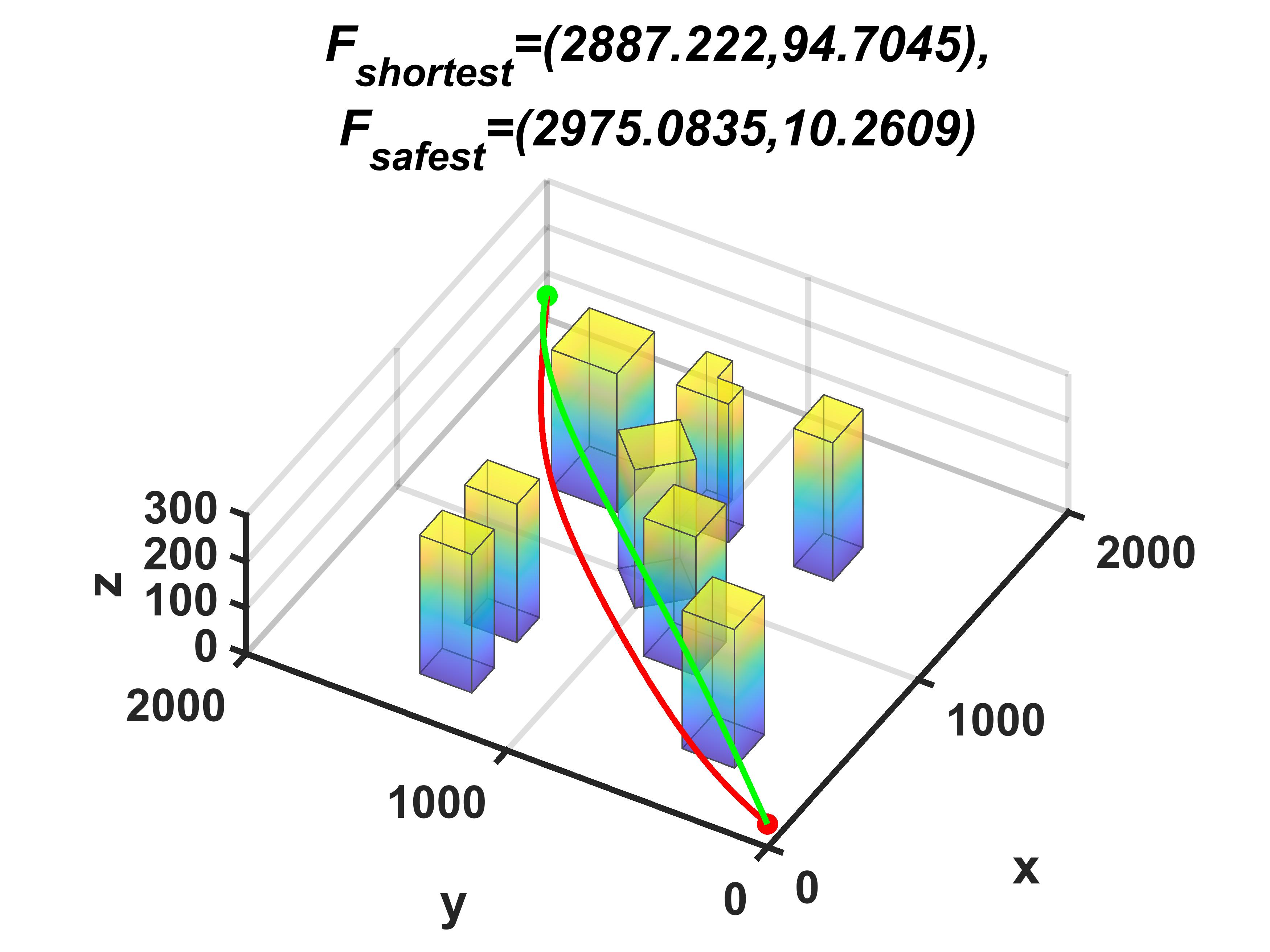}}\centerline{\footnotesize{(b1)}}
	\end{minipage}
     \begin{minipage}{0.24\linewidth}
		\vspace{3pt}\centerline{\includegraphics[width=4.5cm]{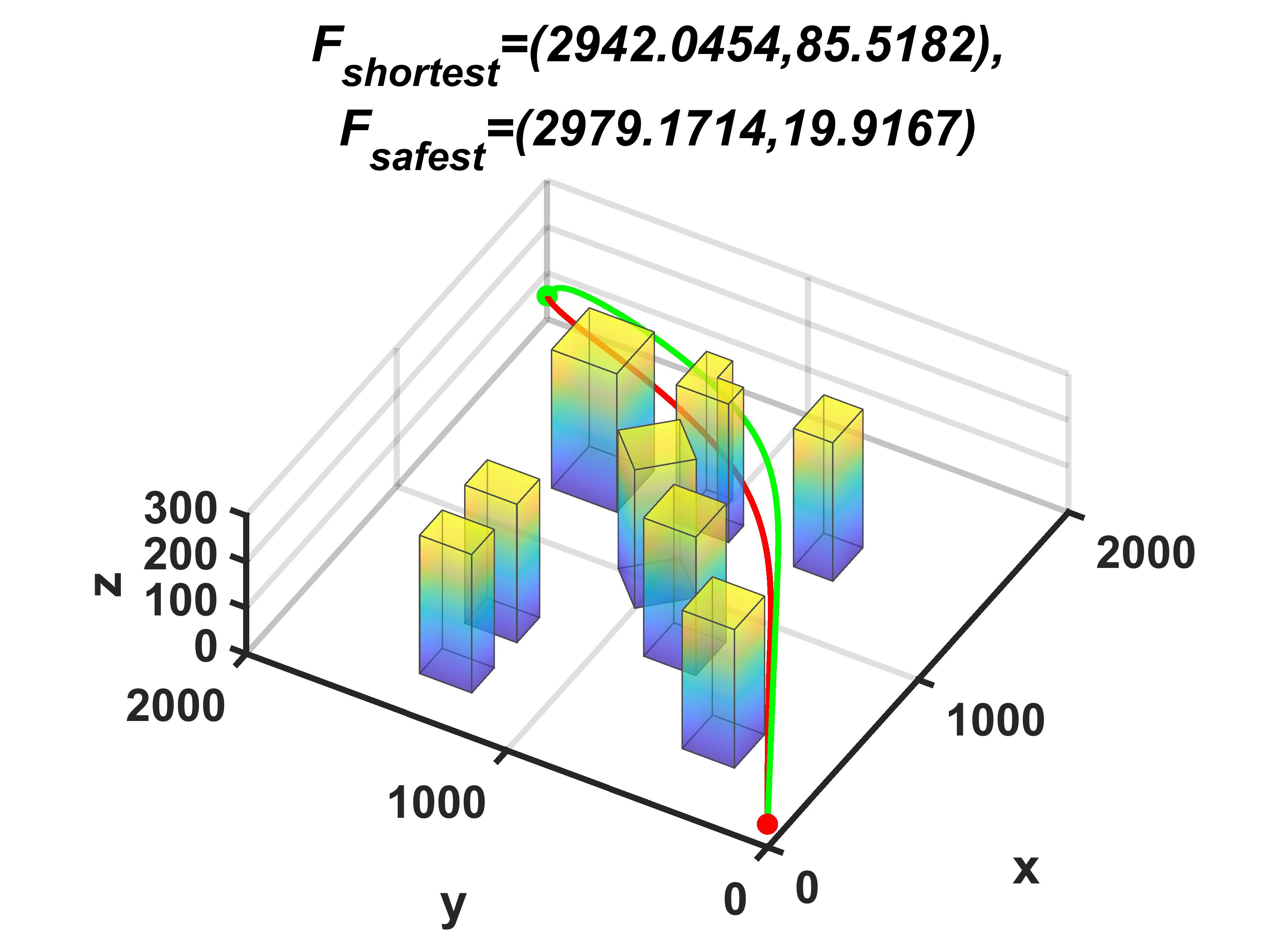}}\centerline{\footnotesize{(b2)}}
	\end{minipage}
     \begin{minipage}{0.24\linewidth}
		\vspace{3pt}\centerline{\includegraphics[width=4.5cm]{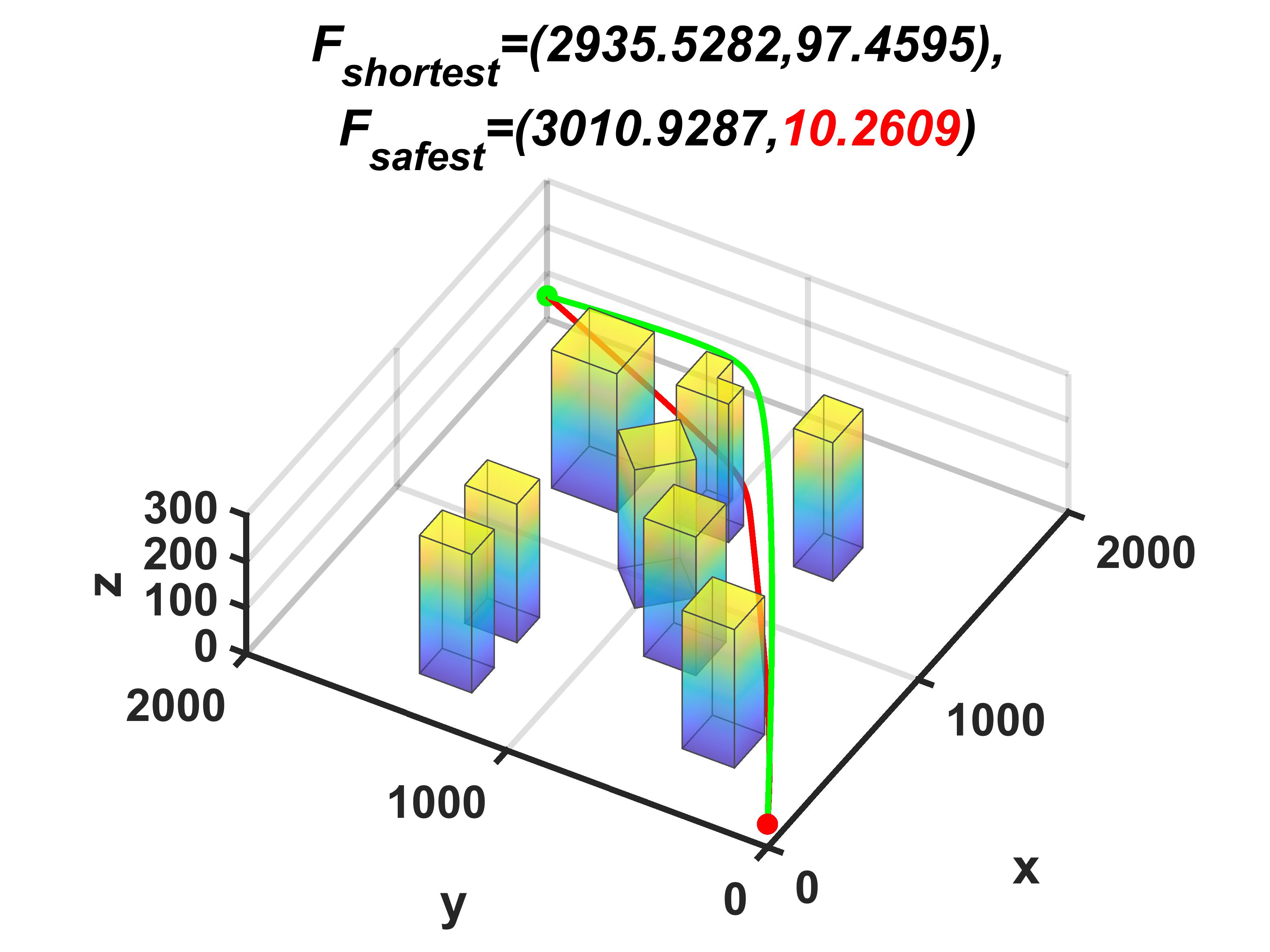}}\centerline{\footnotesize{(b3)}}
	\end{minipage}
     \begin{minipage}{0.24\linewidth}
		\vspace{3pt}\centerline{\includegraphics[width=4.5cm]{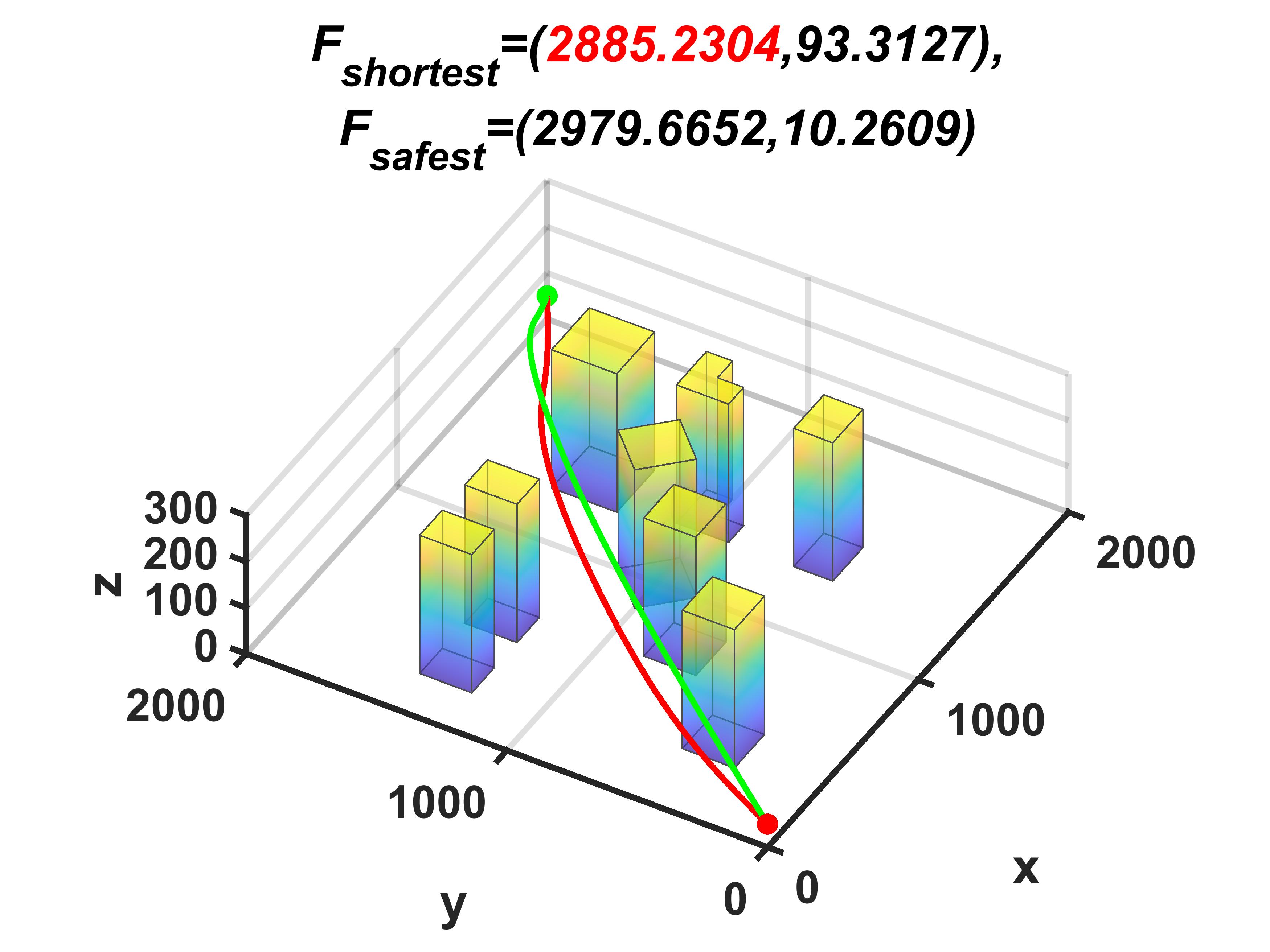}}\centerline{\footnotesize{(b4)}}
	\end{minipage}
 
    \begin{minipage}{0.24\linewidth}
		\vspace{3pt}\centerline{\includegraphics[width=4.5cm]{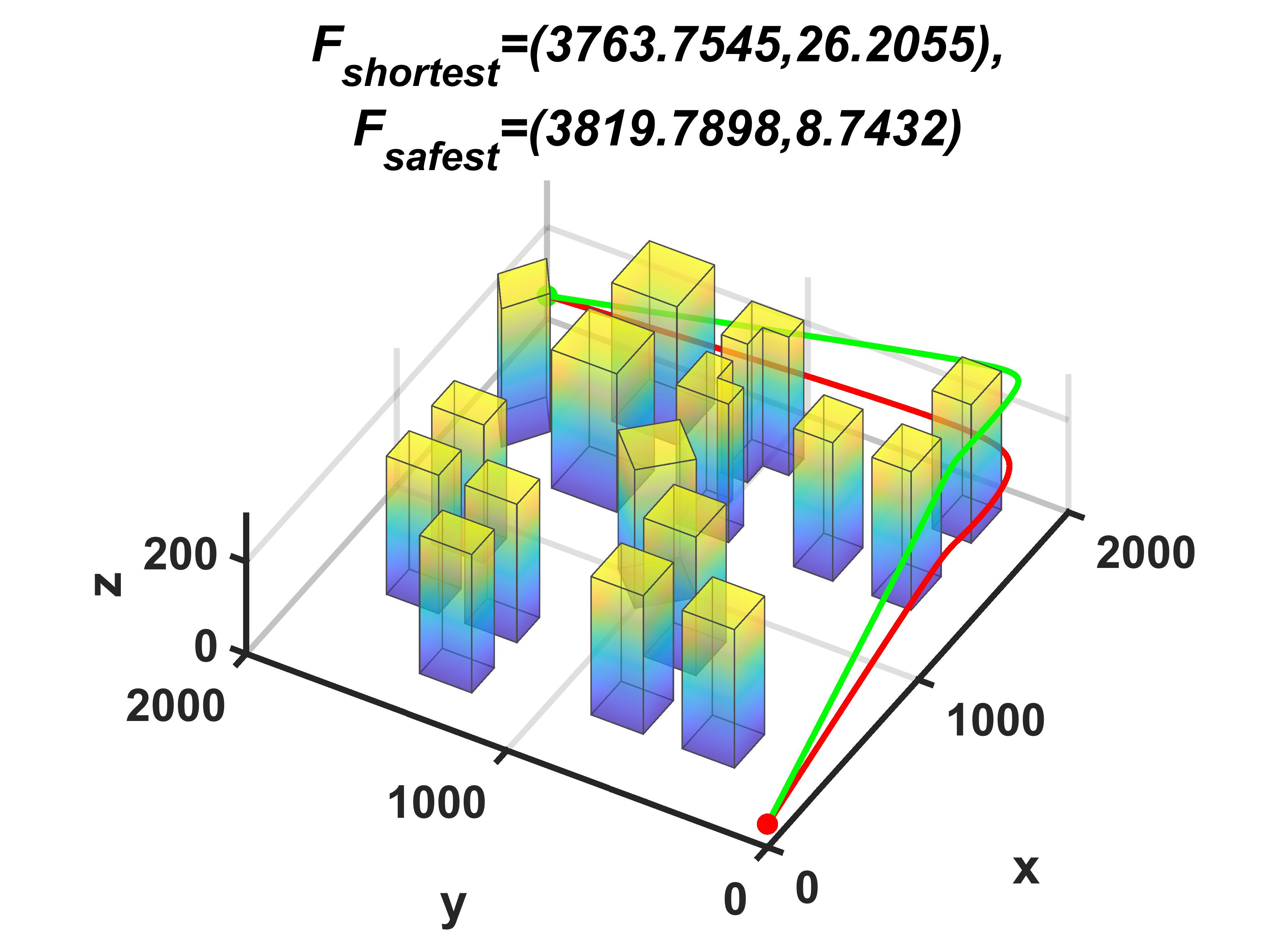}}\centerline{\footnotesize{(c1)}}
	\end{minipage}
     \begin{minipage}{0.24\linewidth}
		\vspace{3pt}\centerline{\includegraphics[width=4.5cm]{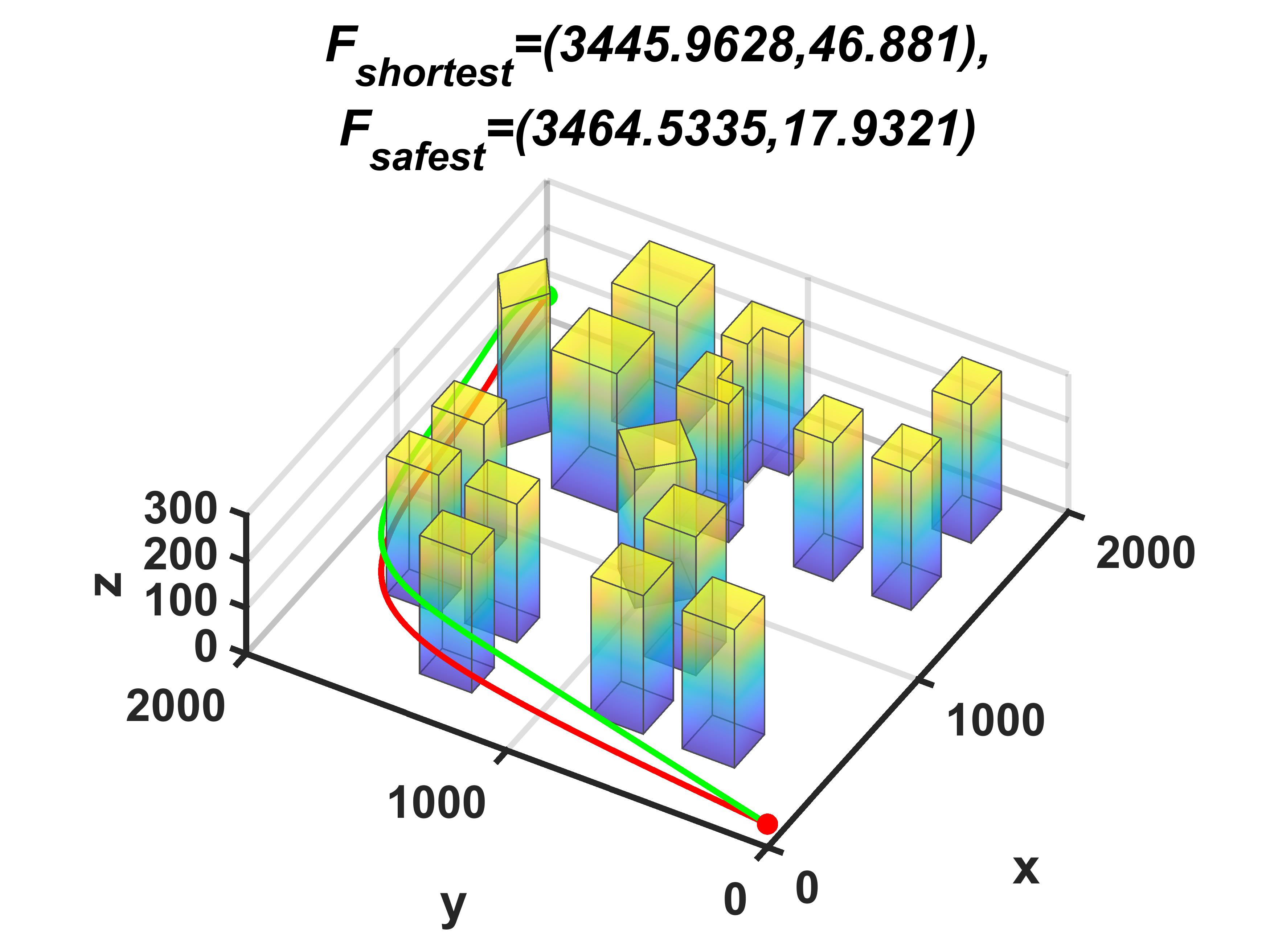}}\centerline{\footnotesize{(c2)}}
	\end{minipage}
     \begin{minipage}{0.24\linewidth}
		\vspace{3pt}\centerline{\includegraphics[width=4.5cm]{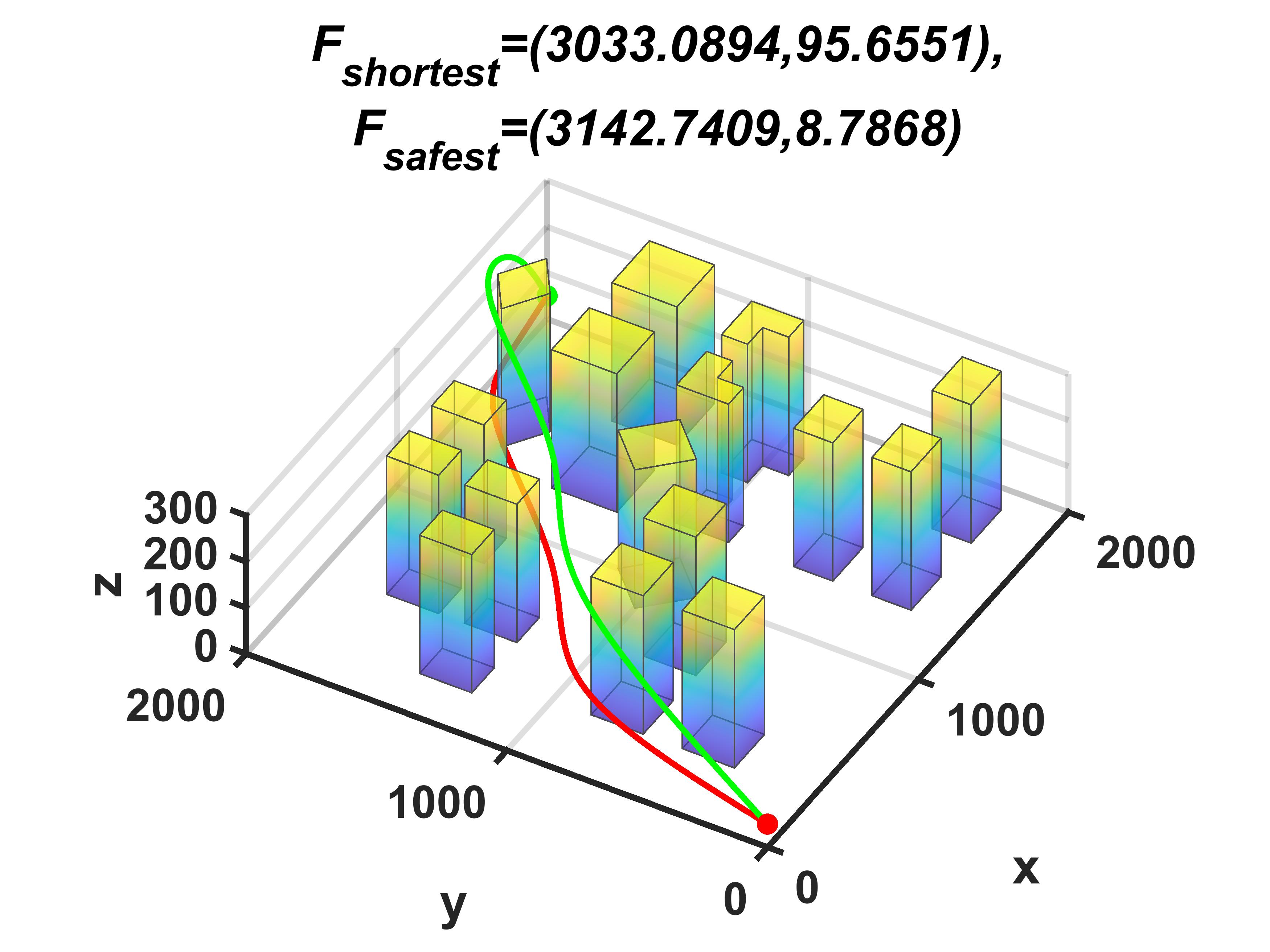}}\centerline{\footnotesize{(c3)}}
	\end{minipage}
     \begin{minipage}{0.24\linewidth}
		\vspace{3pt}\centerline{\includegraphics[width=4.5cm]{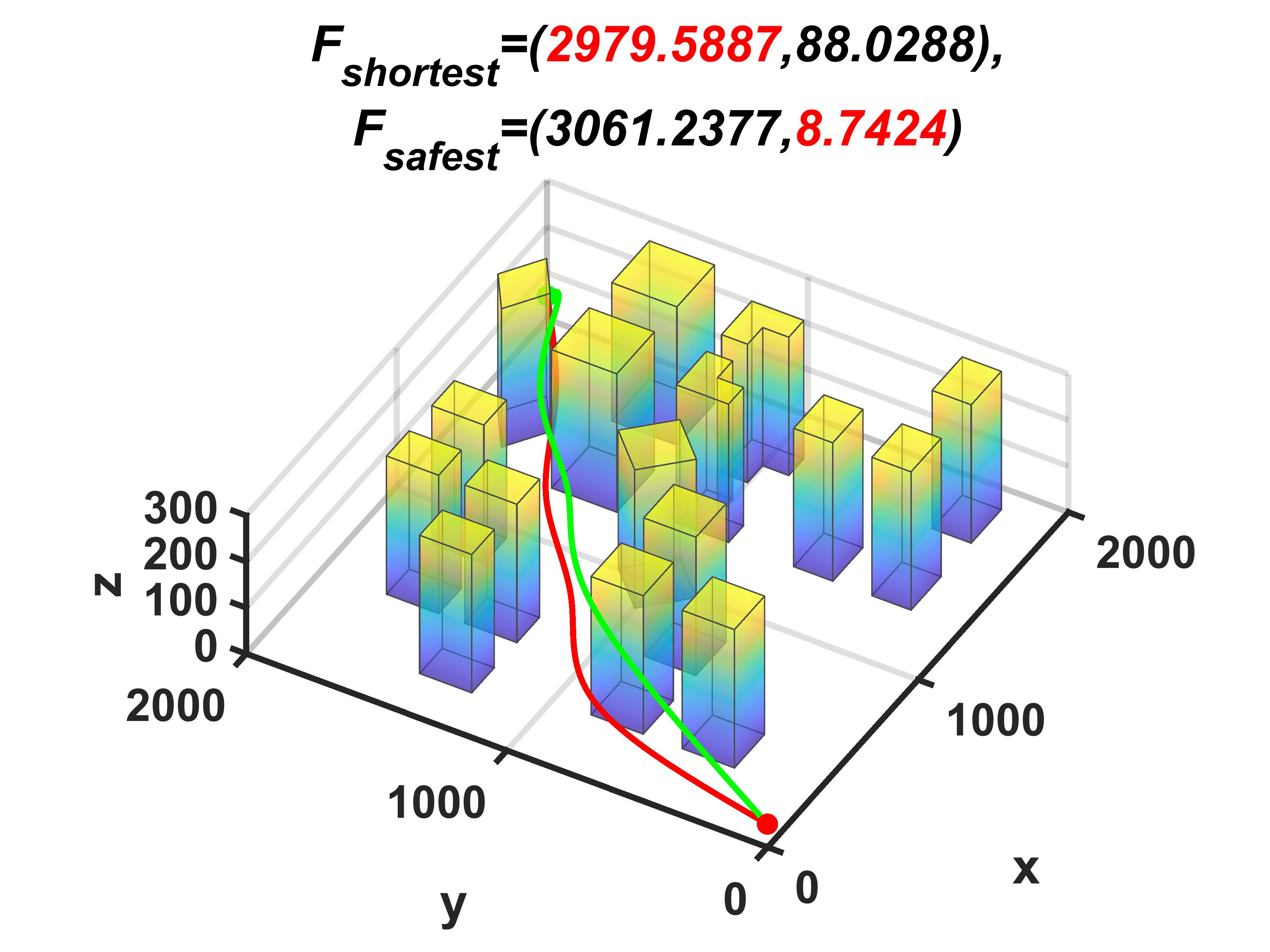}}\centerline{\footnotesize{(c4)}}
	\end{minipage}
 
	\begin{minipage}{0.24\linewidth}
		\vspace{3pt}\centerline{\includegraphics[width=4.5cm]{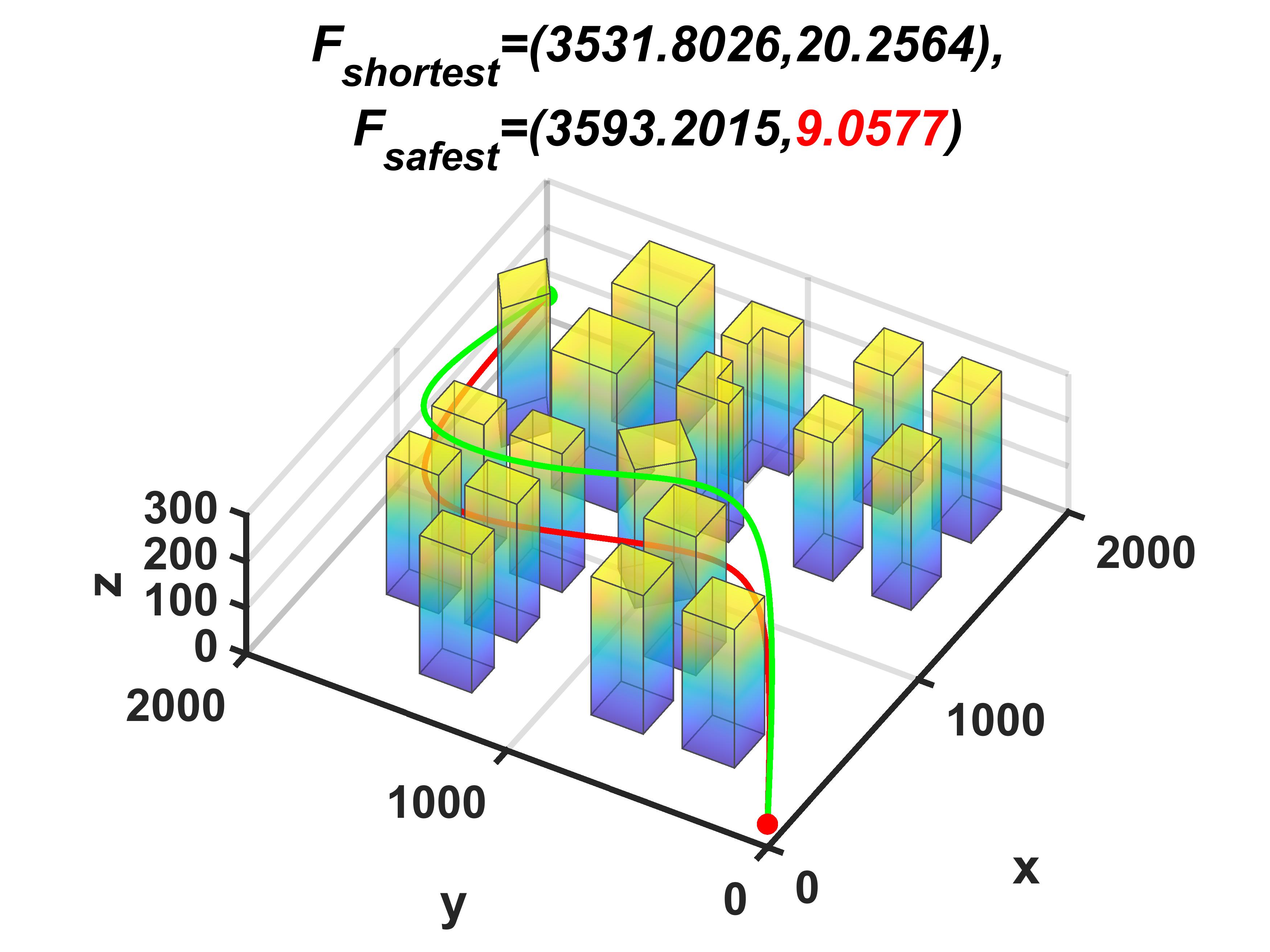}}\centerline{\footnotesize{(d1)}}
	\end{minipage}
    \begin{minipage}{0.24\linewidth}
		\vspace{3pt}\centerline{\includegraphics[width=4.5cm]{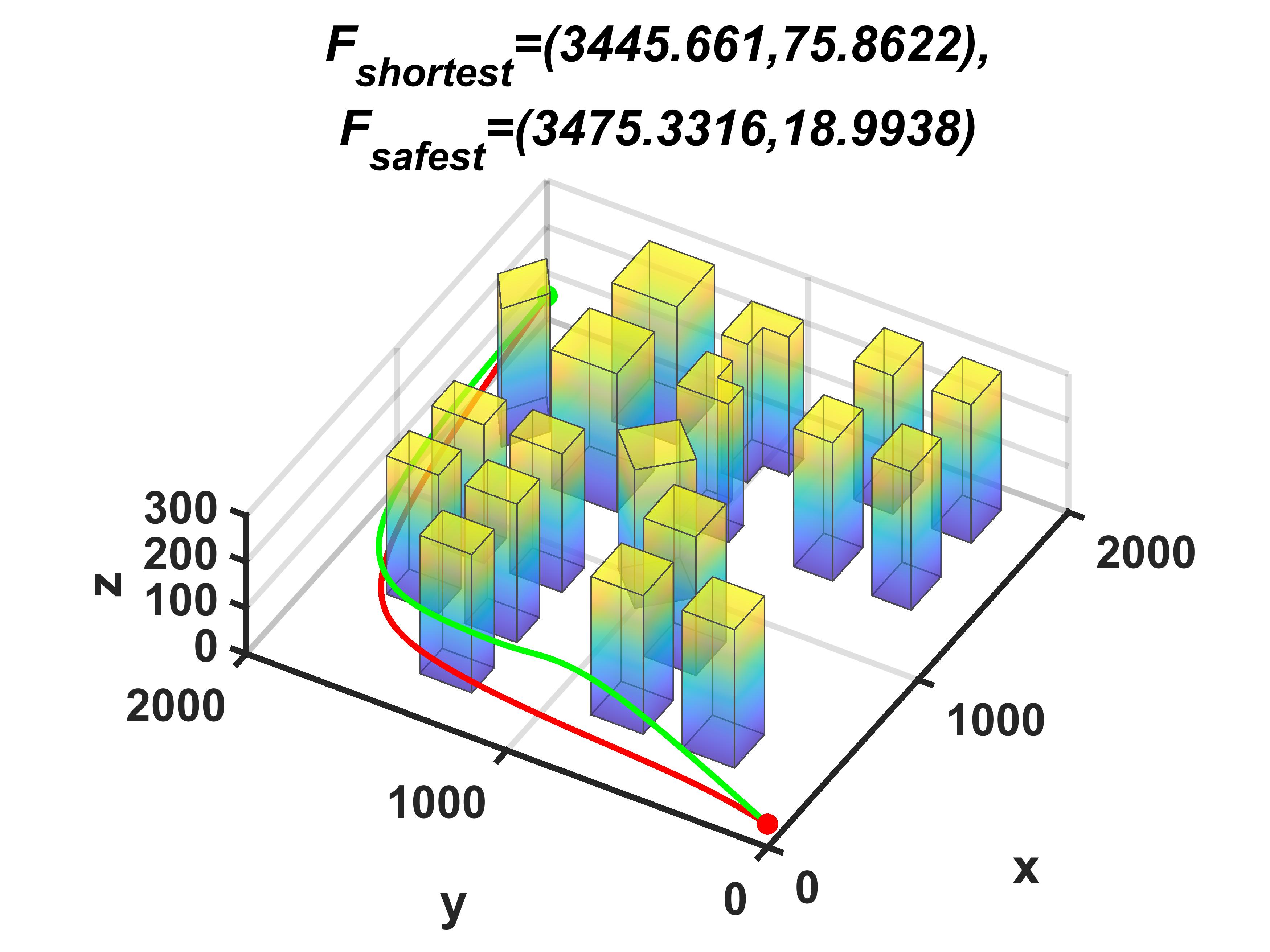}}\centerline{\footnotesize{(d2)}}
	\end{minipage}
    \begin{minipage}{0.24\linewidth}
		\vspace{3pt}\centerline{\includegraphics[width=4.5cm]{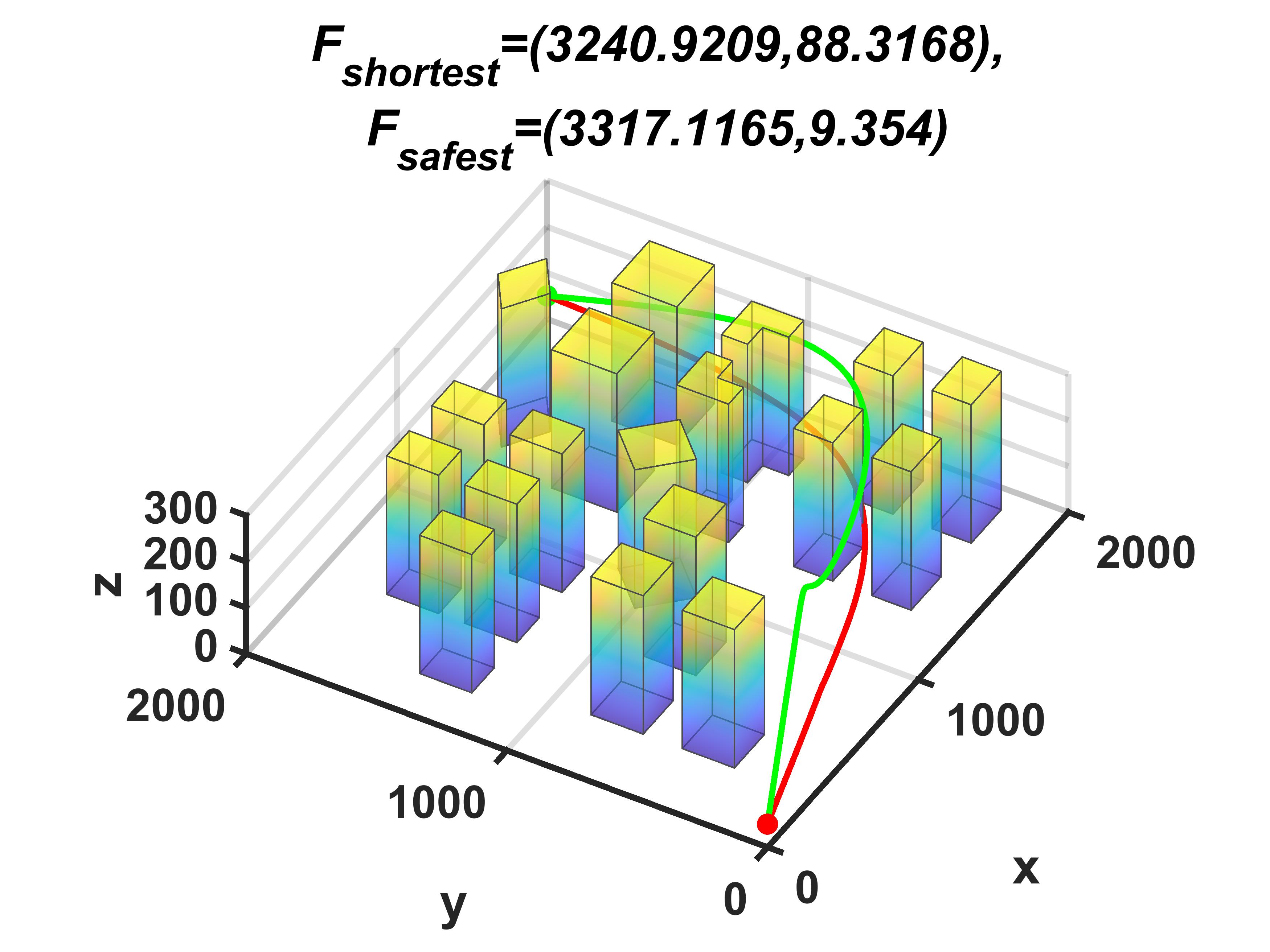}}\centerline{\footnotesize{(d3)}}
	\end{minipage}
    \begin{minipage}{0.24\linewidth}
		\vspace{3pt}\centerline{\includegraphics[width=4.5cm]{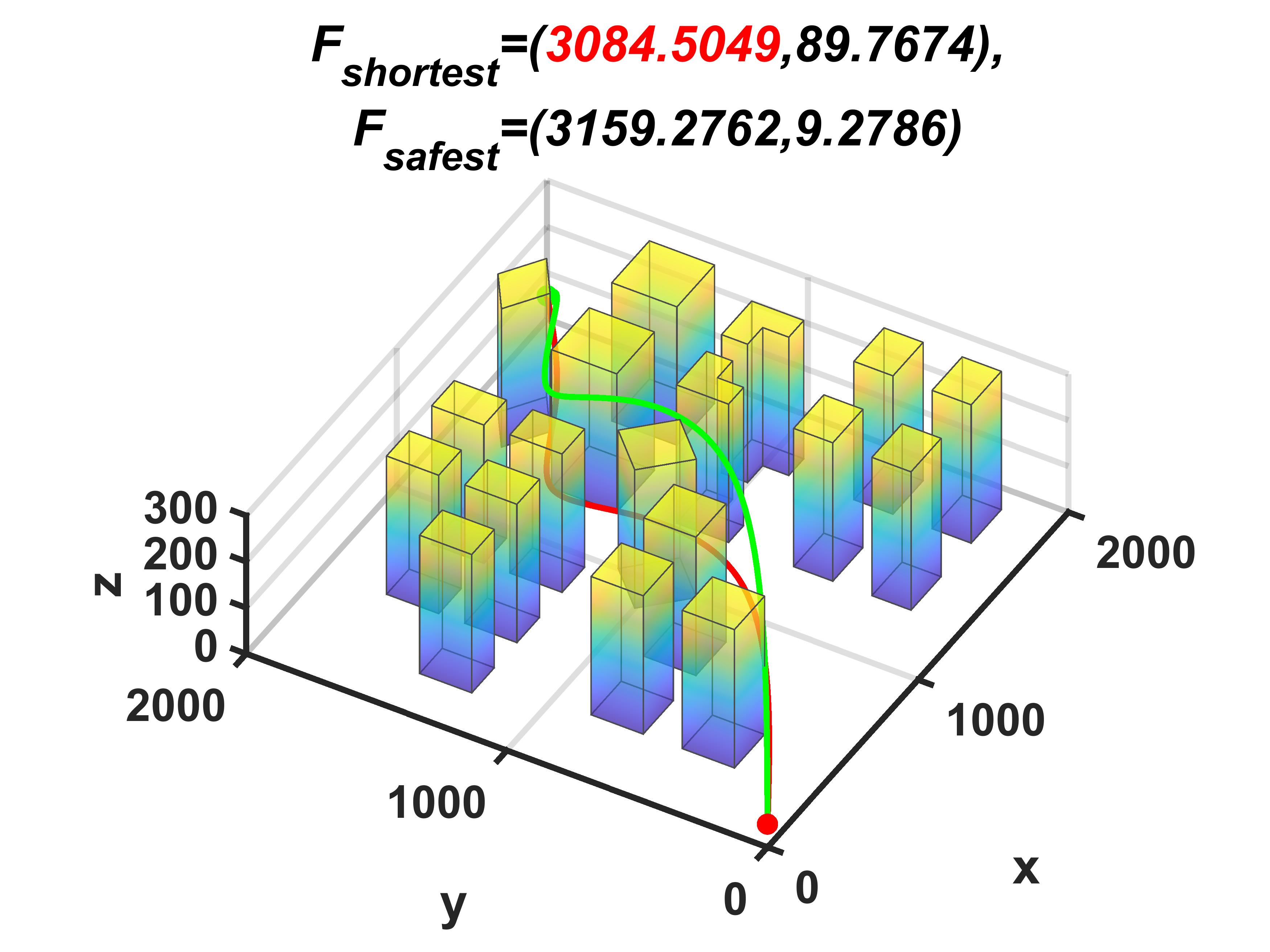}}\centerline{\footnotesize{(d4)}\vspace{1.5ex}}
	\end{minipage}
 
    \caption{The shortest path (red path) and the safest path (green path) in urban environments. (a1)-(a4): C11; (b1)-(b4): C14; (c1)-(c4): C18; (d1)-(d4): C19. (a1)-(d1): NSGA-III; (a2)-(d2): C-MOEA/D; (a3)-(d3): MOEA/D-AWA; (a4)-(d4): MOEA/D-AAWA.}
    \label{fig9}
\end{figure}
Table \ref{tbl3} displays the experimental results of the four MOEAs over 30 independent runs regarding to the mean and Std values of HV and PD in urban environments. Similarly, rank the mean values, and mark the optimal mean value and the highest ranking in boldface. On the one hand, there is little difference of the mean HV values between C-MOEA/D, MOEA/D-AWA and MOEA/D-AAWA in all cases. From the cases C17-C20, the mean HV values of NSGA-III are significantly smaller than that of the other three algorithms. It indicates that NSGA-III is not suitable for handling complex environments with dense urban obstacles. On the other hand, MOEA/D-AWA and MOEA/D-AAWA obtaind better mean values of PD in all cases, which demonstrates the advantages of weight adjustment strategies in diversity. The reason is that weight adjustment strategies guide individuals to evolve towards sparse areas, where there are valuable solutions. In the all cases C11-C20, the proposed algorithm achieves seven optimal mean values of HV (except C15, C19 and C20), seven optimal mean values of PD (except C12, C16 and C20) and the best overall ranking. It indicates that our weight adjustment strategy (i.e., AAWA) is superior to AWA.

Fig. \ref{fig8} presents the final nondominated solutions at the run with the median score in cases C11, C14, C18 and C19. And Fig. \ref{fig9} presents the corresponding shortest path (red path) and safest path (green path). From Fig. \ref{fig8}, the PF also has a sharp peak and a low tail in the urban environments, but it is more gentle than that in the mountain environments. The final nondominated solutions obtained by the proposed algorithm achieves superior diversity. In addition, its convergence is better than other algorithms, especially in cases C18 and C19. By analyzing paths of cases C18 and C19 in Fig. \ref{fig9}, the reason is that the paths of proposed algorithm tend to pass through the middle of the map. 

\subsubsection{Experiments under realistic scenarios}\label{cha5.4.3}

\begin{table}[pos=htb, width=\textwidth]
    \caption{Mean and Std values of HV and PD in realistic scenarios}
    \label{tbl4}
    \begin{tabular*}{\tblwidth}{@{}LLLLLLL@{}}
        \toprule
        \multirow{2}{*}[-2pt]{Case}&\multirow{2}{*}[-2pt]{Metric}&\multicolumn{4}{c}{Algorithm}\\
		\cmidrule(lr){3-6}
		&&NSGA-III&C-MOEA/D&MOEA/D-AWA&MOEA/D-AAWA\\\midrule
  
        \multirow{2}{*}[-2pt]{C21}
        &HV &0.3184±0.4581 (4)&0.4612±0.3573 (3)&0.5164±0.4922 (2)&\textbf{0.62}±0.4799 \textbf{(1)}\\
        &PD($\times 10^{4}$) &0.231±0.3388 (3)&0.2066±0.1697 (4)&0.785±0.8889 (2)&\textbf{0.8378}±0.7473 \textbf{(1)}\\
        \multirow{2}{*}[-2pt]{C22}
        &HV &0.7328±0.4113 (4)&0.7566±0.0249 (3)&\textbf{0.9771}±0.0113 \textbf{(1)}&0.972±0.0136 (2)\\
        &PD($\times 10^{4}$) &0.4866±0.2941 (3)&0.4145±0.1057 (4)&\textbf{1.2474}±0.2983 \textbf{(1)}&1.1769±0.4781 (2)\\
        \multirow{2}{*}[-2pt]{C23}
        &HV &0.7557±0.3846 (4)&0.7628±0.0718 (3) (3)&0.9053±0.2466 (2)&\textbf{0.968}±0.0209 \textbf{(1)}\\
        &PD($\times 10^{4}$) &0.4976±0.277 (3)&0.4591±0.113 (4)&0.9053±0.2466 (2)&\textbf{1.4611}±0.6646 \textbf{(1)}\\
        \multirow{2}{*}[-2pt]{C24}
        &HV &0.5749±0.4463 (4)&0.7091±0.0219 (3)&0.9775±0.0084 (2)&\textbf{0.9798}±0.0064 \textbf{(1)}\\
        &PD($\times 10^{4}$) &0.2828±0.2281 (3)&0.2422±0.059 (4)&1.027±0.3195 (2)&\textbf{1.0444}±0.3038 \textbf{(1)}\\\midrule

        \multicolumn{2}{c}{Overall ranking}&(28)&(28)&(14)&\textbf{(10)}\\
        \bottomrule
    \end{tabular*}
\end{table}
\begin{figure}[pos=htb]
    \centering
        \subfigure[]{\includegraphics[width=4cm]{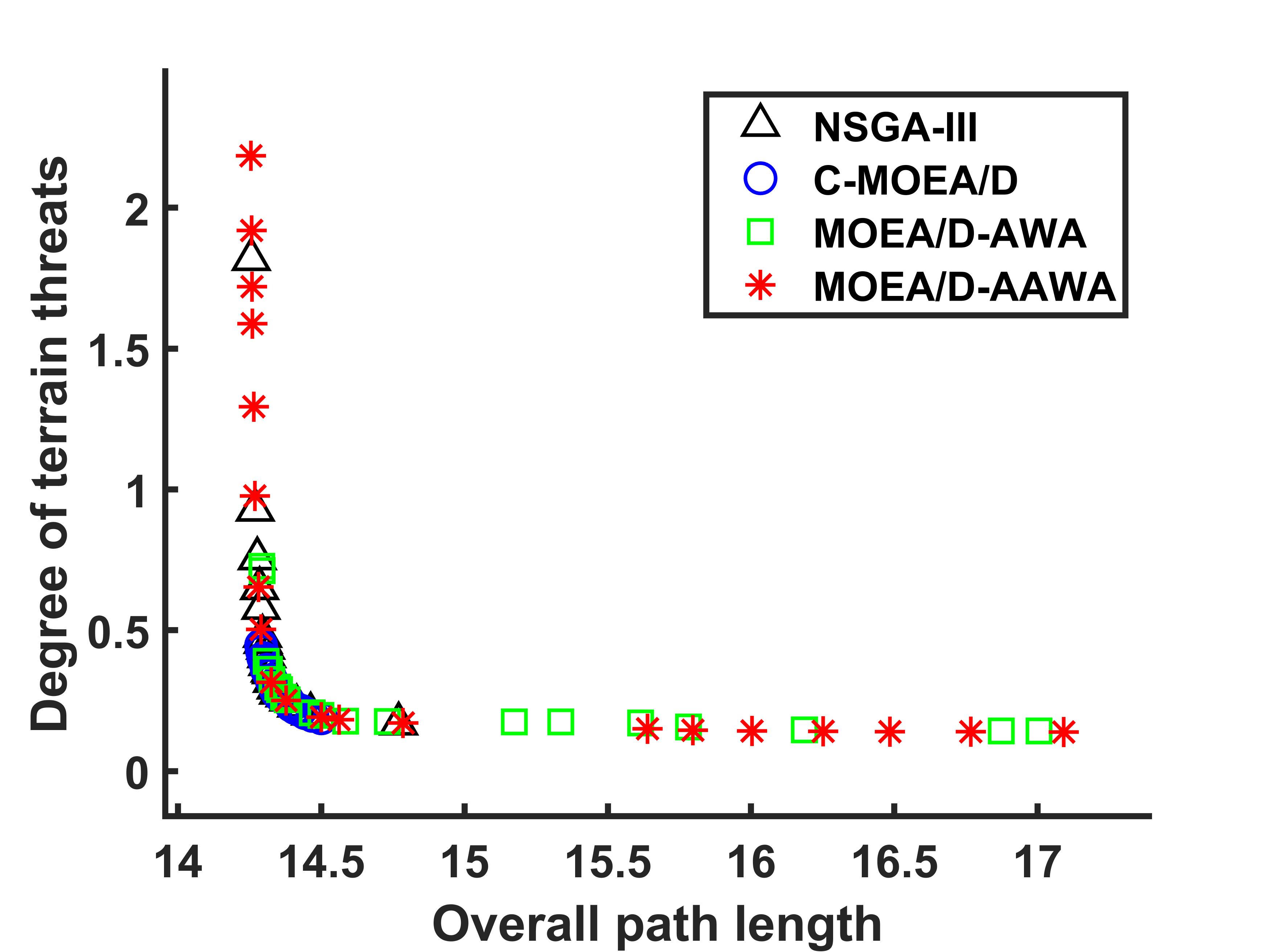}}
        \subfigure[]{\includegraphics[width=4cm]{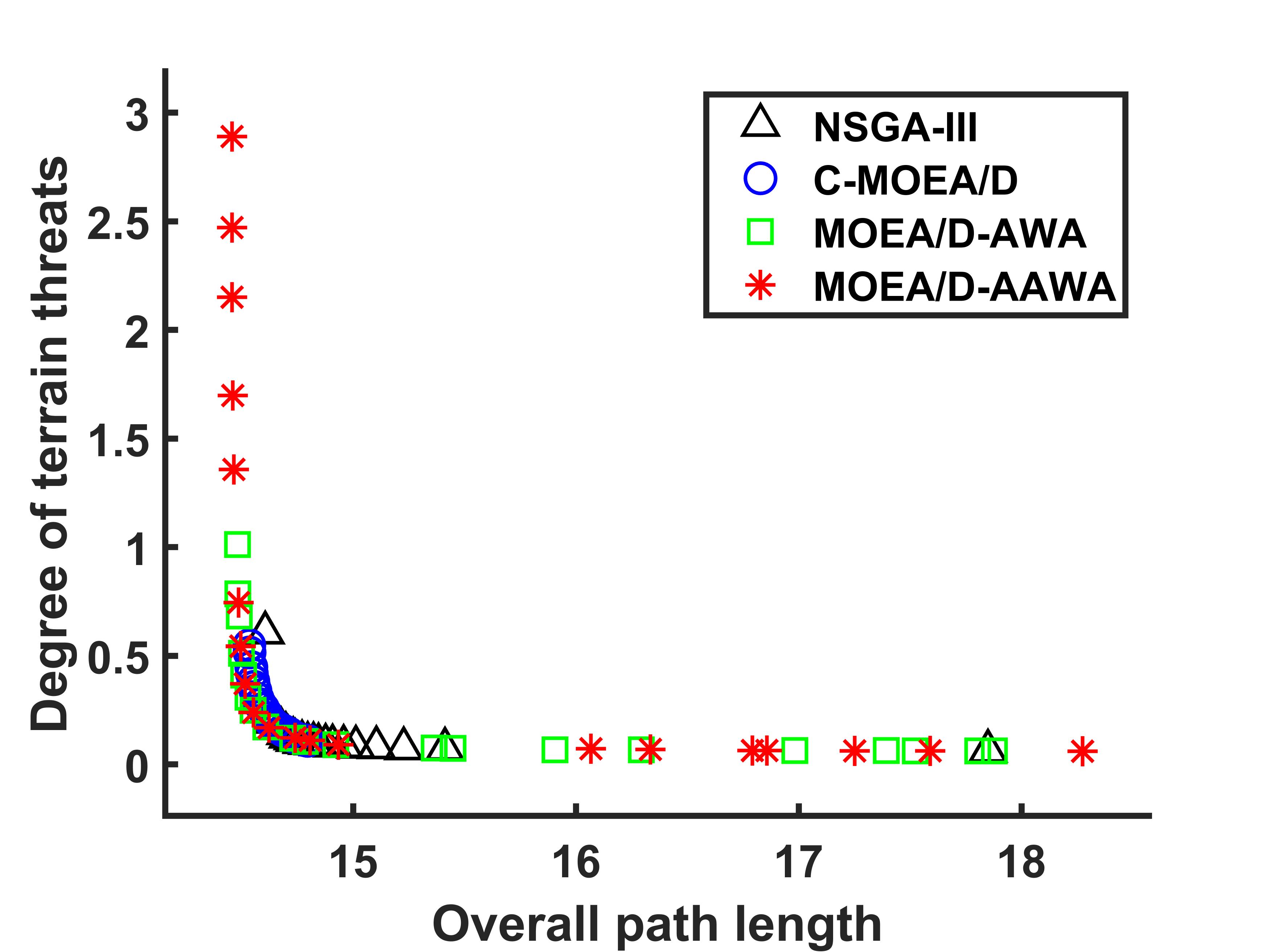}}
        \subfigure[]{\includegraphics[width=4cm]{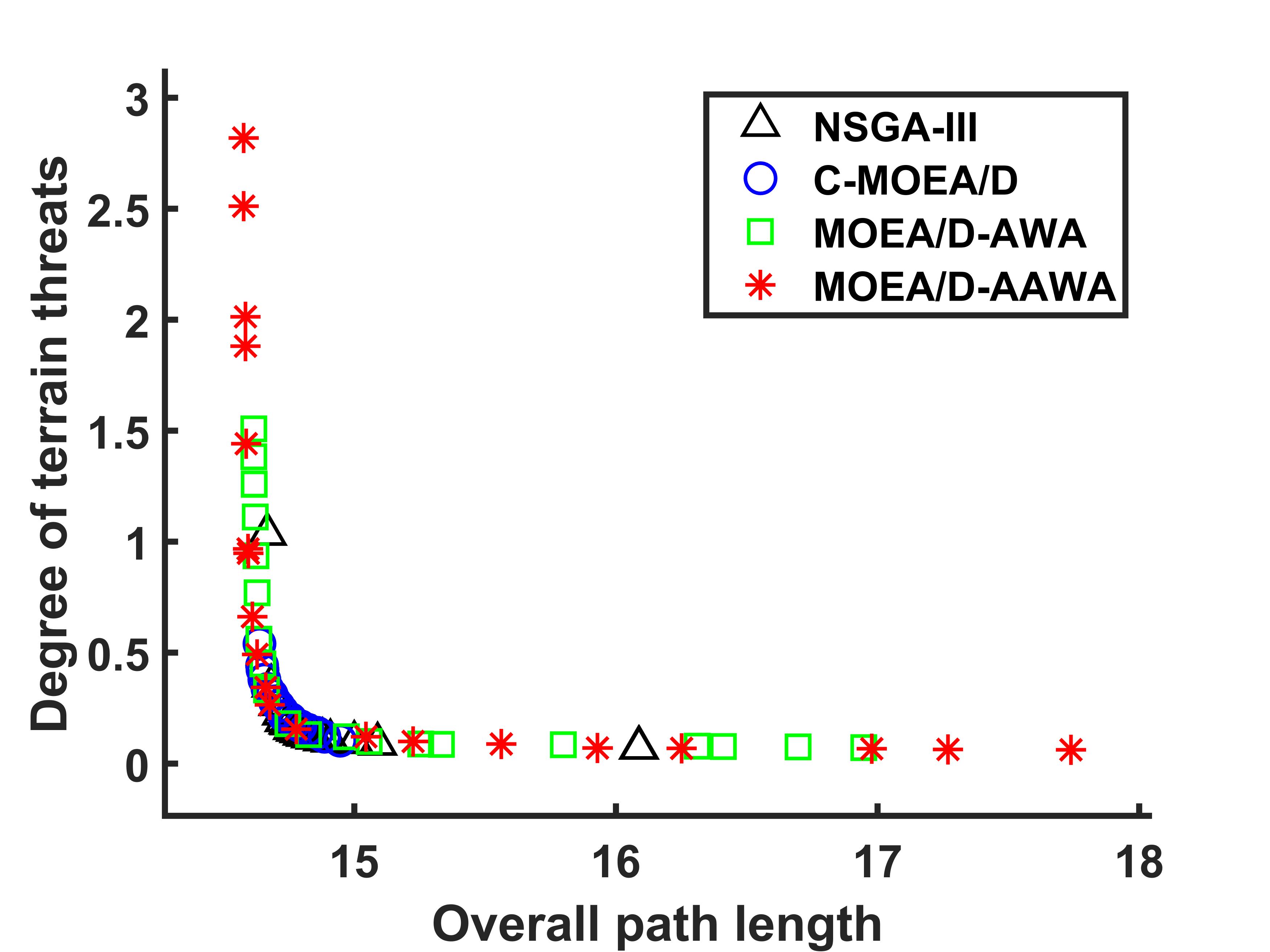}}
        \subfigure[]{\includegraphics[width=4cm]{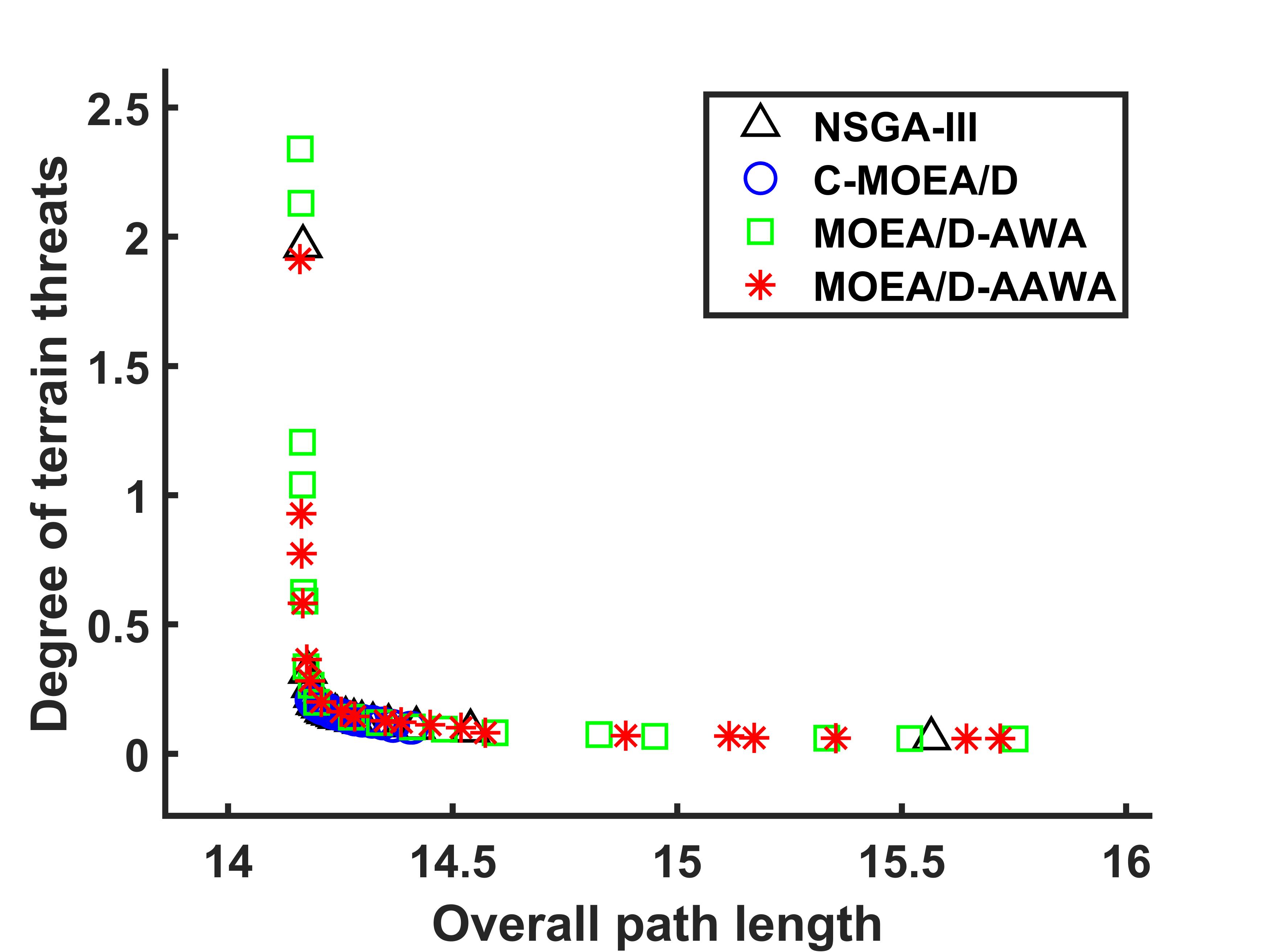}}
    \caption{Final nondominated solutions in realistic scenarios. (a): C21; (b): C22; (c): C23; (d): C24.}
    \label{fig10}
\end{figure}
\begin{figure}[pos=htb]
	\begin{minipage}{0.24\linewidth}
		\vspace{3pt}\centerline{\includegraphics[width=4.5cm]{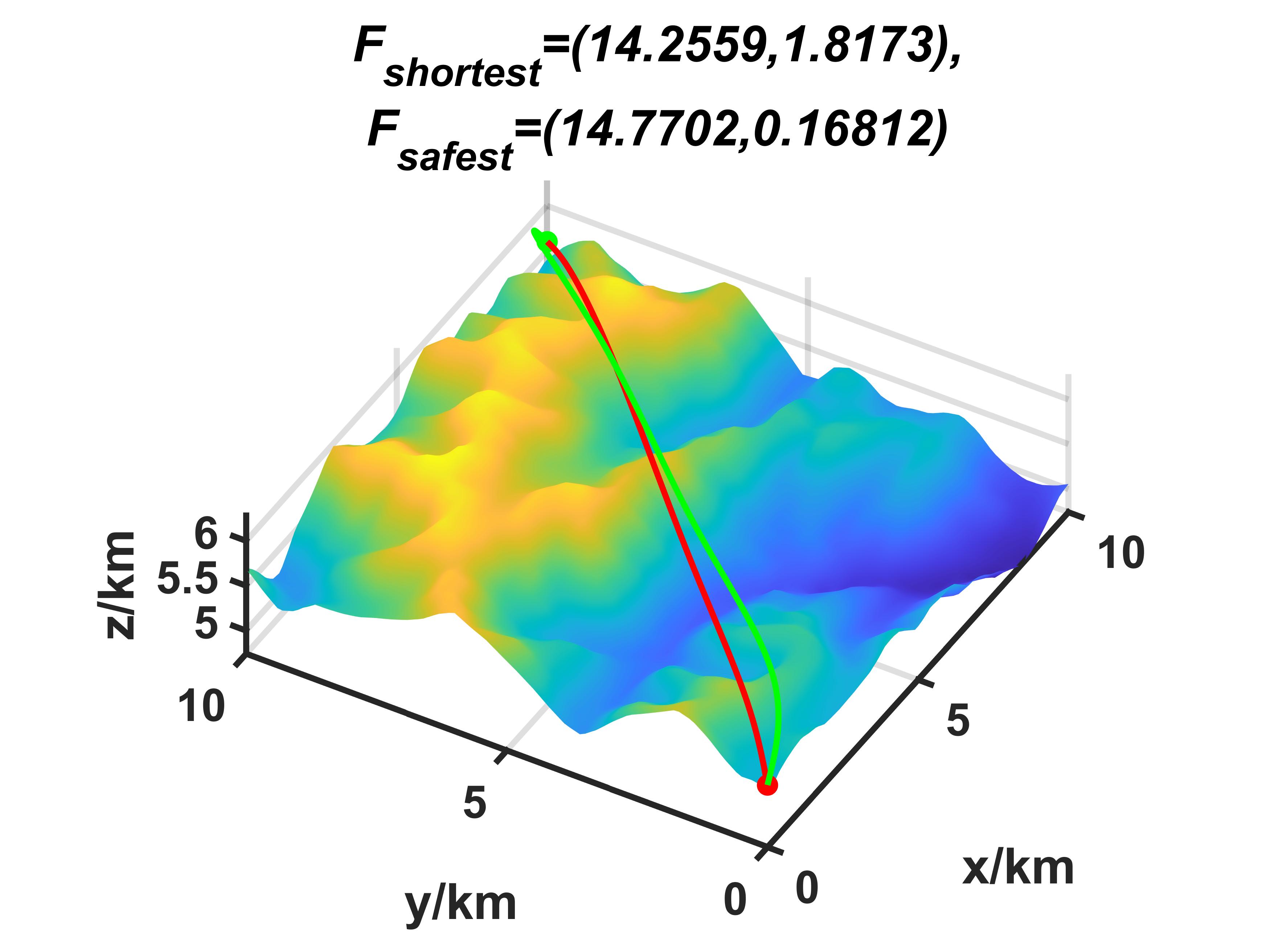}}\centerline{\footnotesize{(a1)}}
	\end{minipage}
    \begin{minipage}{0.24\linewidth}
		\vspace{3pt}\centerline{\includegraphics[width=4.5cm]{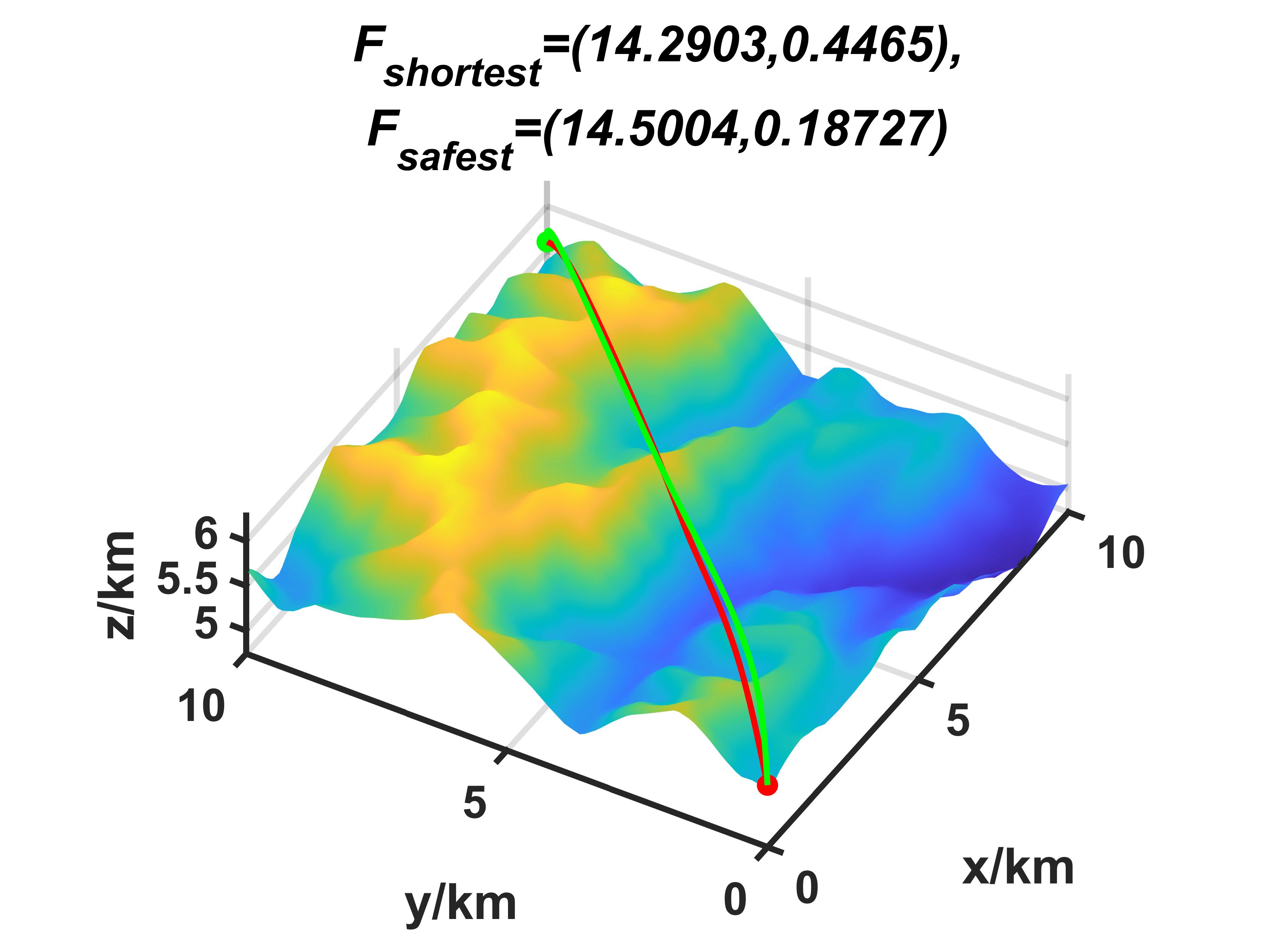}}\centerline{\footnotesize{(a2)}}
	\end{minipage}
    \begin{minipage}{0.24\linewidth}
		\vspace{3pt}\centerline{\includegraphics[width=4.5cm]{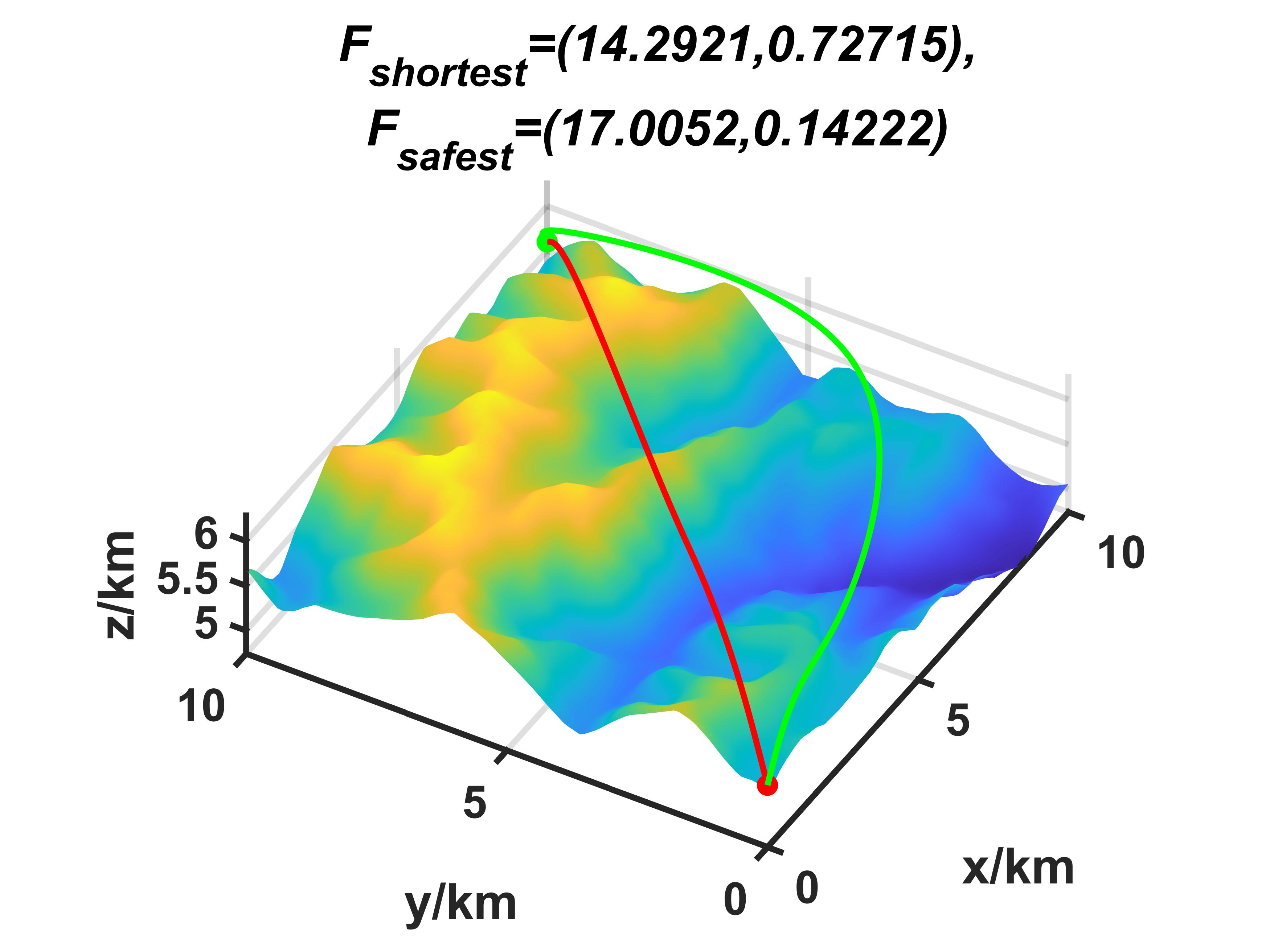}}\centerline{\footnotesize{(a3)}}
	\end{minipage}
    \begin{minipage}{0.24\linewidth}
		\vspace{3pt}\centerline{\includegraphics[width=4.5cm]{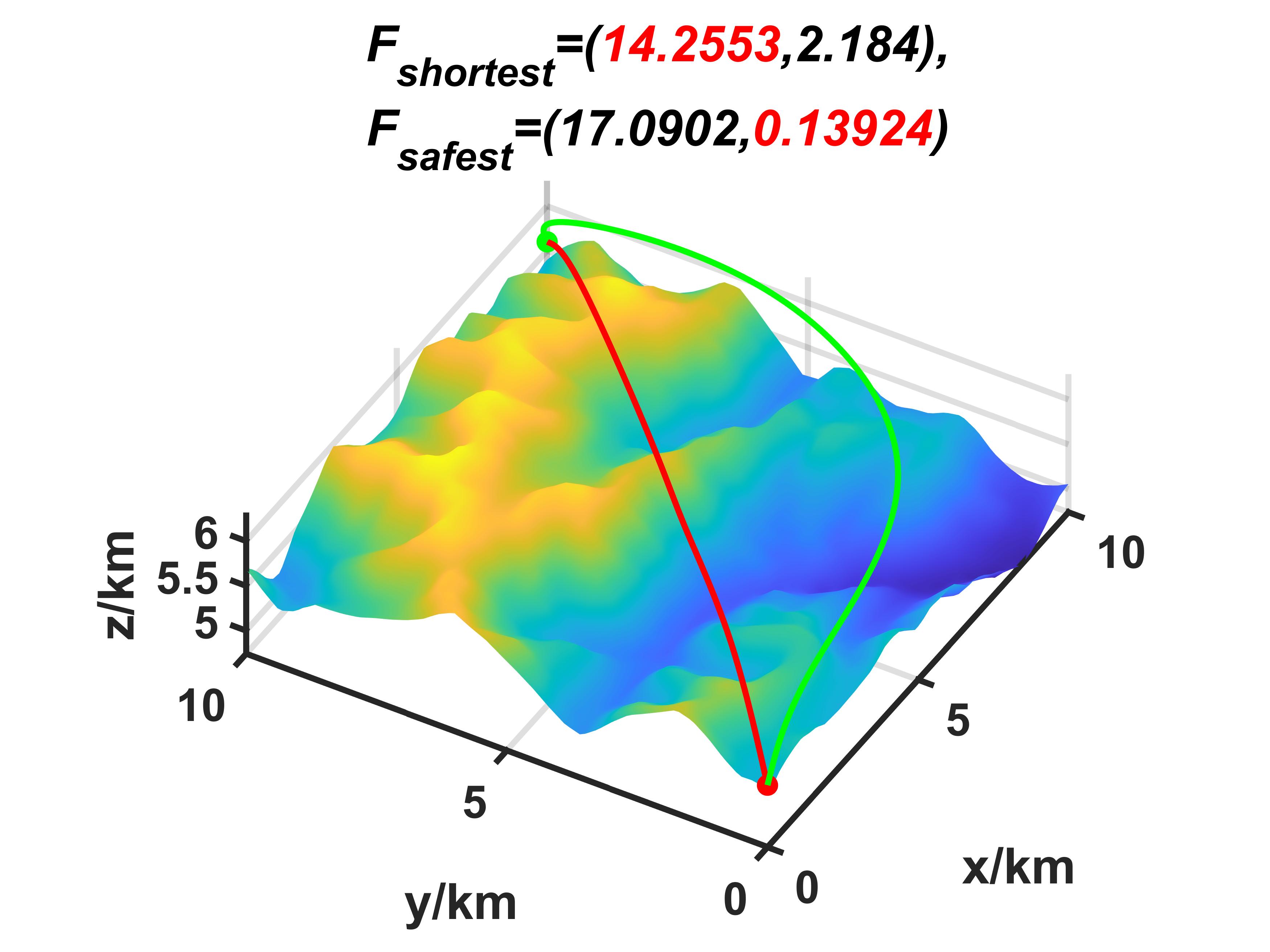}}\centerline{\footnotesize{(a4)}}
	\end{minipage}
 
    \begin{minipage}{0.24\linewidth}
		\vspace{3pt}\centerline{\includegraphics[width=4.5cm]{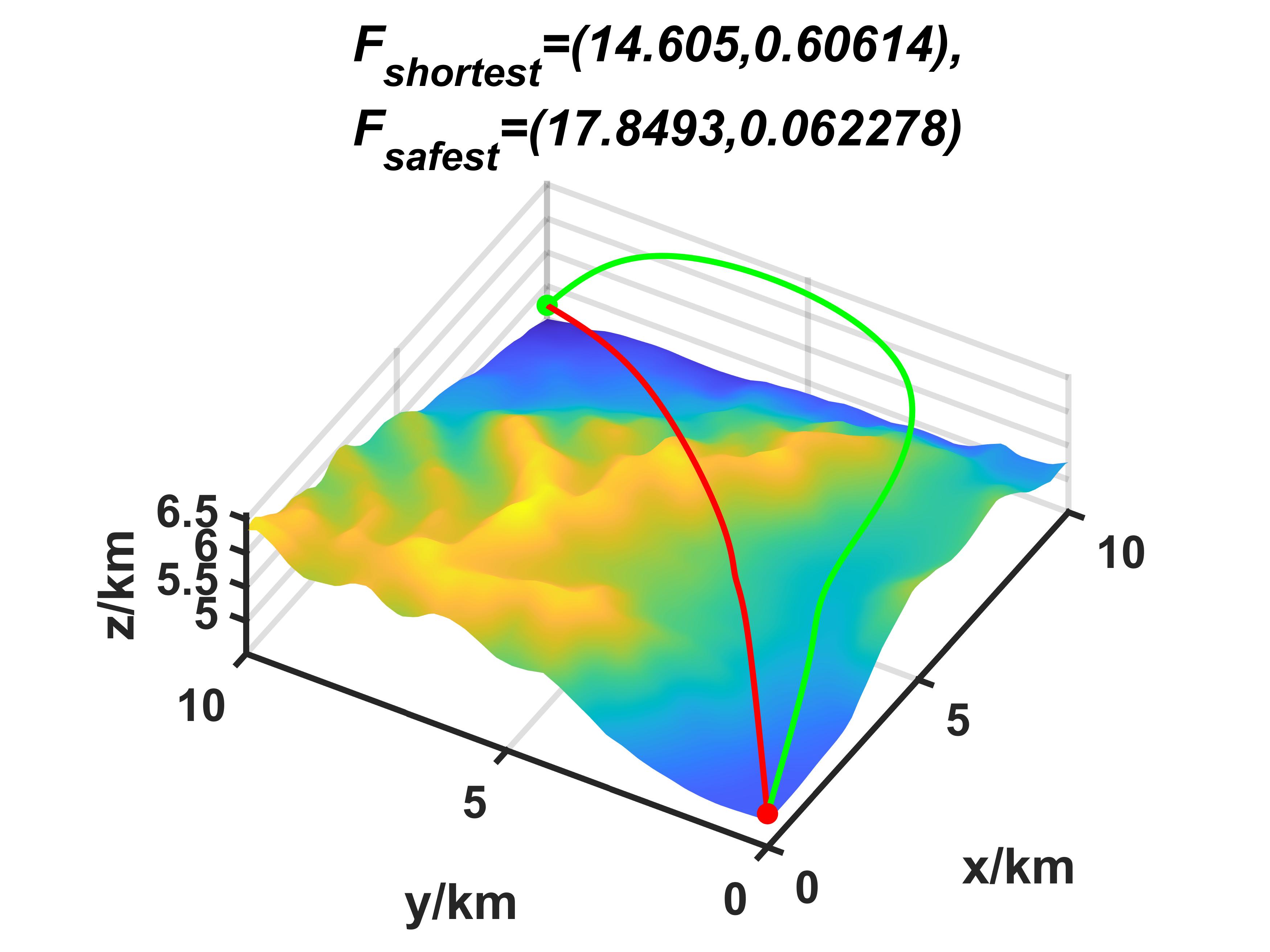}}\centerline{\footnotesize{(b1)}}
	\end{minipage}
     \begin{minipage}{0.24\linewidth}
		\vspace{3pt}\centerline{\includegraphics[width=4.5cm]{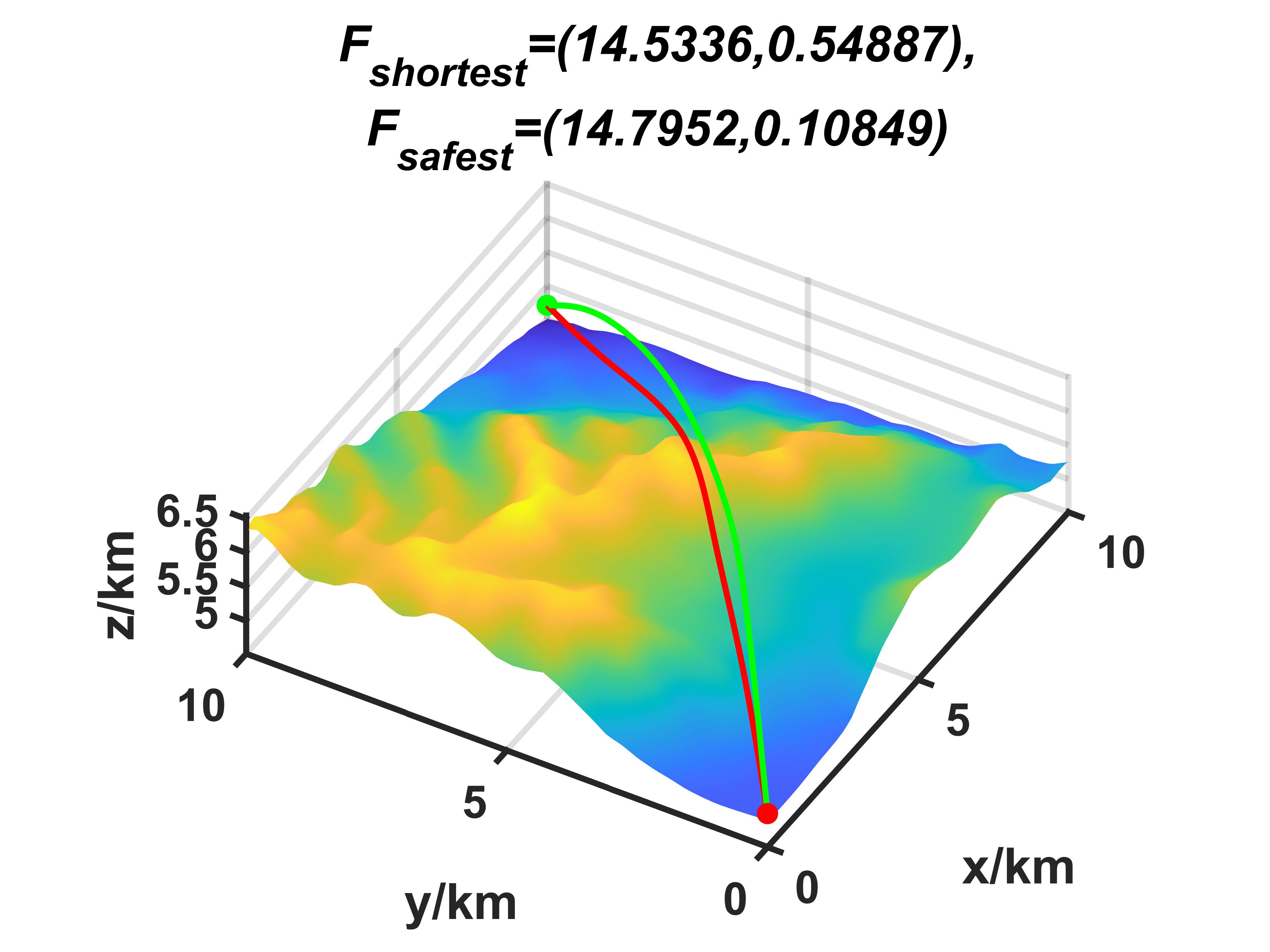}}\centerline{\footnotesize{(b2)}}
	\end{minipage}
     \begin{minipage}{0.24\linewidth}
		\vspace{3pt}\centerline{\includegraphics[width=4.5cm]{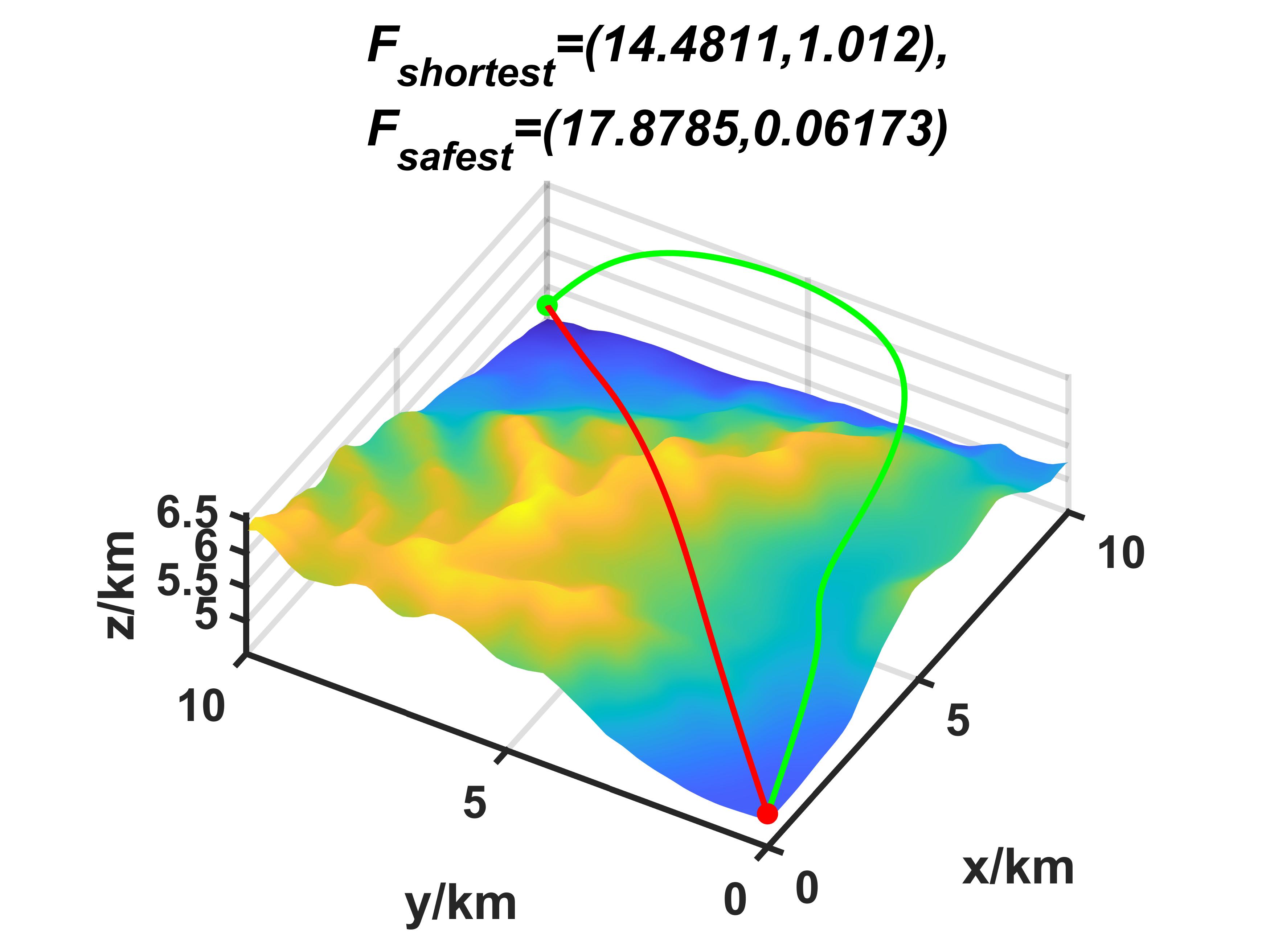}}\centerline{\footnotesize{(b3)}}
	\end{minipage}
     \begin{minipage}{0.24\linewidth}
		\vspace{3pt}\centerline{\includegraphics[width=4.5cm]{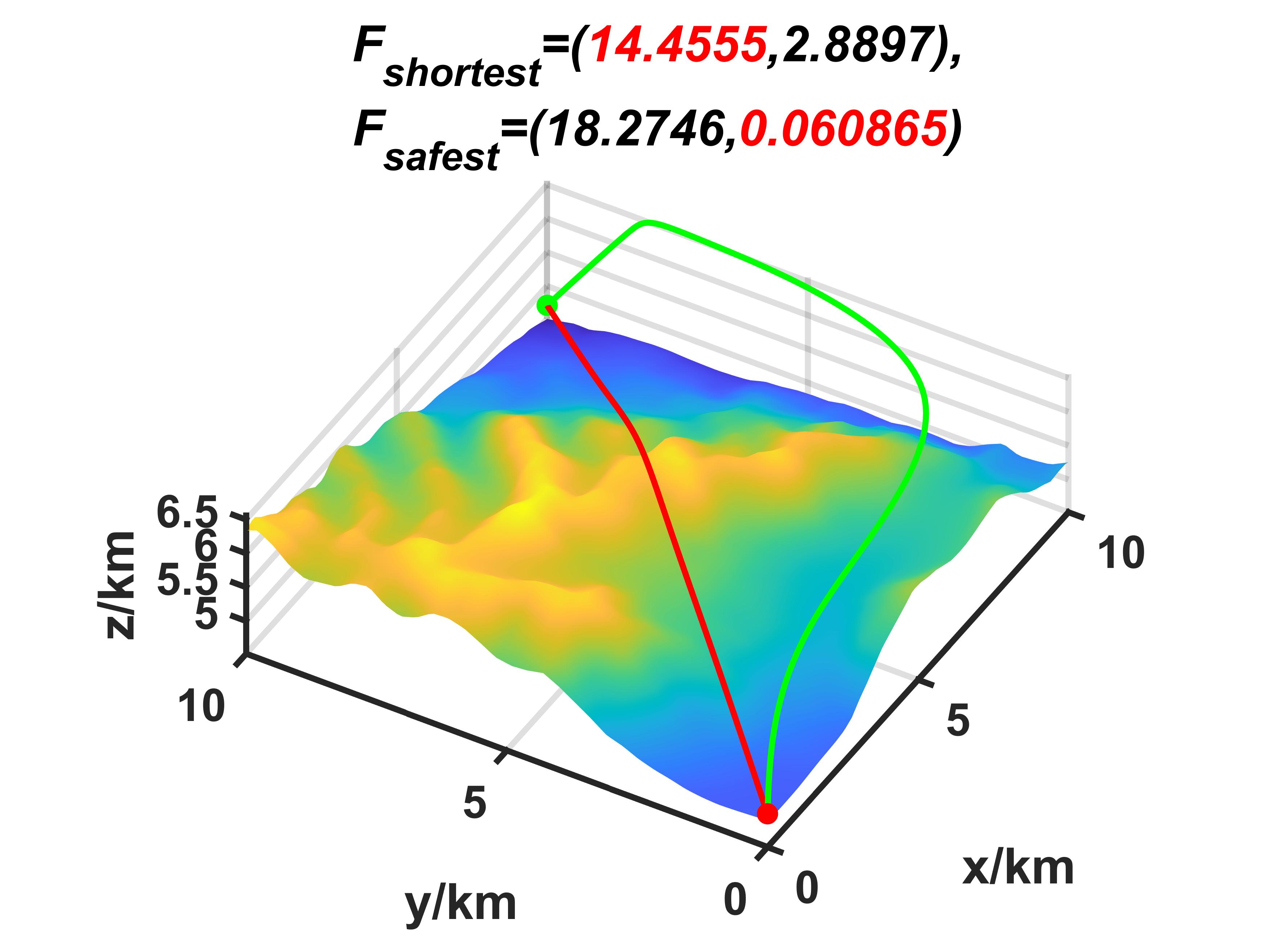}}\centerline{\footnotesize{(b4)}}
	\end{minipage}
 
    \begin{minipage}{0.24\linewidth}
		\vspace{3pt}\centerline{\includegraphics[width=4.5cm]{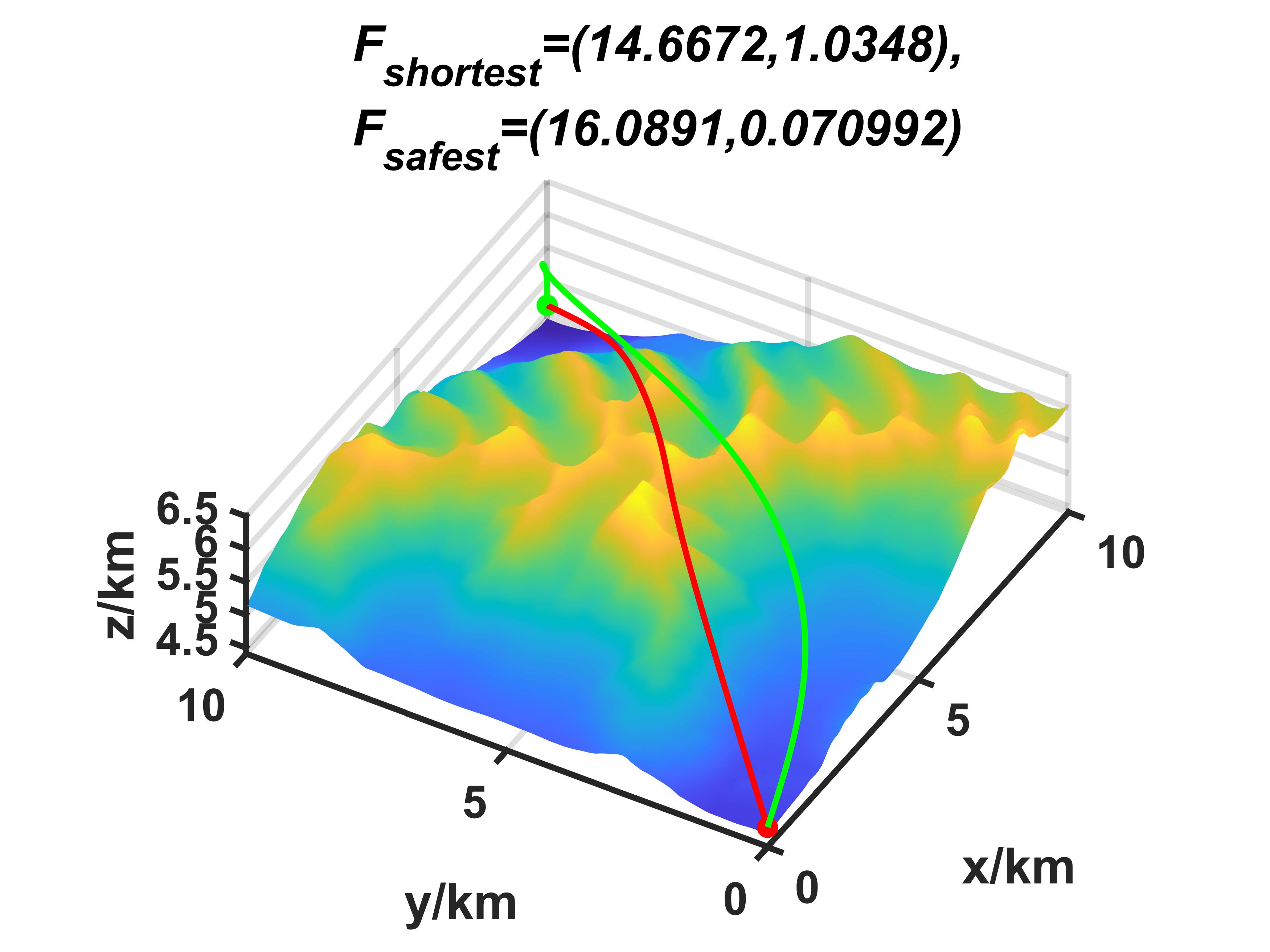}}\centerline{\footnotesize{(c1)}}
	\end{minipage}
     \begin{minipage}{0.24\linewidth}
		\vspace{3pt}\centerline{\includegraphics[width=4.5cm]{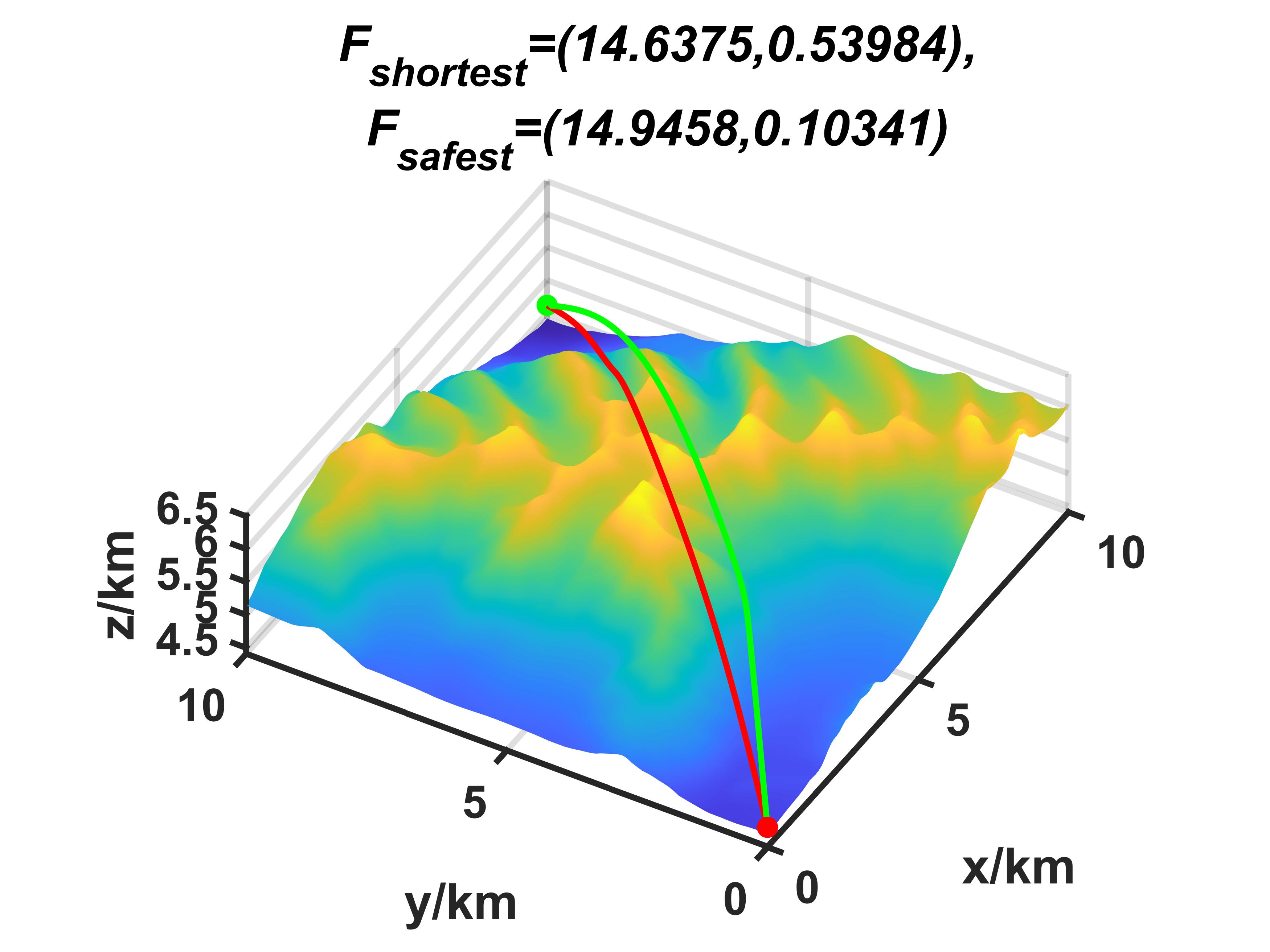}}\centerline{\footnotesize{(c2)}}
	\end{minipage}
     \begin{minipage}{0.24\linewidth}
		\vspace{3pt}\centerline{\includegraphics[width=4.5cm]{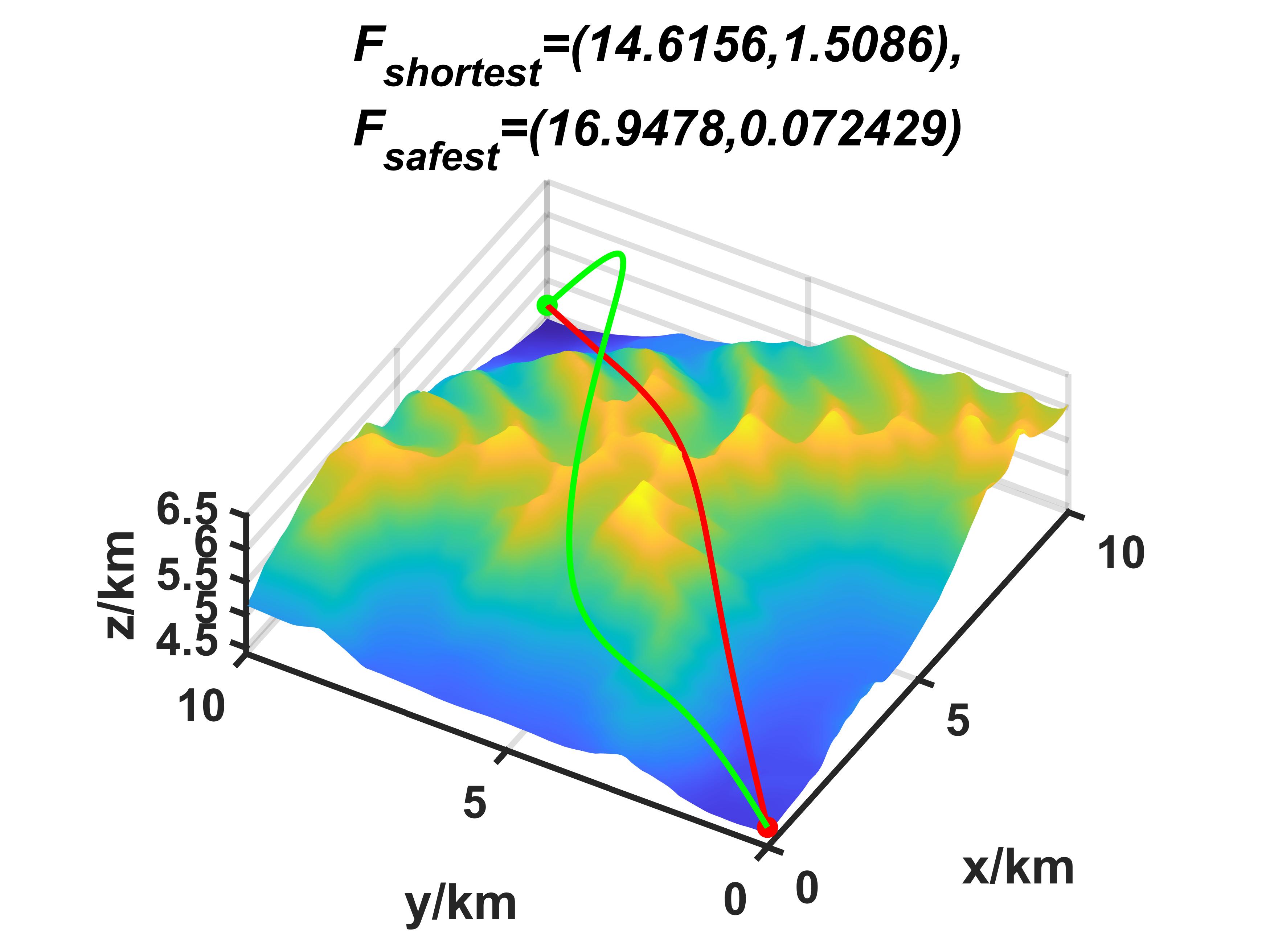}}\centerline{\footnotesize{(c3)}}
	\end{minipage}
     \begin{minipage}{0.24\linewidth}
		\vspace{3pt}\centerline{\includegraphics[width=4.5cm]{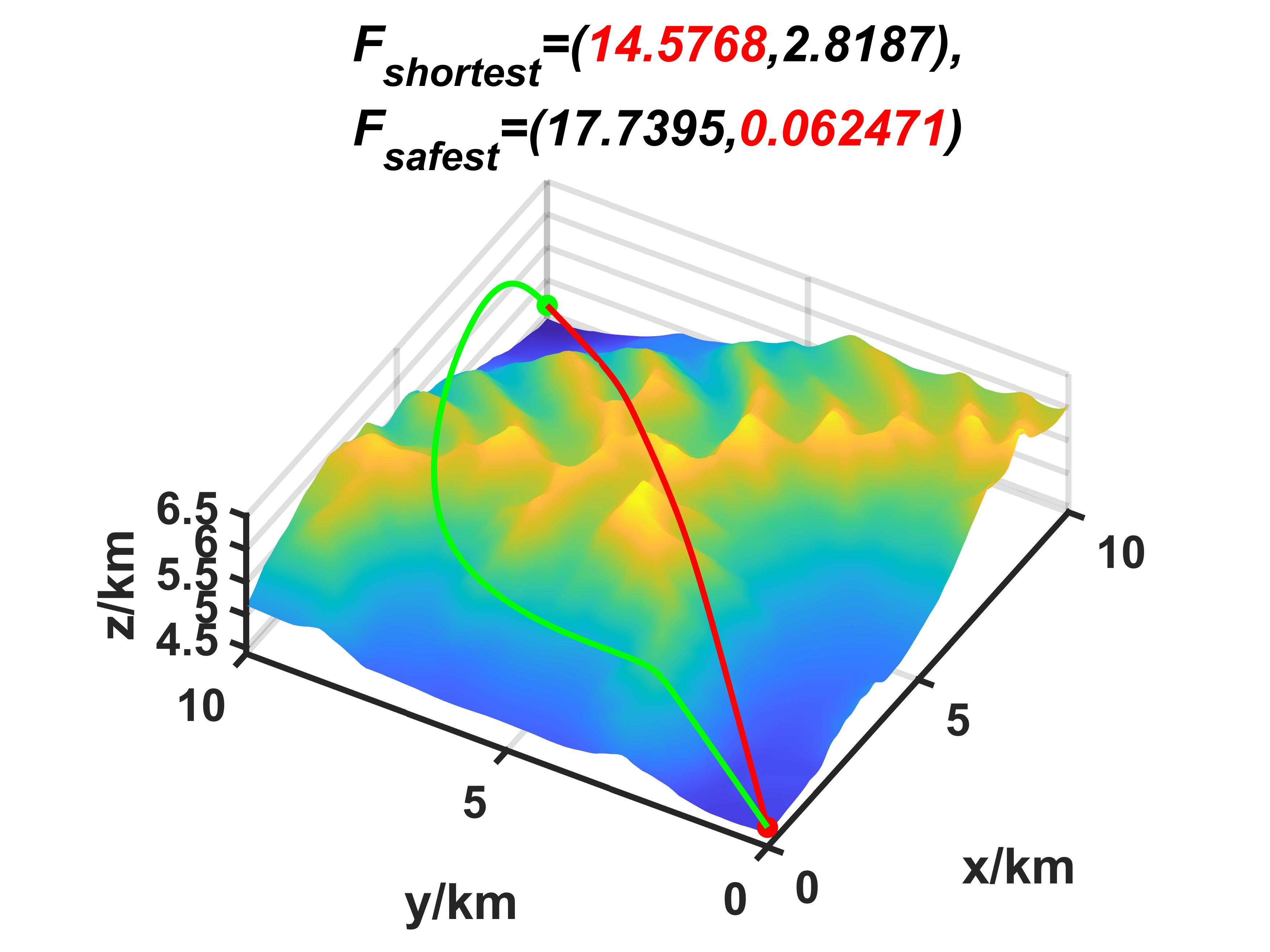}}\centerline{\footnotesize{(c4)}}
	\end{minipage}
 
	\begin{minipage}{0.24\linewidth}
		\vspace{3pt}\centerline{\includegraphics[width=4.5cm]{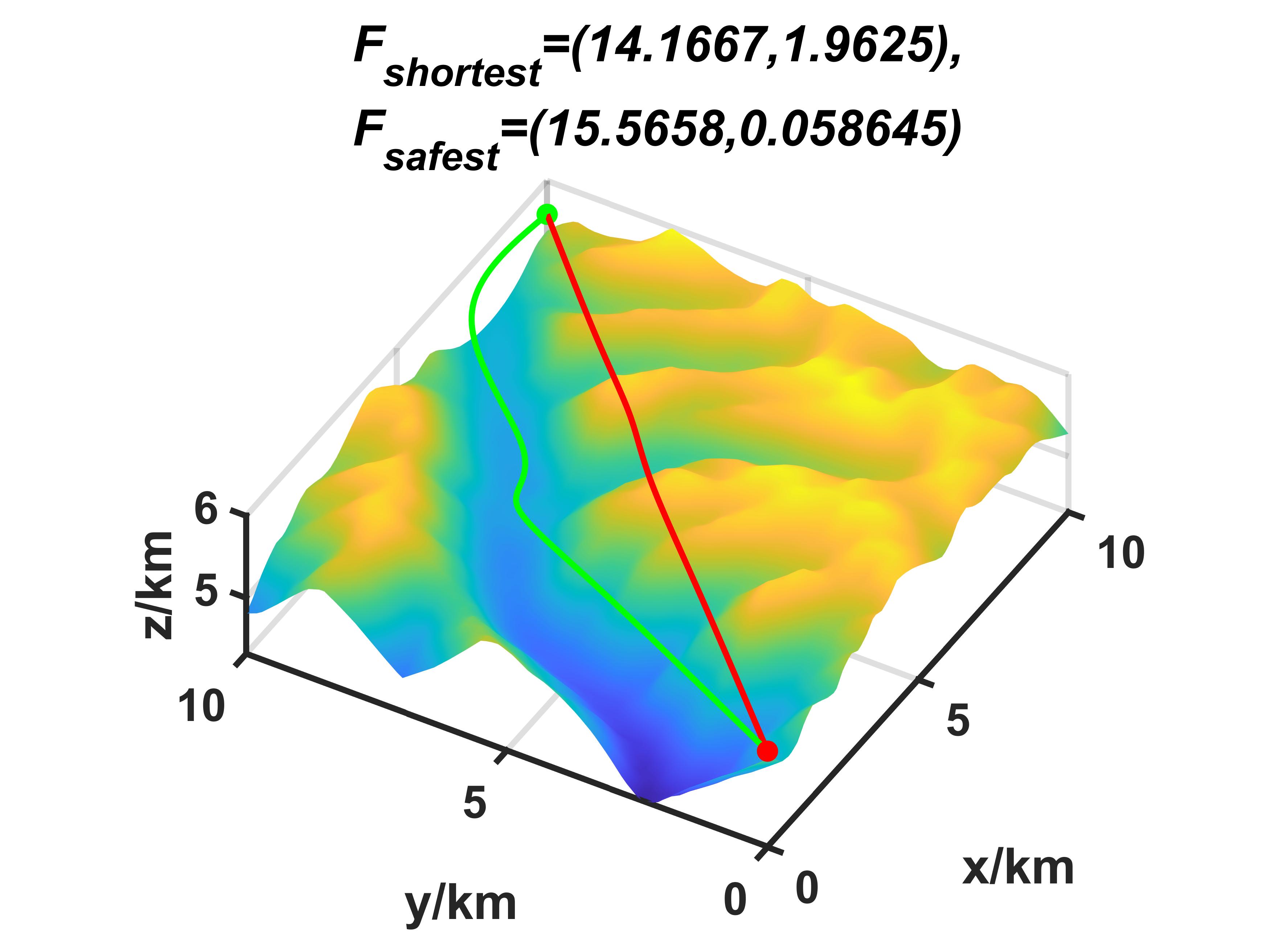}}\centerline{\footnotesize{(d1)}}
	\end{minipage}
    \begin{minipage}{0.24\linewidth}
		\vspace{3pt}\centerline{\includegraphics[width=4.5cm]{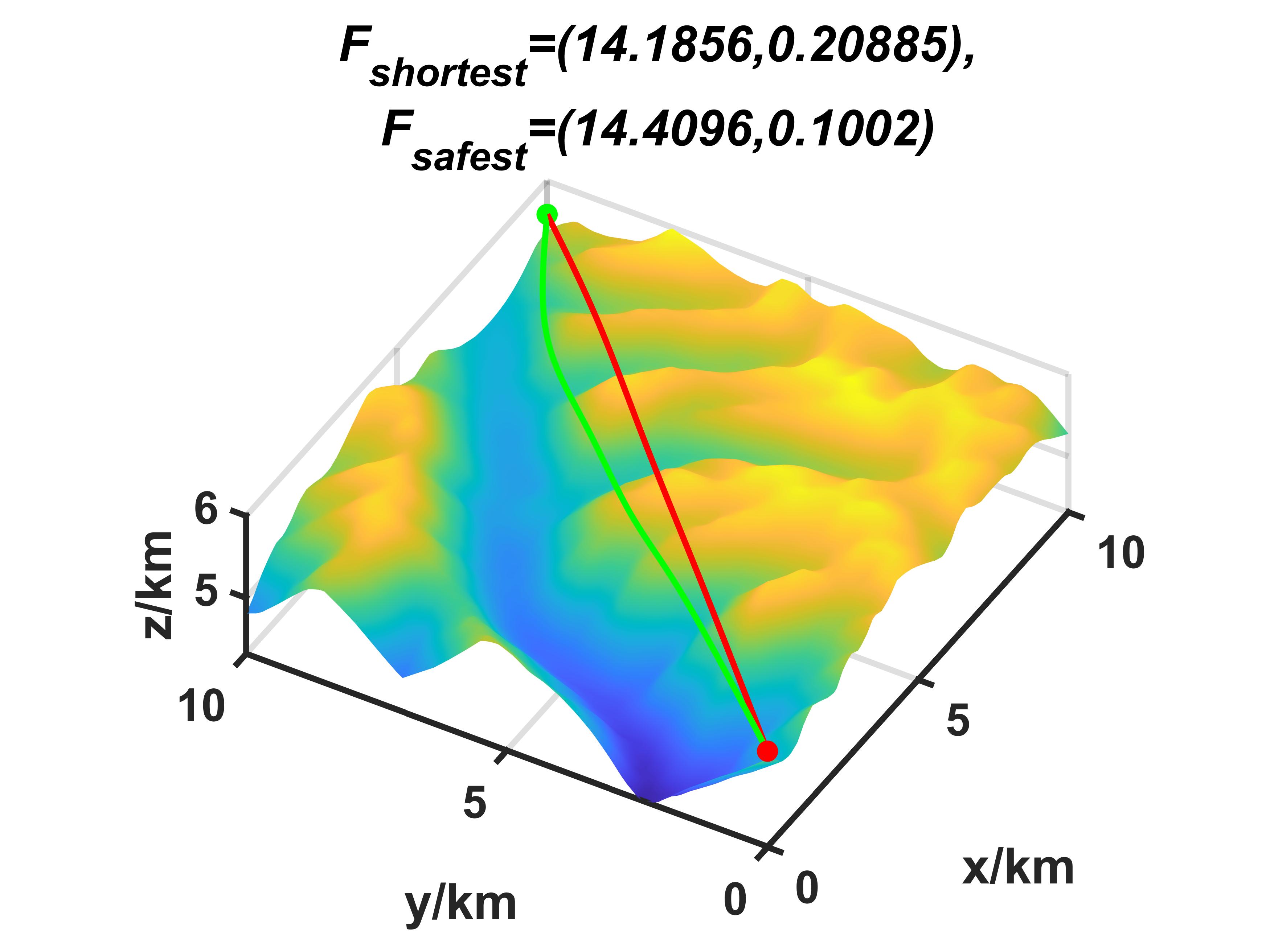}}\centerline{\footnotesize{(d2)}}
	\end{minipage}
    \begin{minipage}{0.24\linewidth}
		\vspace{3pt}\centerline{\includegraphics[width=4.5cm]{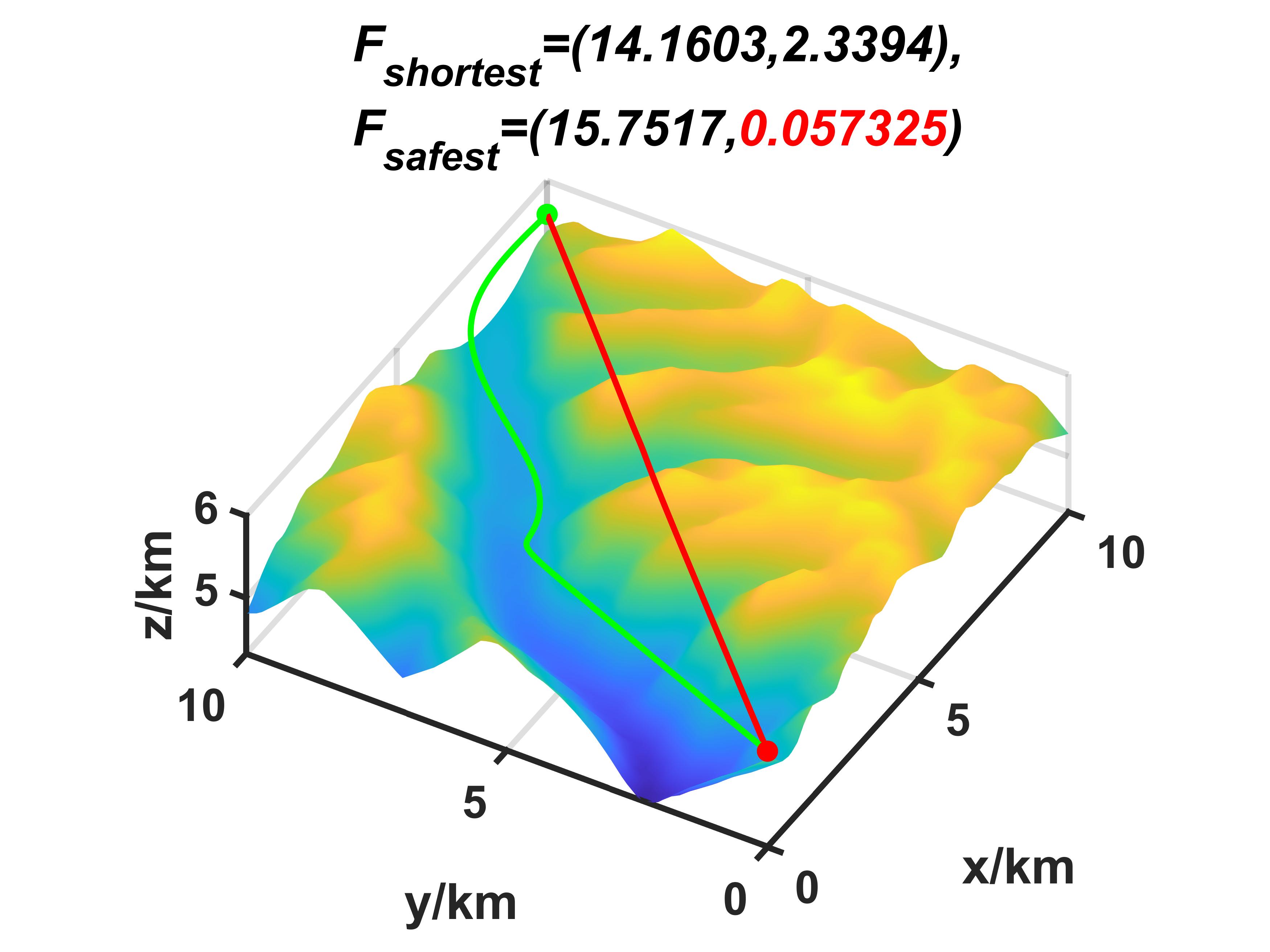}}\centerline{\footnotesize{(d3)}}
	\end{minipage}
    \begin{minipage}{0.24\linewidth}
		\vspace{3pt}\centerline{\includegraphics[width=4.5cm]{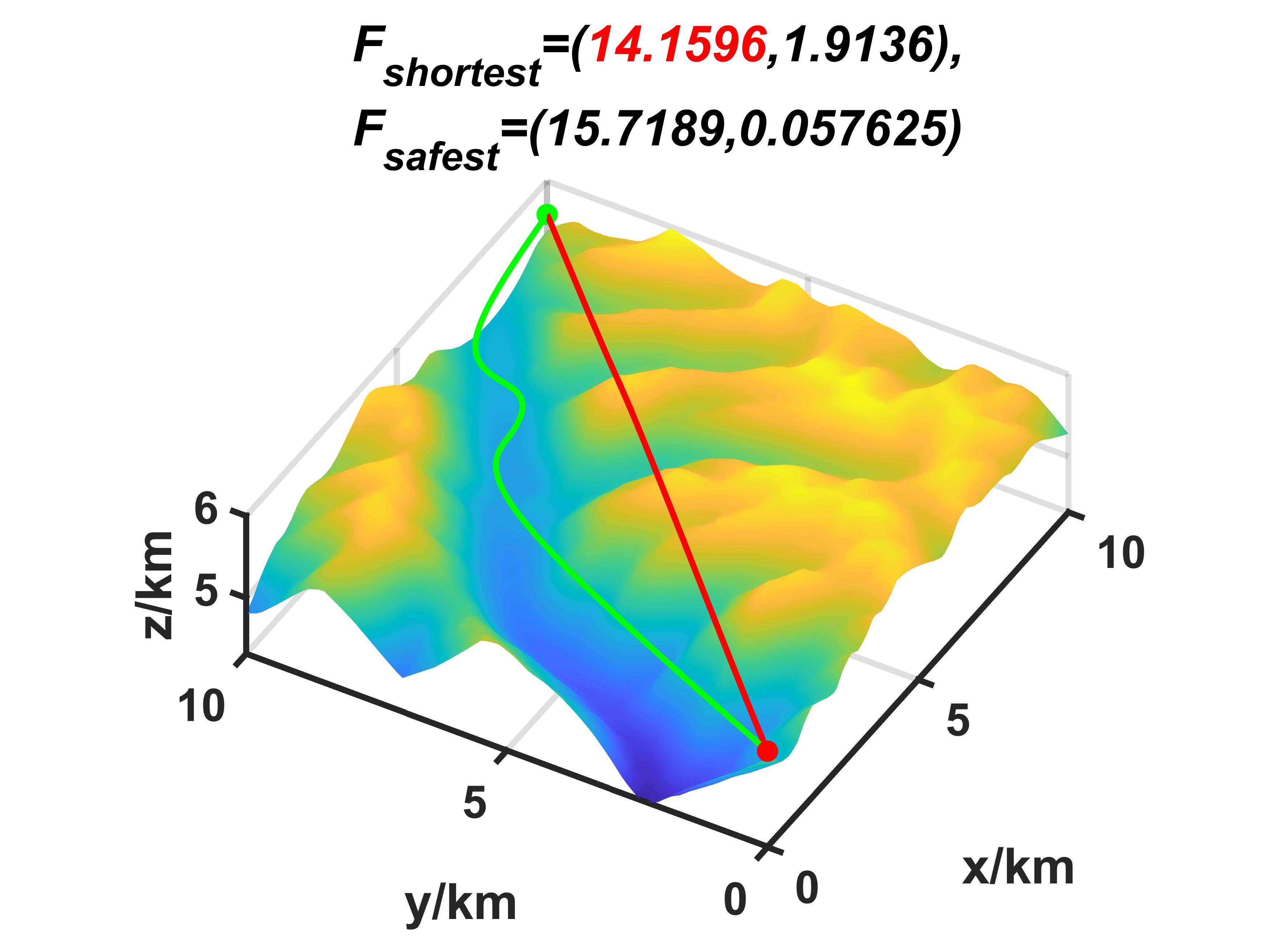}}\centerline{\footnotesize{(d4)}\vspace{1.5ex}}
	\end{minipage}
 
    \caption{The shortest path (red path) and the safest path (green path) in realistic scenarios. (a1)-(a4): C21; (b1)-(b4): C22; (c1)-(c4): C23; (d1)-(d4): C24. (a1)-(d1): NSGA-III; (a2)-(d2): C-MOEA/D; (a3)-(d3): MOEA/D-AWA; (a4)-(d4): MOEA/D-AAWA.}
    \label{fig11}
\end{figure}
The mean and Std values of HV and PD over 30 independent runs in all four real-data cases are shown in Table \ref{tbl4}. Sort and bold the data of the table in the same way. The proposed algorithm obtains the optimal result of both HV and PD mean values in cases C21, C23 and C24, and ranks first overall. Fig. \ref{fig10} presents the final nondominated solutions at the run with the median score. NSGA-III and C-MOEA/D without a weight adjustment strategy cannot reach the sharp peak and low tail. It seems that MOEA/D-AWA reaches these two areas, however, sometimes it can only reach one of them. Taking Fig. \ref{fig10}(a)-(c) as examples, the coverage of sharp peak by MOEA/D-AWA is insufficient. The proposed algorithm cannot only obtain feasible solutions in the middle area of PFs, but also simultaneously search for valuable individuals in sharp peak and low tail.

Fig. \ref{fig11} presents the shortest path (red path) and the safest path (green path) in all four real-data cases. As shown in Fig. \ref{fig11}(a2)-(d2), the shortest and safest path of C-MOEA/D are very similar. It is because that, its safest path is almost a straight line from the starting point to the target point, rather than avoiding dangerous terrain. The path of NSGA-III have the same defect shown in Fig.\ref{fig11} (a1) and (c1). Conversely, the safest paths of MOEA/D-AWA and the proposed algorithm reduce the collision possibility by bypassing from the sides of map or raising the height of the UAV, thereby improving the flight safety. However, the proposed algorithm obtains both the shortest path and the safest path in cases C21, C22 and C23, and the shortest flight paths in case C24. It indicates that the proposed algorithm has more superior performance.

\section{Conclusion}\label{cha6}

To solve UAV 3-D path planning problems with sharp peak and low tail PFs, we propose a multi-objective path planning method based on improved MOEA/D. In this paper, we introduce a multi-objective path planning problem minimizing overall path length and degree of threats for a UAV in 3-D task scenarios. In order to improve population diversity, we propose an adaptive areal weight adjustment strategy, i.e., deleting crowded subproblems in the population and adding new subproblems in the sparse area of the population. Furthermore, the newly-added weight vector is calculated by the objective function value of its neighbors to navigate the newly-added individual to the sparser area on PFs.

To verify the effectiveness of our proposed algorithm, we conduct extensive experiments under synthetic (mountain and urban environments) and realistic scenarios. The experiment results indicate that the proposed algorithm is superior to the other three MOEAs in terms of the convergence and diversity. Besides, the proposed algorithm has the capability of searching shorter paths and safer paths in both synthetic and realistic scenarios.

The improvement of the proposed method in future work mainly includes the following two aspects. First, it is necessary to verify the performance of MOEA/D-AAWA in many-objective optimization problems. Second, more realistic flight constraints and parameters should be simulated to make the simulation closer to reality, such as acceleration of UAVs, flight speed. Furthermore, the effectiveness of the method needs to be verified in other types of scenarios, such as mazes, combat environments with radars and artillery.







\printcredits





\end{document}